\documentclass[11pt,a4paper]{report}%

\usepackage{upthesis}         
\usepackage[latin1]{inputenc} 
\usepackage{epsf}             
\usepackage[noadjust]{cite}
\usepackage{latexsym, amsmath, amsfonts, amssymb}
\usepackage{epsfig}
\usepackage{subfigure}
\usepackage{mathrsfs}
\usepackage{color}
\usepackage{url}
\usepackage{fancyhdr}

\newcommand{\IN}{\mathbb{N}}
\newcommand{\IR}{\mathbb{R}}
\newcommand{\bbbr}{\mathbb{R}}
\newcommand{\abs}[1]{\lvert#1\rvert}

\usepackage{fancyhdr}
\renewcommand{\thefootnote}%
      {\fnsymbol{footnote}}

\usepackage{setspace}

\begin{document}
\pagenumbering{roman}

\title{Classification of Ordinal Data}
\submitionplace{Tese submetida à Faculdade de Ciências da \\
    Universidade do Porto para obtenção do grau de Mestre \\ 
    em Engenharia Matemática}
\author{Jaime dos Santos Cardoso}
\department{Departamento de Matemática Aplicada\\ Faculdade de
Ci\^encias da Universidade do Porto}
\faculty{Faculdade de Ci\^encias da Universidade do Porto}
\submitdate{Setembro de 2005}

\onehalfspacing

\beforepreface
\dedicationpage{To \Large${\cal T}ina$ and ${\cal B}ino$, my parents}
\section*{Preface} 
\addcontentsline{toc}{chapter}{Preface}
I have just gone through my email archive. The first exchange of messages with M. J. Cardoso, M. D., dates from the end of 2001. 
It was Pedro Cardoso, her brother and my superior at INESC Porto, where I was a researcher and developer since 1999, who had put us in contact. She had just started her PhD and would probably need some assistance from someone skilled in software development and mathematics. Although I was already enrolled to start my own PhD in the beginnings of 2002, I accepted. 

Simultaneously, since the end of 2000, I had been working at INESC Porto for the MetaVision project. The MetaVision project proposed an innovative electronic production system to reduce the cost of film production and to allow more artistic flexibility in shooting and film editing.  It also provided the enabling technology for the integration of real and virtual images at source quality for film production and in TV studios in the compressed domain.

2004 brought with it the end of the MetaVision project. That represented some free time that was exploited to fill some gaps detected in my mathematical background, by engaging in a masters in engineering mathematics. This master offers a solid formation in diverse areas of applied mathematics, divided in four main areas, comprising the analysis and processing of information.

Coincidently, 2004 would also be the year of tight collaboration with M. J. Cardoso. 
Her aim was to develop an objective measure for the overall cosmetic result of breast cancer conservative treatment. 
When confronted with the problem, a machine learning approach (a topic that I was delving in the master's classes) emerged as the right move. A suggestion was made to predict the overall cosmetic result from a few simple measures taken from the patient. I knew already some tools to tackle the problem but only superficially. That led me to select the \textsl{automatic classification and pattern recognition}, lectured by Professor Joaquim F. Pinto da Costa, as one of the modules to attend.

The application of some of the state of the art methods for ordinal data, to the problem at hand, sparkled the interest on this specific topic of classification. What I learned, the breakthroughs that were accomplished, is what I would like to share with you.

\vspace{0.4cm}
This work became possible due to the support of Professor Joaquim F. Pinto Costa, my supervisor. I also discussed ideas presented in this thesis with L. Gustavo Martins, Lu\'is F. Teixeira and M. Carmo Sousa. They also read the manuscript, providing important feedback. I would like to express my deep gratitude to all of them.

\vspace{0.4cm}
\emph{Jaime dos Santos Cardoso}\\
\emph{September 2005}
\cleardoublepage
\section*{Abstract}  
\addcontentsline{toc}{chapter}{Abstract}
Predictive learning has traditionally been a standard inductive learning, where different sub-problem formulations have been identified. One of the most representative is {\em classification}, consisting on the estimation of a mapping from the feature space into a finite class space. Depending on the cardinality of the finite class space we are left with binary or multiclass classification problems. Finally, the presence or absence or a ``natural'' order among classes will separate nominal from ordinal problems.

Although two-class and nominal classification problems have been dissected in the literature, the ordinal sibling has not yet received a lot of attention, even with many learning problems involving classifying examples into classes which have a natural order.
Scenarios in which it is natural to rank instances occur in many fields, such as information retrieval, collaborative filtering, econometric modeling and natural sciences. 

Conventional methods for nominal classes or for regression problems could be employed to solve ordinal data problems; however, the use of techniques designed specifically for ordered classes yields simpler classifiers, making it easier to interpret the factors that are being used to discriminate among classes, and generalises better. 
Although the ordinal formulation seems conceptually simpler than nominal, some technical difficulties to incorporate in the algorithms this piece of additional information -- the order -- may explain the widespread use of conventional methods to tackle the ordinal data problem.

This dissertation addresses this void by proposing a nonparametric procedure for the classification of ordinal data based on the extension of the original dataset with additional variables, reducing the classification task to the well-known two-class problem. 
This framework unifies two well-known approaches for the classification of ordinal categorical data, the minimum margin principle and the generic approach by Frank and Hall. It also presents a probabilistic interpretation for the neural network model.
A second novel model, the unimodal model, is also introduced and a parametric version is mapped into neural networks. 
Several case studies are presented to assert the validity of the proposed models. 
\paragraph*{Keywords:} machine learning, classification, ordinal data, neural networks, support vector machines

\cleardoublepage
\section*{Resumo}  
\addcontentsline{toc}{chapter}{Resumo}

Tradicionalmente, a aprendizagem automática predictiva tem sido uma aprendizagem indutiva padrão, onde diferentes sub-problemas foram sendo formulados. 
Um dos mais representativos é o da classificação, que consiste na estimação de uma função do espaço dos atributos para um espaço finito de classes. 
Dependendo da cardinalidade do espaço das classes temos um problema de classificação binário ou multi-classe. Finalmente, a existência ou ausência de uma ordem ``natural'' entre as classes distingue problemas multi-classe nominais de problemas multi-classe ordinais.

Embora os problemas de classificação binária e multi-classe nominal tenham sido dissecados na literatura, o problema-irmão de dados ordinais tem passado despercebido, mesmo com muitos problemas de aprendizagem automática envolvendo a classificação de dados que possuem uma ordem natural. Cenários em que é natural ordenar exemplos ocorrem nas mais diversas áreas, tais como pesquisa ou recuperação de informação, filtragem colaborativa, modelação económica e ciências naturais.

Os métodos convencionais para classes nominais ou para problemas de regressão podem ser empregues para resolver o problema ordinal; contudo, a utilização de técnicas desenvolvidas especificamente para classes ordenadas produz classificadores mais simples, facilitando a interpretação dos factores que estão a desempenhar um papel importante para discriminar as classes, e generaliza melhor. 
Embora a formulação ordinal aparente ser conceptualmente mais simples que a nominal, algumas dificuldades técnicas para incorporar nos algoritmos este pedaço de informação adicional -- a ordem -- pode explicar o uso generalizado de métodos convencionais para atacar o problema de dados ordinais.

Esta dissertação aborda este vazio, propondo um método não-paramétrico para a classificação de dados ordinais baseado na extensão do conjunto de dados original com variáveis adicionais, reduzindo o problema de classificação ao familiar problema de classificação binária. A metodologia proposta unifica duas abordagens bem estabelecidas para a classificação de dados ordinais, o princípio da margem mínima e o método genérico de Frank e Hall. É também apresentado uma interpretação probabilística para o modelo mapeado em redes neuronais. Um segundo modelo, o modelo unimodal, é também introduzido e uma versão paramêtrica é mapeada em redes neuronais. Vários casos de estudo são apresentados para evidenciar a validade dos modelos propostos. 
 
\paragraph*{Palavras-chave:} aprendizagem automática, classificação, dados ordinais, redes neuronais, máquinas de vectores de suporte

\afterpreface
\pagestyle{fancy}

\chapter{Introduction}
\setcounter{footnote}{1}
\section{Problem formulation}
Predictive learning has traditionally been a standard inductive learning, with two modes of inference: system identification (with the goal of density estimation) and system imitation (for generalization). 
Nonetheless, predictive learning does not end with inductive learning. While with inductive learning the main assumptions are a finite training set and a large (infinite), unknown test set, other problem settings may be devised.

The transduction formulation \cite{Vapnik1998} assumes a given set of labeled, training data and a {\em finite}, {\em known} set of unlabeled test points, with the interest to estimate the class labels {\em only} at these points.
The selection type of inference is, in some sense, even simpler than transduction: given a set of labeled training data and unlabeled test points, select a subset of test points with the highest probability of belonging to one class. Selective inference needs only to select a subset of $m$ test points, rather than assign class labels to all test points.
An hierarchy of types of inference can be, not exhaustively, listed \cite{tutorialIJCNN2005}:
identification, imitation, transduction, selection, etc. 

Under the traditional inductive learning, different (sub-)problem formulations have been identified. Two of the most representative are {\em regression} and {\em classification}. While both consist on estimating a mapping from the feature space, the regression looks for a real-valued function defined in the feature space, whereas classification maps the feature space into a finite class space. 
Depending on the cardinality of the finite class space we are left with two-class or multiclass classification problems. Finally, the presence or absence of a ``natural'' order among classes will separate nominal from ordinal problems:

\begin{figure}
\begin{center}
\includegraphics[width=0.6\linewidth]{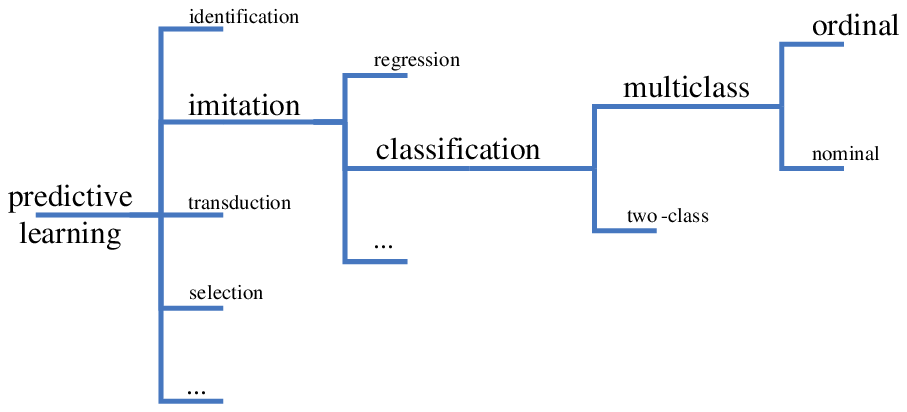}
\label{PredictiveLearning}
\end{center}
\end{figure}

\section{Motivation}
Although two-class and nominal data classification problems have been dissected in the literature, the ordinal sibling has not yet received a lot of attention, even with many learning problems involving classifying examples into classes which have a ``natural'' order.
Settings in which it is natural to rank instances
arise in many fields, such as information retrieval \cite{Herbrich1999B}, collaborative filtering \cite{Shashua2002A}, econometric modeling \cite{Mathieson1995A} and natural sciences \cite{JaimeNN2005}.\footnote{It is worth pointing out that distinct tasks of relation learning, where an example is no longer associated with a class or rank, which include preference learning and reranking \cite{Shen2005}, are topics of research on their own.}

Conventional methods for nominal classes or for regression problems could be employed to solve ordinal data problems (\cite{Duda2001,Theodoridis2003,Ripley1996}); however, the use of techniques designed specifically for ordered classes results in simpler classifiers, making it easier to interpret the factors that are being used to discriminate among classes \cite{Mathieson1995A}. 
Although the ordinal formulation seems conceptually simpler than nominal, some technical difficulties to incorporate the piece of additional information -- the order -- in the algorithms may explain the widespread use of conventional methods to tackle the ordinal data problem.

\section{Tools}
As seen, there are relatively few predictive learning formulations; however, the number of learning algorithms, especially for the inductive case, is overwhelming. Many frameworks, adaptations to real-life problems, intertwining of base algorithms were, and continue to be, proposed in the literature; 
ranging from statistical approaches to state of the art machine learning algorithms, parametric to non parametric procedures, a plethora of methods is available to users.

Our study will not attempt to cover them all. Limited by time (and competence to add significant contributions), two major algorithms will be the ``horsepower'' of our work: support vector machines and neural networks. Other base approaches, such as decision trees, for which interesting algorithms for ordinal data have already been proposed (\cite{Lee2002,Lee2003,Potharst1999}), will have to wait for a next opportunity.

\subsection{The ABC of support vector machines}
Consider briefly how the SVM binary classification
problem is formulated \cite{Vapnik1998}.\footnote{The following introduction to SVMs is based largely on \cite{Shilton2005}.} 

For two training classes linearly separable in the selected feature space, the distinctive idea of SVM is to define a linear discriminant function $g(\textbf{x}) = \textbf{w}^t\textbf{x}+b$ in the feature space bisecting the two training classes and 
characterized by $g(\textbf{x}) = 0$.
However, there may be infinitely
many such surfaces. To select the surface best suited to the task,
the SVM maximizes the distance between the decision surface
and those training points lying closest to it (the support vectors).
Considering the training set 
$\{\textbf{x}_i^{(k)}\}$, where $k=1, 2$ denotes the class number, $i=1,\cdots,\ell_k$ is the index within
each class, it is easy to show \cite{Vapnik1998} that maximizing this
distance is equivalent to solving
\begin{equation}
\label{eq:stdsvm1}
\begin{split}
& \min_{\textbf{w}, b} \quad \frac{1}{2}\textbf{w}^t\textbf{w} \\
& \begin{array}{ll}
s.t. 	& \begin{array}{ll} -(\textbf{w}^t\textbf{x}_i^{(1)} + b) \geq +1 & i = 1,\cdots,\ell_1\\
			 +(\textbf{w}^t\textbf{x}_i^{(2)} + b) \geq +1 & i = 1,\cdots,\ell_2
\end{array}\end{array} 
\end{split}
\end{equation}

If the training classes are not linearly separable in feature
space, the inequalities in \eqref{eq:stdsvm1} can be relaxed using slack variables
and the cost function modified to penalise any failure to meet the
original (strict) inequalities. The problem becomes
\begin{equation}
\label{eq:stdsvm2a}
\begin{split}
&\min_{\textbf{w}, b, \xi_i} \quad \frac{1}{2}\textbf{w}^t\textbf{w} + C\sum_{k=1}^2\sum_{i=1}^{\ell_k} \mbox{ sgn }(\xi_i^{(k)}) \\
&\begin{array}{ll}
s.t. & \begin{array}{ll}-(\textbf{w}^t\textbf{x}_i^{(1)} + b) \geq +1- \xi_i^{(1)} & i = 1,\cdots,\ell_1\\
			 +(\textbf{w}^t\textbf{x}_i^{(2)} + b) \geq +1- \xi_i^{(2)} & i = 1,\cdots,\ell_2\\
			 \xi_i^{(k)} \geq 0 & 
\end{array}
\end{array}
\end{split}
\end{equation}

The constraint
parameter $C$ controls the tradeoff between the dual objectives
of maximizing the margin of separation and minimizing
the misclassification error.
For an error to occur, the corresponding $\xi_i$ must exceed unity so $\sum_{k=1}^2 \sum_{i=1}^{\ell_k} \mbox{ sgn }(\xi_i^{(k)})$ is 
an upper bound on the number of the training errors, that is $\sum l_{0-1}(f(\textbf{x}_i^{(k)}), k)$, where
 $f(\textbf{x}_i^{(k)})$ is the classification rule induced by the hyperplane $\textbf{w}^t\textbf{x}+b$.
Hence the added penalty component is a natural way to assign an extra cost for errors. 

However, optimization of the above is difficult since it involves a
discontinuous function $\mbox{ sgn }()$. As it is common in such cases, we choose to optimize
a closely related cost function, and the goal becomes to 
\begin{equation}
\label{eq:stdsvm3}
\min_{\textbf{w}, b, \xi_i} \quad \frac{1}{2}\textbf{w}^t\textbf{w} + C\sum_{k=1}^2\sum_{i=1}^{\ell_k} \xi_i^{(k)} \\
\end{equation}
under the same set of constraints as \eqref{eq:stdsvm2a}.

In order to account for different misclassification costs or sampling bias, the model can be extended
to penalise the slack variables according to different weights in the objective function \cite{Lin2002}:
\begin{equation}
\label{eq:stdsvm4}
\min_{\textbf{w}, b, \xi_i} \quad \frac{1}{2}\textbf{w}^t\textbf{w} + \sum_{k=1}^2\sum_{i=1}^{\ell_k}C_i \xi_i^{(k)} \\
\end{equation}

\subsection{The ABC of neural networks}
Neural networks were originally developed from attempts to model the communication and processing information in the human brain.
Analogous to the brain, a neural network consists of a number of inputs (variables), each of which is multiplied by a weight, 
which is analogous to a dendrite. The products are summed and transformed in a ``neuron'' (i.e. simple processing unit) and the result becomes an input value for another neuron \cite{Thomas2002}.

A multilayer feedforward neural network consists of an input layer of signals, an output layer of output signals, and a number of layers
of neurons in between, called hidden layers \cite{Haykin1999,Tsoukalas1996,Bishop1995}. 
It was shown that, under mild conditions, these models can approximate any decision function and its derivatives to any degree of accuracy.

To use a neural network for classification, we need to construct an equivalent function approximation problem by assigning a target value for each class. For a two-class problem we can use a network with a single output, and binary target values: 1 for one class, and 0 for the other. We can thus interpret the network's output as an estimate of the probability that a given pattern belongs to the '1' class.
The training of the network is commonly performed using the popular mean square error.

For multiclass classification problems ($1$-of-$K$, where $K > 2$) we use a network with $K$ outputs, one corresponding to each class, and target values of 1 for the correct class, and 0 otherwise. Since these targets are not independent of each other, however, it is no longer appropriate to use the same error measure. 
The correct generalization is through a special activation function (the \emph{softmax}) designed so as to satisfy the normalization constraint on the total probability \cite{Ripley1996}.

However, this approach does not retain the ordinality or rank order of the classes and is not, therefore, appropriate for ordinal multiclass classification problems. 
An clear exception is the PRank algorithm by Crammer \cite{Crammer2001}, and its improvement by Harrington \cite{Harrington2003}, which is a variant of the perceptron algorithm. As we progress in this work, several other approaches will be presented, making use of generic neural networks.

\section{Thesis' structure}
This thesis introduces in chapter \ref{dataReplication} the data replication method, a nonparametric procedure for the classification of ordinal data based on the extension of the original dataset with additional variables, reducing the classification task to the well known two-class problem. Starting with the simpler linear case, the chapter evolves to the nonlinear case; from there the method is extended to incorporate the procedure of Frank and Hall \cite{Frank1999}. Finally, the generic version of the data replication method is presented, allowing partial constraints on variables.

In chapter \ref{Mapping} the data replication method is mapped into two important machine learning algorithms: support vector machines and neural networks. A comparison is made with a previous SVM approach introduced by Shashua \cite{Shashua2002A}, the minimum margin principle, showing that the data replication method leads {\em essentially} to the same solution, but with some key advantages. The chapter is elegantly concluded with a reinterpretation of the neural network model as a generalization of the ordinal logistic regression model.

The second novel model, the unimodal model, is introduced in chapter \ref{chap:unimodal}, and a parametric version is mapped into neural networks. A parallelism of this approach with regression models concludes the chapter.

Chapter \ref{chap:experimental} introduces the experimental methodology and the algorithms that were compared in the conducted experiments reported in the succeeding chapters. Finally, results are discussed, conclusions are drawn and future work is oriented in chapter \ref{chap:discussion}.

\section{Contributions}
We summarize below the contributions of this thesis towards more efficient and parsimonious methods for classification of ordinal data. In this thesis we have
\begin{enumerate}
\item introduced in the machine learning community the \emph{data replication method}, a nonparametric procedure for the classification of ordinal categorical data. Presented also the mapping of this method for neural networks and support vector machines; 

\item unified under this framework two well-known approaches for the classification of ordinal categorical data, the minimum margin principle \cite{Shashua2002A} and the generic approach by Frank and Hall \cite{Frank1999}. It was also presented a probabilistic interpretation for the neural network model;

\item introduced the unimodal model, mapped to neural networks, a second approach for the classification of ordinal data, and established links to previous works.

\end{enumerate}

\subsection*{Publications related to the thesis}

\cite{JaimeIJCNN2005} J.~S. Cardoso, J.~F.~P. da~Costa, and M.~J. Cardoso, ``{SVM}s applied to
  objective aesthetic evaluation of conservative breast cancer treatment,'' in
  \emph{Proceedings of International Joint Conference on Neural Networks
  {(IJCNN)} 2005}, 2005, pp. 2481--2486.
  
\cite{JaimeNN2005} J.~S. Cardoso, J.~F.~P. da~Costa, and M.~J. Cardoso, ``Modelling ordinal
  relations with {SVMs}: an application to objective aesthetic evaluation of
  breast cancer conservative treatment,'' \emph{{(ELSEVIER)}Neural Networks},
  vol.~18, pp. 808--817, june-july 2005. 

\cite{JFPCostaECML2005} J.~F.~P. da~Costa and J.~S. Cardoso, ``Classification of ordinal data using
  neural networks,'' in \emph{Proceedings of European Conference Machine
  Learning {(ECML)} 2005}, 2005, pp. 690--697.
    
\cite{JaimeMachineLearning2006} J.~S. Cardoso and J.~F.~P. da~Costa, ``Learning to classify ordinal data: the
  data replication method,'' \emph{(submitted) Journal of Machine Learning Research}.

\chapter[The data replication method]{The data replication method\footnotemark[4]}\footnotetext[4]{Some portions of this chapter appeared in \cite{JaimeNN2005}.}
\label{dataReplication}

\setcounter{footnote}{1}

Let us formulate the problem of separating $K$ ordered classes ${\cal C}_1, \cdots, {\cal C}_K$.
Consider the training set 
$\{\textbf{x}_i^{(k)}\}$, where $k=1, \cdots, K$ denotes the class number, $i=1,\cdots,\ell_k$ is the index within
each class, and $\textbf{x}_i^{(k)} \in \IR^p$, with $p$ the dimension of the feature space. Let $\ell = \sum_{k=1}^K \ell_{k}$ be the total number of training examples.
 
Suppose that a $K$-class classifier was forced, by design, to have $K-1$ \emph{noncrossing boundaries}, with boundary $i$ discriminating classes ${\cal C}_1, \cdots, {\cal C}_i$ against classes ${\cal C}_{i+1}, \cdots, {\cal C}_K$. 
As the intersection point of two boundaries would indicate an example with three or more classes equally probable -- not plausible with ordinal classes --, this strategy imposes an (arguably) intuitive restriction.
With this constraint emerges a monotonic model, where a better value in an attribute does not lead to a lower decision class.
For the linear case, this translates to choosing the same weighted sum for all decisions -- the classifier would be just a set of weights, one for each feature, and a set of biases, the {\em scale} in the weighted sum. 
By avoiding the intersection of any two boundaries, this simplified model captures better the essence of the ordinal data problem.  
Another strength of this approach is the reduced number of parameters to estimate, which may lead to a more robust 
classifier, with greater capacity for generalization.

This rationale leads to a straight-forward generalization of the two-class separating hyperplane \cite{Shashua2002A}. 
Define $K-1$ separating hyperplanes that separate the training data into $K$ ordered classes by modeling the
ranks as intervals on the real line -- an idea with roots in the classical cumulative model, \cite{Herbrich1999B,McCullagh1989}.
The geometric interpretation of this approach is to look for $K-1$ parallel hyperplanes represented by vector $\textbf{w}\in \IR^p$ and scalars $b_1, \cdots, b_{K-1}$,
such that the feature space is divided into equally ranked regions by the decision boundaries
$\textbf{w}^t\textbf{x}+b_r, \ r = \{1, \cdots, K-1\}$. 

It would be interesting to accommodate this formulation under the two-class problem. That would allow the use of mature and optimized algorithms, developed for the two-class problem. The data replication method allows us to do precisely that.
\section{Data replication method -- the linear case}
\label{sec:linearreplication}
\begin{figure}[!ht]
\begin{center}
\subfigure[Original dataset in $\IR^2$, $K=3$.]{
        \label{fig:proposedModel1}
        \includegraphics[width=0.45\linewidth]{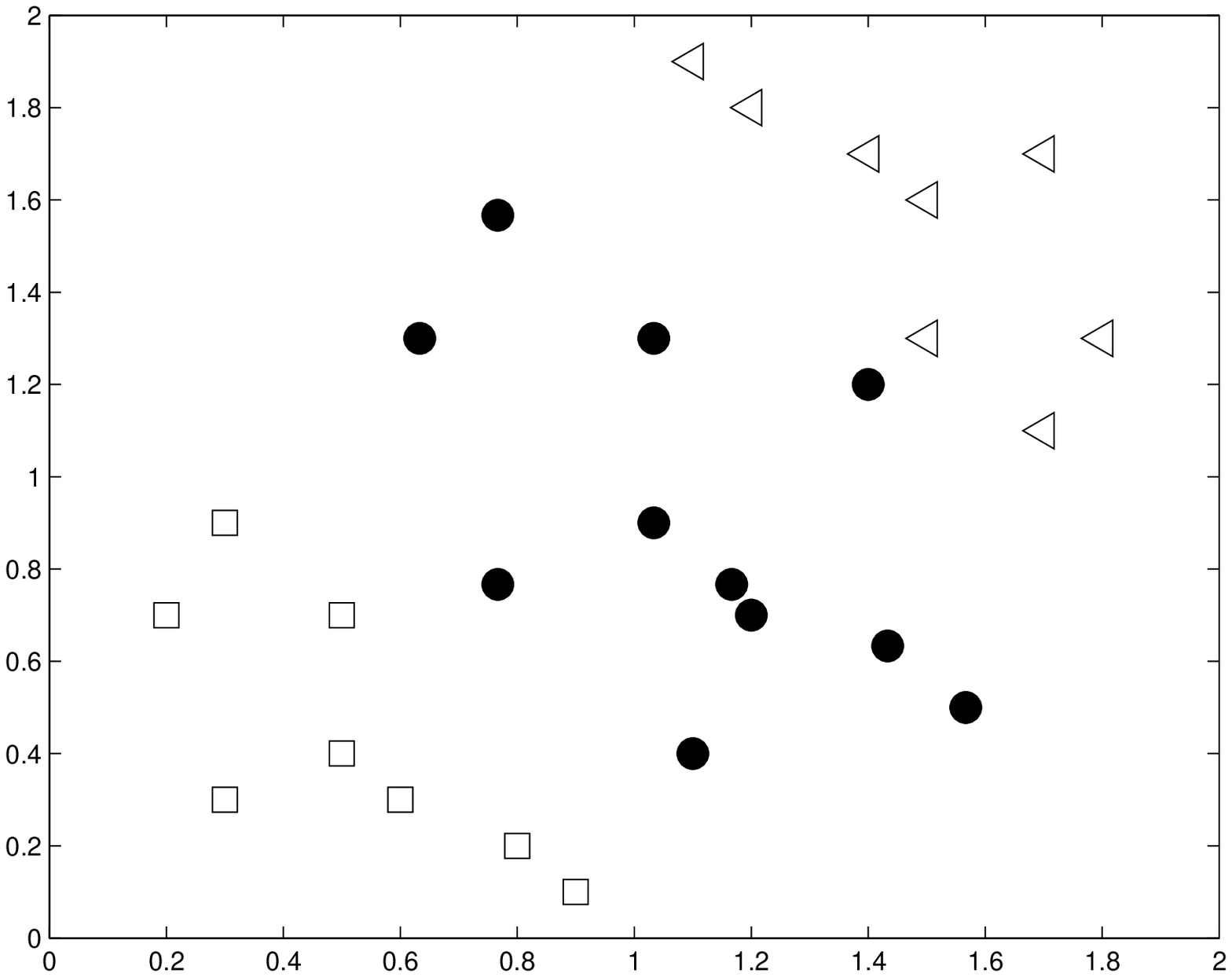}}
\subfigure[Data set in $\IR^3$, with samples replicated ($h=1$).]{
        \label{fig:proposedModel2}
        \includegraphics[width=0.45\linewidth]{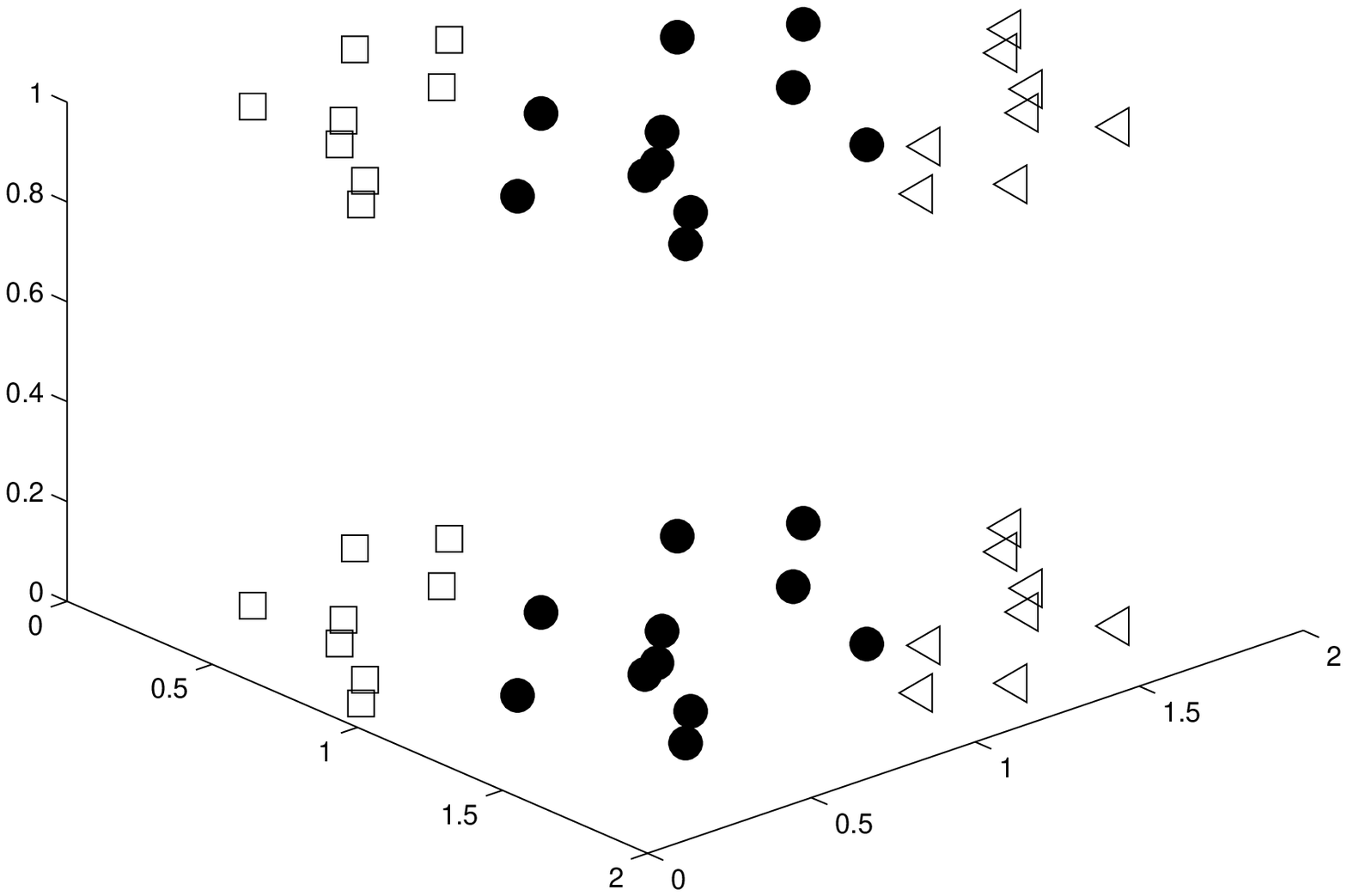}}
\subfigure[Transformation into a binary classification problem.]{
        \label{fig:proposedModel3}
        \includegraphics[width=0.45\linewidth]{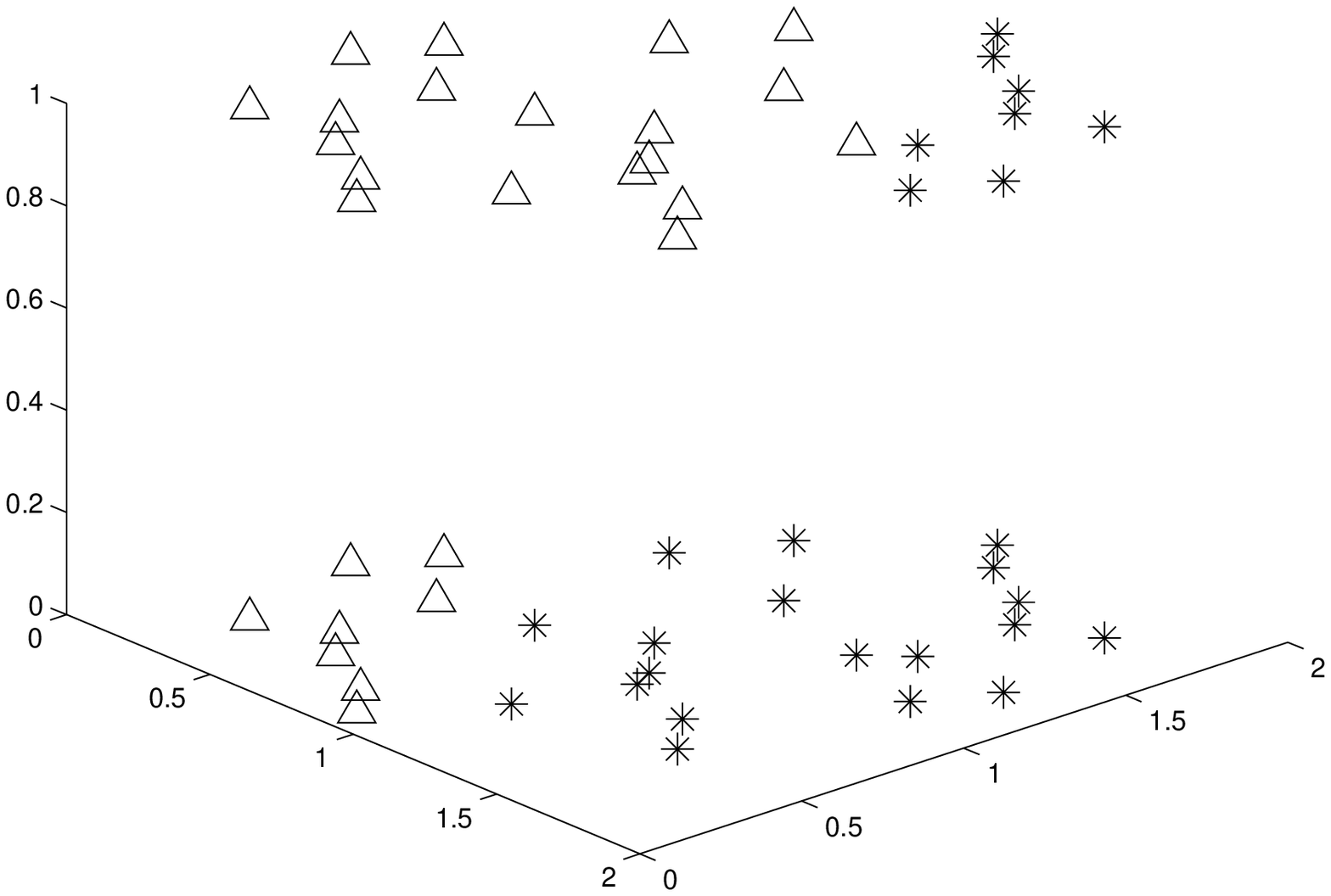}}
\subfigure[Linear solution to the binary problem.]{
        \label{fig:proposedModel4}
        \includegraphics[width=0.45\linewidth]{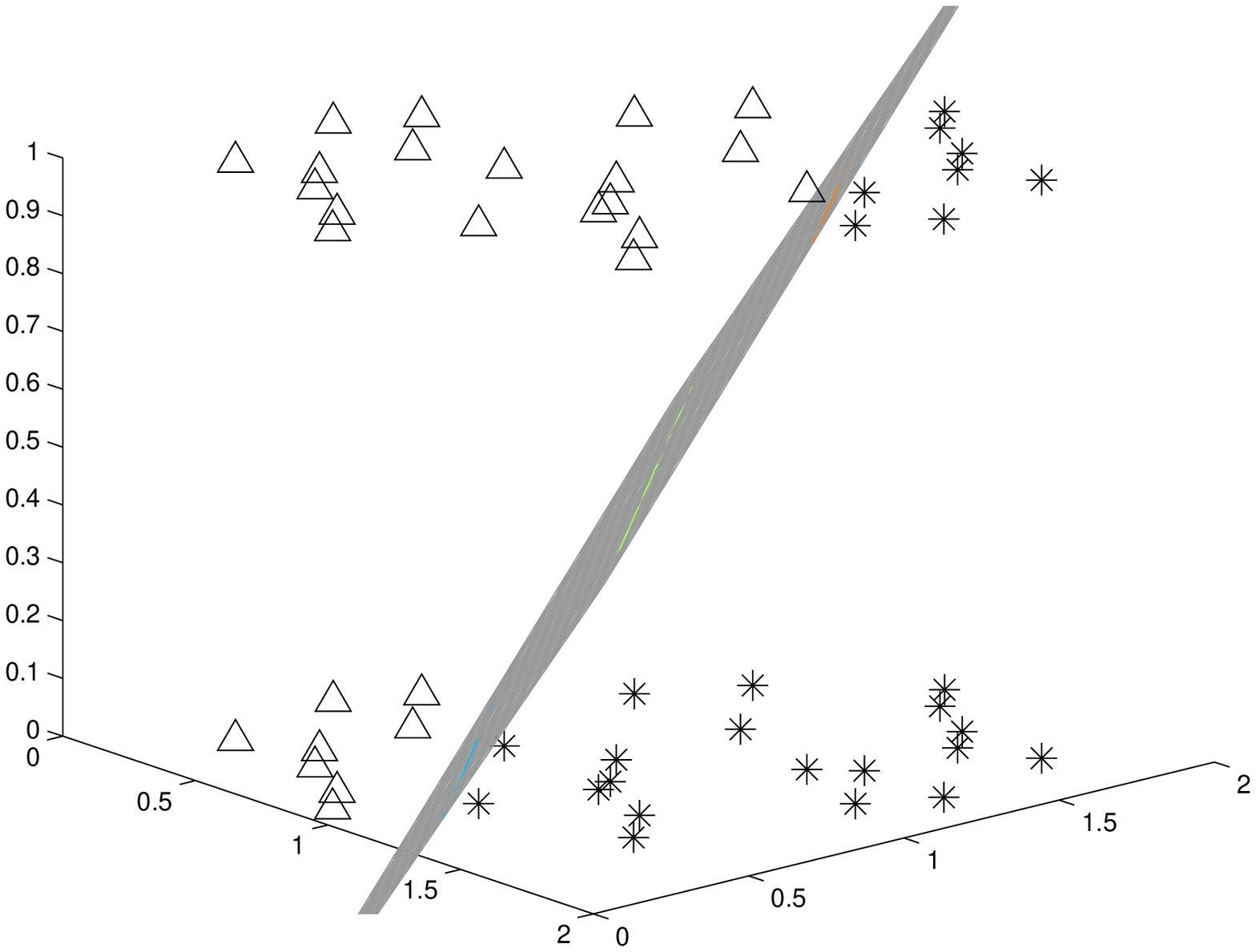}}
\subfigure[Linear solution in the original dataset.]{
        \label{fig:proposedModel5}
        \includegraphics[width=0.45\linewidth]{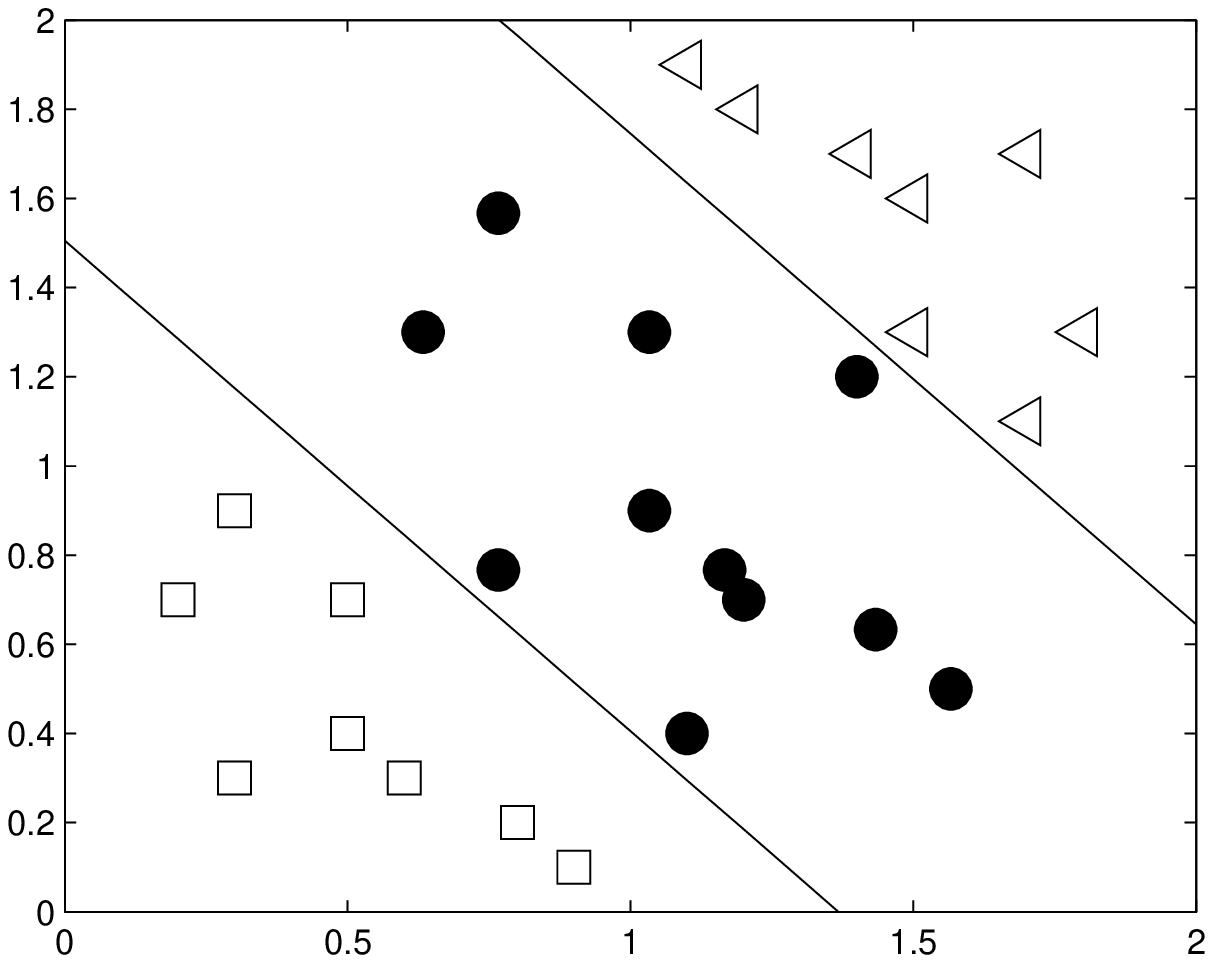}}
\caption{Proposed data extension model in a toy example.}
\label{fig:proposedModel}
\end{center}
\end{figure}

To outline the rationale behind the proposed model for the linear case, consider first an hypothetical, simplified scenario with three classes
in $\IR^2$.
The plot of the dataset is presented in figure \ref{fig:proposedModel1}.

Using a transformation from the $\IR^2$ initial feature-space to a $\IR^3$ feature space, 
replicate each original point, according to the rule (figure \ref{fig:proposedModel2}):
\[
\textbf{x} \in \IR^2 
\begin{matrix}
\nearrow \\
\searrow 
\end{matrix}
\begin{matrix}
\left[\begin{smallmatrix}\textbf{x}\\ 0\end{smallmatrix}\right] \in \IR^{3}\\
\\
\left[\begin{smallmatrix}\textbf{x}\\ h\end{smallmatrix}\right] \in \IR^{3}
\end{matrix}, \text{ where $h=$ const} \in \IR
\]
Observe that each any two points created from the same starting point differ only in the new variable.

Define now a binary training set in the high-dimensional space according to (figure \ref{fig:proposedModel3}):
\begin{equation}
\label{toyReplication}
\left[\begin{smallmatrix}\textbf{x}_i^{(1)}\\ 0\end{smallmatrix}\right], 
\left[\begin{smallmatrix}\textbf{x}_i^{(1)}\\ h\end{smallmatrix}\right], 
\left[\begin{smallmatrix}\textbf{x}_i^{(2)}\\ h\end{smallmatrix}\right] \in \overline{{\cal C}}_1 \quad
\left[\begin{smallmatrix}\textbf{x}_i^{(2)}\\ 0\end{smallmatrix}\right], 
\left[\begin{smallmatrix}\textbf{x}_i^{(3)}\\ 0\end{smallmatrix}\right], 
\left[\begin{smallmatrix}\textbf{x}_i^{(3)}\\ h\end{smallmatrix}\right] \in \overline{{\cal C}}_2 
\end{equation}
A linear two-class classifier can now be applied to the extended dataset, yielding a hyperplane separating the two classes -- figure \ref{fig:proposedModel4}.
The intersection of this hyperplane with each of the subspace replicas (by setting $x_3=0$ and $x_3=h$ in the equation of the hyperplane) can be used to derive the boundaries in the original dataset -- figure \ref{fig:proposedModel5}.

\begin{figure}[!ht]
\begin{center}
\includegraphics[width=0.45\linewidth]{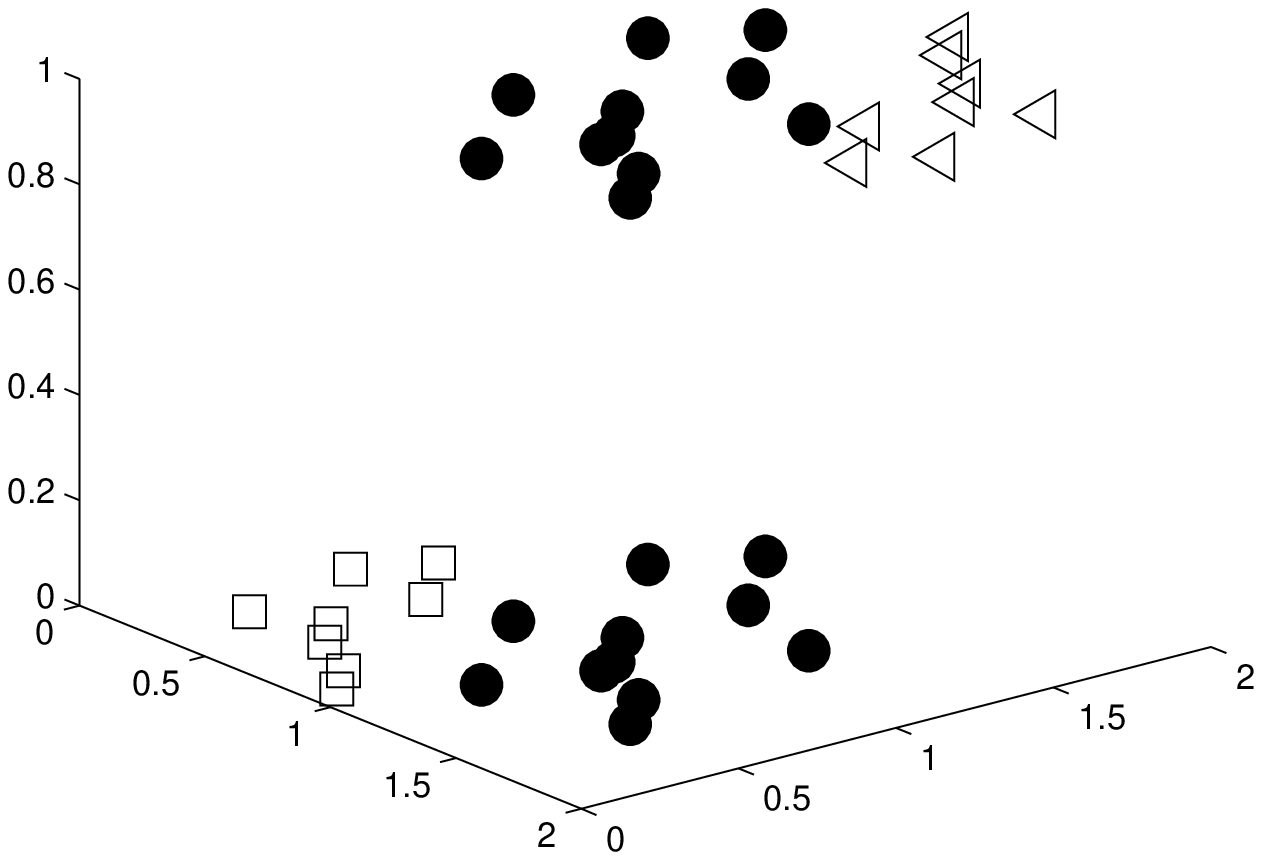}
\caption{Toy dataset replicated in $\IR^3$, $h=1$, $s=1$.}
\label{fig:proposedModels1}
\end{center}
\end{figure}

Although the foregoing analysis enables to classify unseen examples in the original dataset, classification can be done directly in the extended dataset, using the binary classifier, without explicitly resorting to the original dataset. 
For a given example $\in \IR^2$, classify each of its two replicas $\in \IR^3$, obtaining a sequence of two labels $\in \{\overline{{\cal C}}_1, \overline{{\cal C}}_2\}^2$. From this sequence infer the class according to the rule 
\[
\overline{{\cal C}}_1 \overline{{\cal C}}_1 \Longrightarrow {\cal C}_1\quad\quad
\overline{{\cal C}}_2 \overline{{\cal C}}_1 \Longrightarrow {\cal C}_2\quad\quad
\overline{{\cal C}}_2 \overline{{\cal C}}_2 \Longrightarrow {\cal C}_3
\]
With the material on how to construct a set of optimal hyperplanes for the toy example, we are now in a position to formally describe the construction of a $K$-class classifier for ordinal classification.
Define $\textbf{e}_0$ as the sequence of $K-2$ zeros and $\textbf{e}_{q}$ as the sequence of $K-2$ symbols $0,\cdots, 0, h, 0, \cdots, 0$, with $h$ in the $q$-th position.
Considering the problem of separating $K$ classes ${\cal C}_1, \cdots, {\cal C}_K$ with training set 
$\{\textbf{x}_i^{(k)}\}$, define a new high-dimensional binary training dataset as\\ 
\begin{footnotesize}
\begin{equation}
\label{generalExt}
\begin{array}{ll} 
\left[\begin{smallmatrix}\textbf{x}_i^{(k)}\\ \textbf{e}_0\end{smallmatrix}\right] &\in 
\begin{cases} 
\overline{{\cal C}}_1 \quad k=1\\
\overline{{\cal C}}_2 \quad k=2, \cdots, \min(K, 1+s)\\
\end{cases}\\
&\vdots\\
\left[\begin{smallmatrix}\textbf{x}_i^{(k)}\\\textbf{e}_{q-1}\end{smallmatrix}\right] &\in 
\begin{cases} 
\overline{{\cal C}}_1 \quad k = \max(1, q-s+1), \cdots, q\\
\overline{{\cal C}}_2 \quad k = q+1, \cdots, \min(K, q+s)\\
\end{cases}\\
&\vdots\\
\left[\begin{smallmatrix}\textbf{x}_i^{(k)}\\ \textbf{e}_{K-2}\end{smallmatrix}\right] &\in 
\begin{cases} 
\overline{{\cal C}}_1 \quad k= \max(1,K-1-s+1), \cdots, K-1\\
\overline{{\cal C}}_2 \quad k=K\\
\end{cases}
\end{array}
\end{equation}
\end{footnotesize}
\noindent where the role of parameter $s \in \{1, \cdots, K-1\}$ is to bound the number of classes, to the `left' and to the `right', involved in the constraints of a boundary.
This allows to control the increase of data points inherent to this method.
The toy example in figure \ref{fig:proposedModel2} was illustrated with $s=K-1=2$; setting $s=1$ would result as illustrated in \ref{fig:proposedModels1}, with essentially the same solution.

Then construct a linear two-class classifier on this extended dataset; to classify an unseen example obtain a sequence of $(K-1)$ labels $\in \{\overline{{\cal C}}_1, \overline{{\cal C}}_2\}^{(K-1)}$ by classifying each of the $(K-1)$ replicas in the extended dataset with the binary classifier.
Note that, because the $(K-1)$ boundaries do not cross each other, there are only $K$ different possible sequences. 
The target class can be obtained by summing one to the number of $\overline{{\cal C}}_2$ labels in the sequence.

\section{Data replication method -- the nonlinear case}
So far we have assumed linear boundaries between classes. There are important situations in which such a restriction does not exist, but the order of the classes is kept. 
Inspired by the data replication method just presented, we can look for  boundaries that are \emph{level curves} of some nonlinear function $G(\textbf{x})$ defined in the feature space. 
For the linear version we take $G(\textbf{x}) = \textbf{w}^t\textbf{x}$. 

\begin{figure}[!ht]
\begin{center}
\subfigure[Nonlinear solution to the binary problem. $\overline{G}(\overline{\textbf{x}}) = 0.4(x_1^2+x_2^2 - 1) + x_3$]{
        \includegraphics[width=0.45\linewidth]{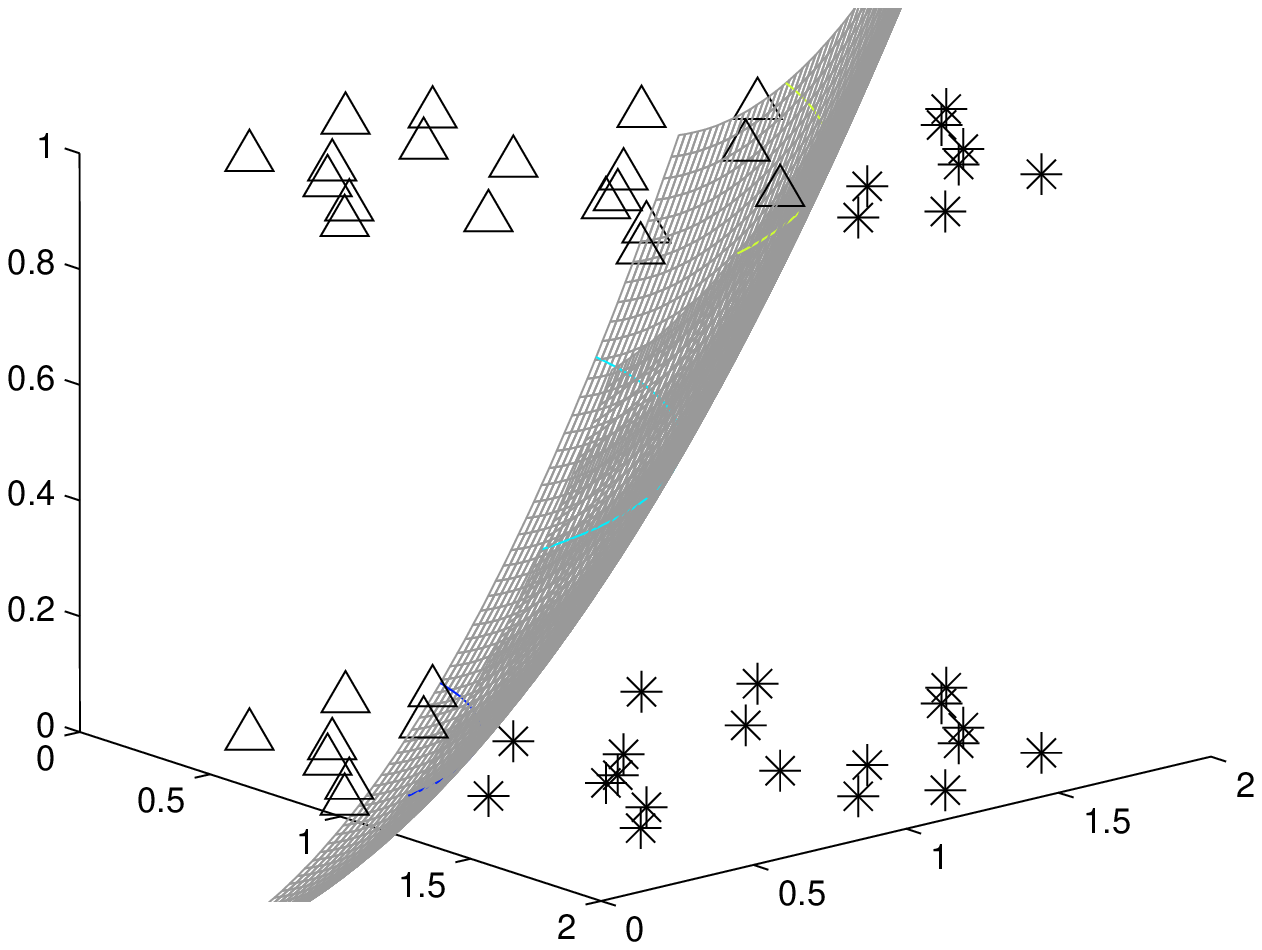}}
\subfigure[Nonlinear solution in the original dataset. $G(\textbf{x}) = x_1^2+x_2^2 - 1$]{
        \includegraphics[width=0.45\linewidth]{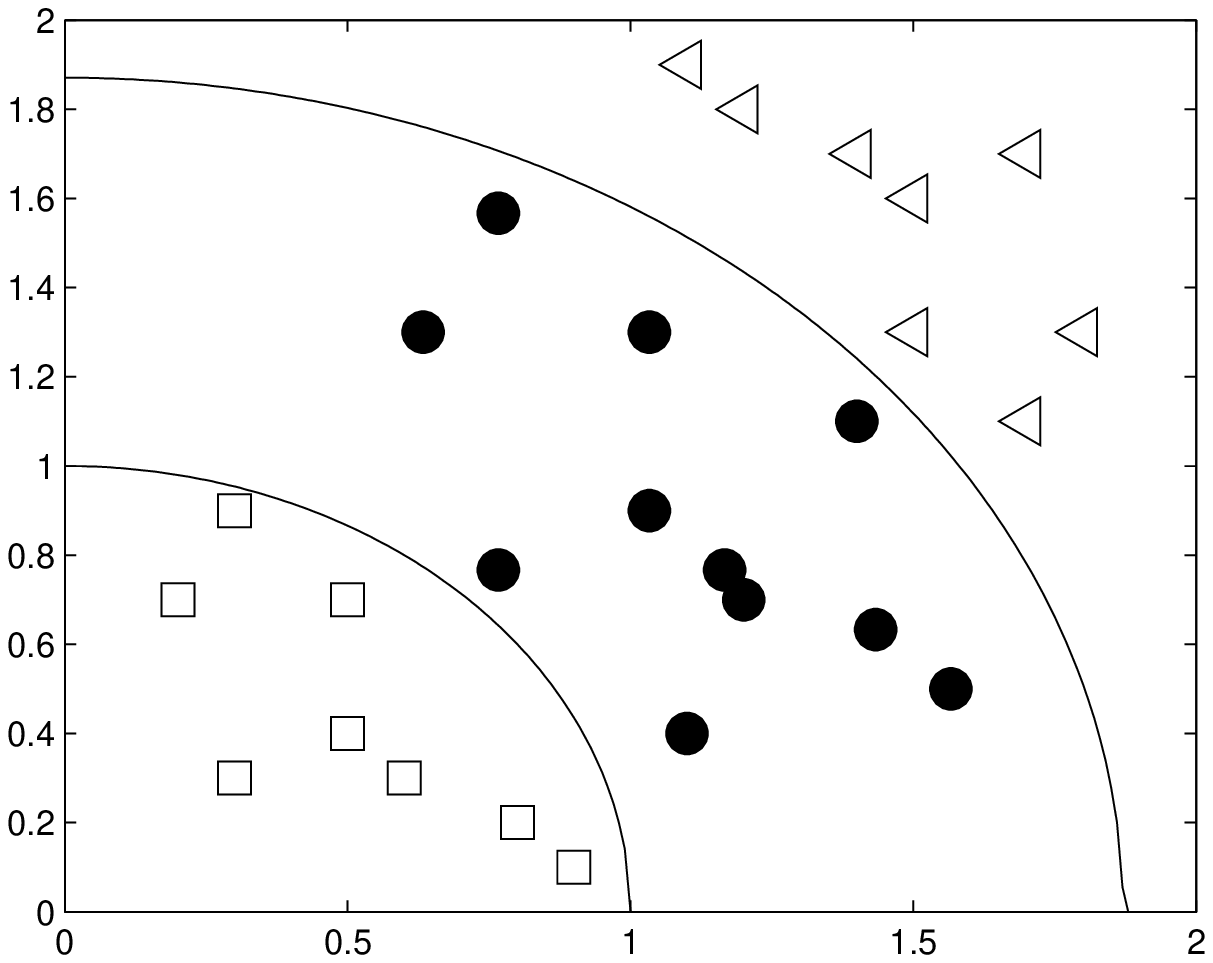}}
\caption{Nonlinear data extension model in the toy example.}
\label{fig:proposedNONLINEARModel}
\end{center}
\end{figure}

Extending the feature space and modifying to a binary problem, as dictated by the data replication method, we can search
for a partially linear (nonlinear in the original variables but linear in the introduced variables) boundary $\overline{G}(\overline{\textbf{x}}) = G(\textbf{x}) + \underline{\textbf{w}}^t\textbf{e}_i = 0$, with $\underline{\textbf{w}} \in \IR ^{K-2}$, and 
$\overline{\textbf{x}} = \left[\begin{smallmatrix}\textbf{x} \\ \textbf{e}_i \end{smallmatrix}\right]$.
The intersection of the constructed high-dimensional boundary with each of the subspace replicas provides the desired $(K-1)$ boundaries.
This approach is plotted in figure \ref{fig:proposedNONLINEARModel} for the toy example.\footnote{Although a partial linear function $\overline{G}(\overline{\textbf{x}})$ is the simplest to 
provide noncrossing boundaries in the original space (level curves of some function $G(\textbf{x})$), 
it is by no means the only type of function to provide them.}

\section{A general framework}

As presented so far the data replication method allows only to search for parallel hyperplanes (\emph{level curves} in the nonlinear case) boundaries. That is, a single direction is specified for all boundaries. 
In the quest for an extension allowing more loosely coupled boundaries, let us start by reviewing a method for ordinal data already presented in the literature.

\subsection{The method of Frank and Hall}
Frank and Hall \cite{Frank1999} introduced a simple algorithm that enables standard classification algorithms to exploit the ordering information in ordinal prediction problems. First, the data is transformed from a $K$-class ordinal problem to $K-1$ binary class problems. 
Training of the $i$-th classifier is performed by converting the ordinal dataset with classes ${\cal C}_1, \cdots, {\cal C}_K$ into a binary dataset, discriminating ${\cal C}_1, \cdots, {\cal C}_i$ against ${\cal C}_{i+1}, \cdots, {\cal C}_K$; in fact it represents the test ${\cal C}_{\textbf{x}} > i$.
To predict the class value of an unseen instance, the $K-1$ binary outputs are combined to produce a single estimation. 
Any binary classifier can be used as the building block of this scheme.

Observe that, under our approach, the $i$-th boundary is also discriminating ${\cal C}_1, \cdots, {\cal C}_i$ against ${\cal C}_{i+1}, \cdots, {\cal C}_K$; the major difference lies in the \emph{independence} of the boundaries found with Frank and Hall method.

\subsection{A parameterized family of classifiers}
Up to now, when replicating the original dataset, the original $p$ variables were the first $p$ variables of the $p+K-2$ variables of the new dataset, for each subspace replica, as seen in \eqref{generalExt}.

Returning to the toy example, assume that the replication was done {\bf not} according to \eqref{toyReplication} but instead using the following rule:

\begin{equation}
\label{toyGeneralReplication}
\left[\begin{smallmatrix}\textbf{x}_i^{(1)}\\ \textbf{0}_2 \\ 0\end{smallmatrix}\right], 
\left[\begin{smallmatrix}\textbf{0}_2 \\ \textbf{x}_i^{(1)}\\ h\end{smallmatrix}\right], 
\left[\begin{smallmatrix}\textbf{0}_2 \\ \textbf{x}_i^{(2)}\\ h\end{smallmatrix}\right] \in \overline{{\cal C}}_1 \quad
\left[\begin{smallmatrix}\textbf{x}_i^{(2)} \\ \textbf{0}_2 \\ 0\end{smallmatrix}\right], 
\left[\begin{smallmatrix}\textbf{x}_i^{(3)} \\ \textbf{0}_2 \\ 0\end{smallmatrix}\right], 
\left[\begin{smallmatrix}\textbf{0}_2 \\ \textbf{x}_i^{(3)}\\ h\end{smallmatrix}\right] \in \overline{{\cal C}}_2 
\end{equation}

where $\textbf{0}_2$ is the sequence of $2$ zeros.
Intuitively, by misaligning variables involved in the determination of different boundaries (variables in different subspaces), we are decoupling those same boundaries.
 
Proceeding this way, boundaries can be designed almost independently (more on this later, when mapping to SVMs).
In the linear case we have now four parameters to estimate, the same as for two independent lines in $\bbbr^2$. Intuitively, this new rule to replicate the data allows the estimation of the direction of each boundary \emph{essentially} independently.  

The general formulation in \eqref{generalExt} becomes

\begin{equation}
\label{moreGeneralExt}
\begin{array}{ll} 
\left[\begin{smallmatrix}\textbf{x}_i^{(k)}\\ \textbf{0}_{p(K-2)} \\ \textbf{e}_0\end{smallmatrix}\right] &\in 
\begin{cases} 
\overline{{\cal C}}_1 \quad k=1\\
\overline{{\cal C}}_2 \quad k=2, \cdots, \min(K, 1+s)\\
\end{cases}\\
&\vdots\\
\left[\begin{smallmatrix}\textbf{0}_{p(q-1)} \\ \textbf{x}_i^{(k)}\\ \textbf{0}_{p(K-q-1)} \\ \textbf{e}_{q-1}\end{smallmatrix}\right] &\in 
\begin{cases} 
\overline{{\cal C}}_1 \quad k = \max(1, q-s+1), \cdots, q\\
\overline{{\cal C}}_2 \quad k = q+1, \cdots, \min(K, q+s)\\
\end{cases}\\
&\vdots\\
\left[\begin{smallmatrix}\textbf{0}_{p(K-2)} \\ \textbf{x}_i^{(k)}\\ \textbf{e}_{K-2}\end{smallmatrix}\right] &\in 
\begin{cases} 
\overline{{\cal C}}_1 \quad k= \max(1,K-1-s+1), \cdots, K-1\\
\overline{{\cal C}}_2 \quad k=K\\
\end{cases}
\end{array}
\end{equation}

where $\textbf{0}_l$ is the sequence of $l$ zeros, $l\in\IN$.

While the linear basic data replication method requires the estimation of $ (p-1)+(K-1)$ parameters,  the new rule necessitates of $(p-1)(K-1)+(K-1)$, the same as the Frank and Hall approach; this corresponds to the number of free parameters in $(K-1)$ independent $p$-dimensional hyperplanes.

While this does not aim at being a practical alternative to Frank's method, it does paves the way for intermediate solutions, filling the gap between the totally coupled and totally independent boundaries.

To constraint only the first $j$ variables of the $p$ initial variables to have the same direction in all boundaries, while leaving the $(p-j)$ final variables unconstrained, we propose to extend the data according to 

\begin{equation}
\label{TotalGeneralExt}
\begin{array}{ll} 
\left[\begin{smallmatrix}\textbf{x}_i^{(k)}(1:j)\\ \textbf{x}_i^{(k)}(j+1:p)\\ \textbf{0}_{(p-j)(K-2)} \\ \textbf{e}_0\end{smallmatrix}\right] &\in 
\begin{cases} 
\overline{{\cal C}}_1 \quad k=1\\
\overline{{\cal C}}_2 \quad k=2, \cdots, \min(K, 1+s)\\
\end{cases}\\
&\vdots\\
\left[\begin{smallmatrix}\textbf{x}_i^{(k)}(1:j)\\ \textbf{0}_{(p-j)(q-1)} \\ \textbf{x}_i^{(k)}(j+1:p)\\ \textbf{0}_{(p-j)(K-q-1)} \\ \textbf{e}_{q-1}\end{smallmatrix}\right] &\in 
\begin{cases} 
\overline{{\cal C}}_1 \quad k = \max(1, q-s+1), \cdots, q\\
\overline{{\cal C}}_2 \quad k = q+1, \cdots, \min(K, q+s)\\
\end{cases}\\
&\vdots\\
\left[\begin{smallmatrix}\textbf{x}_i^{(k)}(1:j)\\ \textbf{0}_{(p-j)(K-2)} \\ \textbf{x}_i^{(k)}(j+1:p)\\ \textbf{e}_{K-2}\end{smallmatrix}\right] &\in 
\begin{cases} 
\overline{{\cal C}}_1 \quad k= \max(1,K-1-s+1), \cdots, K-1\\
\overline{{\cal C}}_2 \quad k=K\\
\end{cases}
\end{array}
\end{equation}

With this rule $[p-1-(j-1)](K-1) + (K-1) + j-1$,  $j \in\ \{1,\cdots,p\}$, parameters are to be estimated.

This general formulation of the data replication method allows the enforcement of only the amount of knowledge (constraints) that is effectively known \emph{a priori}, building the right amount of parsimony into the model.

\chapter{Mapping the data replication method to learning algorithms}
\setcounter{footnote}{1}
\label{Mapping}
Suppose that examples in a classification problem come from one of $K$ classes, numbered from $1$ to $K$, 
corresponding to their natural order if one exists, and arbitrarily otherwise.
The learning task is to select a prediction function $f(\textbf{x})$ from a family of possible functions 
that minimizes the expected {\em loss}.

In the absence of reliable information on relative costs, a natural approach for unordered classes is to treat every
misclassification as equally likely. This translates to adopting the non-metric indicator function
$l_{0-1}(f(\textbf{x}), y) = 0$ if $f(\textbf{x}) = y$ and $l_{0-1}(f(\textbf{x}), y) = 1$ if $f(\textbf{x}) \neq y$,
where $f(\textbf{x})$ and $y$ are the predicted and true classes, respectively.
Measuring the performance of a classifier using the $l_{0-1}$ loss function is equivalent to simply considering
the misclassification error rate.
However, for ordered classes, losses that increase with the absolute difference between the class
numbers are more natural choices in the absence of better information \cite{Mathieson1995A}. This loss should be naturally incorporated during the training period of the learning algorithm.

A risk functional that takes into account the ordering of the classes can be defined as
\begin{equation}
\label{eq:riskFunct}
R(f) = \textbf{E}\left[l^s\left(f(\textbf{x}^{(k)}),k\right)\right]
\end{equation}
with
$$
l^s\left(f(\textbf{x}^{(k)}),k\right) = \min\left(|f(\textbf{x}^{(k)})-k|, s\right)
$$

The empirical risk is the average of the number of mistakes,
where the magnitude of a mistake is related to the total ordering: 
$R_{emp}^s(f)=\frac{1}{\ell}\sum_{k=1}^K\sum_{i=1}^{\ell_k} l^s\left(f(\textbf{x}_i^{(k)}),k\right)$.

Arguing as \cite{Herbrich1999B}, we see that the role of parameter $s$ (bounding the loss incurred in each example) is to allow for an incorporation of a priori knowledge about the probability of the classes, conditioned by $\textbf{x}$, $P({\cal C}_k|\textbf{x})$. This can be treated as an assumption on the concentration of the probability around a ``true'' rank. Let us see how all this finds its place with the data replication method.

\section{Mapping the data replication method to SVMs}

\subsection{The minimum margin principle}
Let us formulate the problem of separating $K$ ordered classes ${\cal C}_1, \cdots, {\cal C}_K$ in the spirit of 
SVMs.
 
Starting from the generalization of the two-class separating hyperplane presented in the beginning of previous section, let us look for
$K-1$ parallel hyperplanes represented by vector $\textbf{w}\in \IR^p$ and scalars $b_1, \cdots, b_{K-1}$,
such that the feature space is divided into equally ranked regions by the decision boundaries
$\textbf{w}^t\textbf{x}+b_r, \ r = 1,\cdots,K-1$.

Going for a strategy to maximize the margin of the closest pair of classes, the goal becomes to
maximize $\min |\textbf{w}^t\textbf{x}+b_i|/||\textbf{w}||$.
Recalling that an algebraic measure of the distance of a point to the hyperplane $\textbf{w}^t\textbf{x}+b$ is given by 
$(\textbf{w}^t\textbf{x}+b) /\|\textbf{w}\|$, we can scale $\textbf{w}$ and $b_{i}$ so that the value of the minimum margin is $2/\|\textbf{w}\|$.

\begin{figure}[!ht]
\begin{center}
\includegraphics[width=0.4\linewidth]{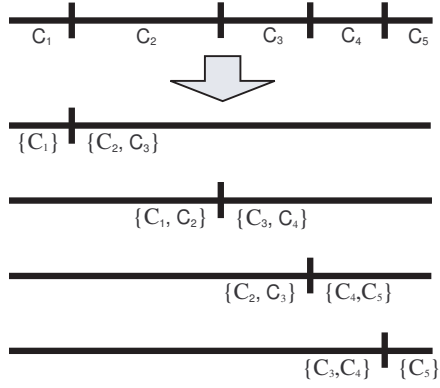}
\caption{Classes involved in the hyperplanes constraints, for $K=5$, $s=2$.}
\label{fig:ordinal2binary}
\end{center}
\end{figure}

The constraints to consider result from the $K-1$ binary classifications related to each hyperplane; 
the number of classes involved in each binary classification can be made dependent on a parameter $s$, as depicted in figure \ref{fig:ordinal2binary}. For the hyperplane $q\in\ \{1,\cdots,K-1\}$, the constraints result as 

$$
\begin{array}{lcl}
-(\textbf{w}^t\textbf{x}_i^{(k)}+ b_{q})& \geq +1 & k = \max(1,q-s+1), \cdots, q\\
+(\textbf{w}^t\textbf{x}_i^{(k)}+ b_{q})& \geq +1 & k = q+1, \cdots, \min(K, q+s)\\
\end{array}
$$

Our model can now be summarized as: 
\begin{footnotesize}
\begin{equation}
\label{eqno1}
\begin{split}
&\min_{\textbf{w}, b_i} \quad	\frac{1}{2}\textbf{w}^t\textbf{w}\\
&\begin{array}{ll}
s.t. &
\begin{array}{lll}
								-(\textbf{w}^t\textbf{x}_i^{(k)}+ b_{1})& \geq +1 & k = 1\\
								+(\textbf{w}^t\textbf{x}_i^{(k)}+ b_{1})& \geq +1 & k = 2, \cdots, \min(K, 1+s)\\
								 \vdots&&\\
								-(\textbf{w}^t\textbf{x}_i^{(k)}+ b_{q})& \geq +1 & k = \max(1,q-s+1), \cdots, q\\
								+(\textbf{w}^t\textbf{x}_i^{(k)}+ b_{q})& \geq +1 & k = q+1, \cdots, \min(K, q+s)\\
								 \vdots&&\\								
								-(\textbf{w}^t\textbf{x}_i^{(k)}+ b_{K-1})& \geq +1 & k = \max(1,K-s), \cdots, K-1\\
								+(\textbf{w}^t\textbf{x}_i^{(k)}+ b_{K-1})& \geq +1 & k = K\\
								 \xi_i^{(k)} \geq 0 &&
\end{array}\end{array}
\end{split}
\end{equation}
\end{footnotesize}
Reasoning as in the two-class SVM for the non-linearly separable dataset, the model becomes
\begin{footnotesize}
\begin{equation}
\label{eqno2}
\begin{split}
&\min_{\textbf{w}, b_i, \xi_i}	\quad \frac{1}{2}\textbf{w}^t\textbf{w} + 
C\sum_{q=1}^{K-1} \sum_{k=\max(1,q-s+1)}^{\min(K, q+s)} \sum_{i=1}^{\ell_k} \mbox{ sgn }(\xi_{i,q}^{(k)})\\
&\begin{array}{ll}
s.t. &
\begin{array}{lll}
								-(\textbf{w}^t\textbf{x}_i^{(k)}+ b_{1})& \geq +1-\xi_{i,1}^{(k)} & k = 1\\
								+(\textbf{w}^t\textbf{x}_i^{(k)}+ b_{1})& \geq +1-\xi_{i,1}^{(k)} & k = 2, \cdots, \min(K, 1+s)\\
								 \vdots&&\\
								-(\textbf{w}^t\textbf{x}_i^{(k)}+ b_{q})& \geq +1-\xi_{i,q}^{(k)} & k = \max(1,q-s+1), \cdots, q\\
								+(\textbf{w}^t\textbf{x}_i^{(k)}+ b_{q})& \geq +1-\xi_{i,q}^{(k)} & k = q+1, \cdots, \min(K, q+s)\\
								 \vdots&&\\								
								-(\textbf{w}^t\textbf{x}_i^{(k)}+ b_{K-1})& \geq +1-\xi_{i,K-1}^{(k)} & k = \max(1,K-s), \cdots, K-1\\
								+(\textbf{w}^t\textbf{x}_i^{(k)}+ b_{K-1})& \geq +1-\xi_{i,K-1}^{(k)} & k = K\\
								 \xi_{i,q}^{(k)} \geq 0 &&
\end{array}\end{array}
\end{split}
\end{equation}
\end{footnotesize}
Since each point $\textbf{x}_i^{(k)}$ is replicated $2.s$ times, it is involved in the definition of  $2.s$ boundaries (see figure \ref{fig:ordinal2binary}); consequently, it can be shown to be misclassified 
$\min(|f(\textbf{x}_i^{(k)})- k|, s) = l^s(f(\textbf{x}_i^{(k)}), k)$ times, where
$f(\textbf{x}_i^{(k)})$ is the class estimated by the model. As with the two-class example, 
$\sum_{q=1}^{K-1} \sum_{k=\max(1,q-s+1)}^{\min(K, q+s)} \sum_{i=1}^{\ell_k} \mbox{ sgn }(\xi_{i,q}^{(k)})$ is 
an upperbound of $\sum_k \sum_i l^s(f(\textbf{x}_i^{(k)}), k)$, proportional to the empirical risk.\footnote{Two parameters named $s$ have been introduced. In section \ref{sec:linearreplication} the $s$ parameter bounds the number of classes involved in the definition of each boundary, controlling this way the growth of the original dataset. The parameter $s$ introduced in equation \eqref{eq:riskFunct} bounds the loss incurred in each example. Here we see that they are the same parameter.}

Continuing the parallelism with the two-class SVM, the function to minimize simplifies to 

\begin{equation}
\label{eqno3}
\min_{\textbf{w}, b_i, \xi_i}	\quad \frac{1}{2}\textbf{w}^t\textbf{w} + 
C\sum_{q=1}^{K-1} \sum_{k=\max(1,q-s+1)}^{\min(K, q+s)} \sum_{i=1}^{\ell_k} \xi_{i,q}^{(k)}
\end{equation}
subject to the same constraints as \eqref{eqno2}.

As easily seen, the proposed formulation resembles the fixed margin strategy in \cite{Shashua2002A}. 
However, instead of using only the two closest classes in the constraints of an hyperplane, more appropriate for the loss function $l_{0-1}()$, we adopt a formulation that captures better the performance of a classifier for ordinal data.

\begin{figure}[!ht]
\begin{center}
\includegraphics[width=0.5\linewidth]{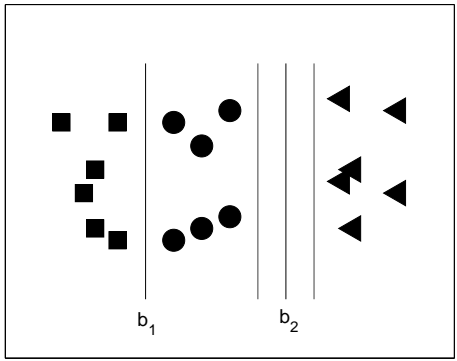}
\caption{Scalar $b_2$ is undetermined over an interval under the fixed margin strategy.}
\label{fig:nonUniqueSolution}
\end{center}
\end{figure}

Two issues were identified in the above formulation. First, this is an incompletely specified model because
the scalars $b_i$ are not well defined. In fact, although the direction of the hyperplanes $\textbf{w}$ is unique under the above formulation (proceeding as \cite{Vapnik1998} for the binary case), 
the scalars $b_1, \cdots, b_{K-1}$ are not uniquely defined, figure \ref{fig:nonUniqueSolution}.

Another issue is that, although the formulation was constructed from the two-class SVM, it is no longer solvable 
with the same algorithms. 
It would be interesting to accommodate this formulation under the two-class problem. 
Both issues are addressed by mapping the data replication method to SVMs.

\subsection{The oSVM algorithm}
In order to get a better intuition of the general result, consider first the toy example previously presented. 

\begin{figure}[!ht]
\begin{center}
\subfigure[Original dataset in $\IR$.]{
        \label{fig:oneDmodelA}
        \includegraphics[width=0.3\linewidth]{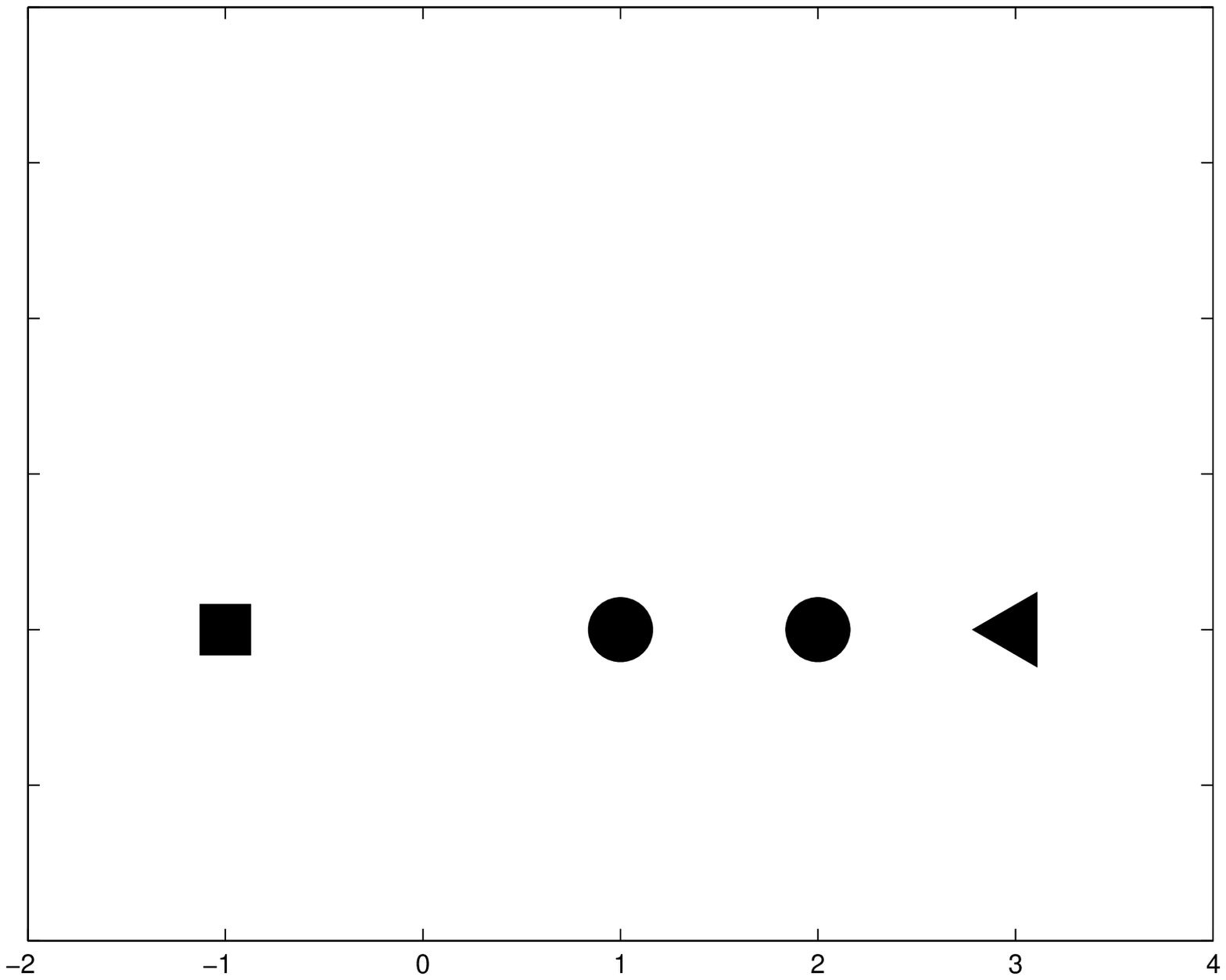}}
\subfigure[Data set in $\IR^2$, with samples replicated, $s=2$ and oSVM solution to the binary problem.]{
        \label{fig:oneDmodelC}
        \includegraphics[width=0.3\linewidth]{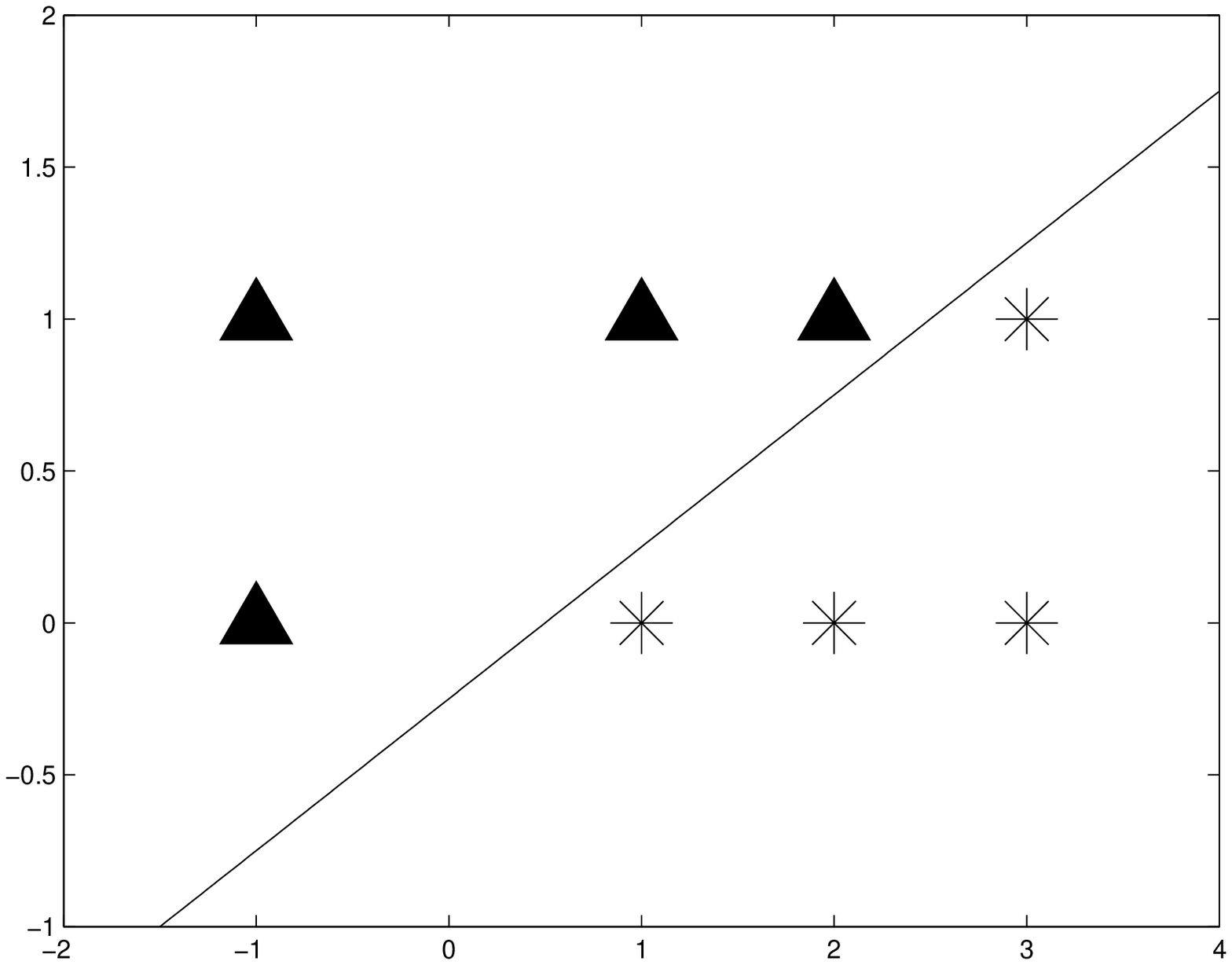}}
\subfigure[oSVM solution in the original feature space.]{
        \label{fig:oneDmodelD}
        \includegraphics[width=0.3\linewidth]{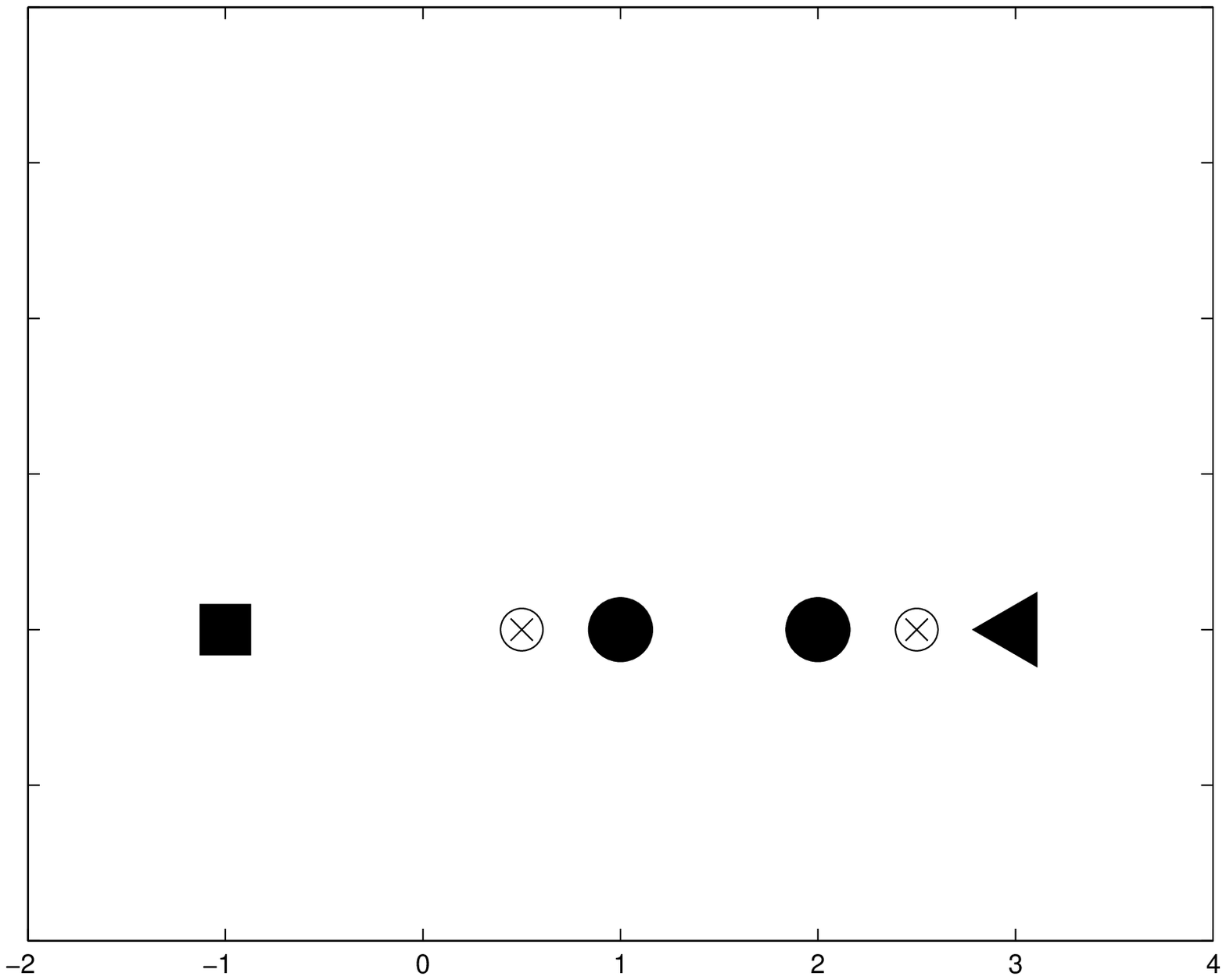}}
\caption{Effect of the regularization member in the oSVM solution.}
\label{fig:oneDmodel}
\end{center}
\end{figure}

The binary SVM formulation for the extended and binarized training set can be described as 
\big (with $\overline{\textbf{w}} = \left[\begin{smallmatrix}\textbf{w} \\ w_{3}\end{smallmatrix}\right], \ \textbf{w} \in  \IR^{2}$\big)\\
\begin{footnotesize}
\begin{equation}
\label{eq:eqno4}
\begin{array}{ll}
\min_{\overline{\textbf{w}}, b} 	& \frac{1}{2}\overline{\textbf{w}}^t\overline{\textbf{w}}  \\ \\

s.t.&\begin{array}{r}	
						  -(\overline{\textbf{w}}^t\left[\begin{smallmatrix}\textbf{x}_i^{(1)}\\ 0\end{smallmatrix}\right]+b)\geq +1  \vspace{1mm}\\
						  +(\overline{\textbf{w}}^t\left[\begin{smallmatrix}\textbf{x}_i^{(2)}\\ 0\end{smallmatrix}\right]+b)\geq +1  \vspace{1mm}\\
						  +(\overline{\textbf{w}}^t\left[\begin{smallmatrix}\textbf{x}_i^{(3)}\\ 0\end{smallmatrix}\right]+b)\geq +1  \vspace{1mm}\\
						  -(\overline{\textbf{w}}^t\left[\begin{smallmatrix}\textbf{x}_i^{(1)}\\ h\end{smallmatrix}\right]+b)\geq +1  \vspace{1mm}\\
						  -(\overline{\textbf{w}}^t\left[\begin{smallmatrix}\textbf{x}_i^{(2)}\\ h\end{smallmatrix}\right]+b)\geq +1  \vspace{1mm}\\
						  +(\overline{\textbf{w}}^t\left[\begin{smallmatrix}\textbf{x}_i^{(3)}\\ h\end{smallmatrix}\right]+b)\geq +1  \vspace{1mm}\\
\end{array}
\end{array}
\end{equation}
\end{footnotesize}
\noindent But because $\begin{cases}\overline{\textbf{w}}^t [\begin{smallmatrix}\textbf{x}_i\\ 0\end{smallmatrix}] = \textbf{w}^t\textbf{x}_i\\
		\overline{\textbf{w}}^t [\begin{smallmatrix}\textbf{x}_i\\ h\end{smallmatrix}] = \textbf{w}^t\textbf{x}_i + w_{3}h\end{cases}$, and renaming $b$ to $b_1$ and $b+w_{3}h$ to $b_2$ the formulation 		
above simplifies to
\begin{footnotesize} 
\begin{equation}
\label{eq:eqno5}
\begin{array}{ll}
\min_{\textbf{w}, b_1, b_2} 	& \frac{1}{2}\textbf{w}^t\textbf{w} + 
												\frac{1}{2}\frac{(b_2-b_1)^2}{h^2} \\ \\
s.t.&\begin{array}{l}												
					    -(\textbf{w}^t\textbf{x}_{i}^{(1)}+b_1)\geq +1  \\
					    +(\textbf{w}^t\textbf{x}_{i}^{(2)}+b_1)\geq +1  \\
						  +(\textbf{w}^t\textbf{x}_{i}^{(3)}+b_1)\geq +1  \\
						  -(\textbf{w}^t\textbf{x}_{i}^{(1)}+b_2)\geq +1  \\
						  -(\textbf{w}^t\textbf{x}_{i}^{(2)}+b_2)\geq +1  \\
						  +(\textbf{w}^t\textbf{x}_{i}^{(3)}+b_2)\geq +1 \\
\end{array}
\end{array}
\end{equation}
\end{footnotesize}
Two points are worth to mention: a) this formulation, being the result of a pure SVM method, has an unique solution \cite{Vapnik1998}; b) this formulation equals the formulation \eqref{eqno3} for ordinal data previously introduced,
with $K=3$, $s=K-1=2$, and a slightly modified objective function by the introduction of a regularization member, proportional to the distance between the hyperplanes.
The oSVM solution is the one that simultaneously minimizes the distance between boundaries and maximizes the minimum of the margins -- figure \ref{fig:oneDmodel}.
The $h$ parameter controls the tradeoff between the objectives
of maximizing the margin of separation and minimizing
the distance between the hyperplanes.

To reiterate, the data replication method enabled us to formulate the classification of ordinal data as a standard SVM problem, removing the ambiguity in the solution by the introduction of a regularization term in the objective function.

With the material on how to construct a set of optimal hyperplanes for the toy example, we are now in a position to formally describe the construction of a support vector machine for ordinal classification.

Consider a general extended dataset, as defined in \eqref{generalExt}.
After the simplifications and change of variables suggested in the toy example,
the binary SVM formulation for this extended dataset yields 

\begin{footnotesize}
\begin{equation}
\label{eqno5}
\min_{\textbf{w}, b_i, \xi_i}	\quad \frac{1}{2}\textbf{w}^t\textbf{w} + \frac{1}{h^2}\sum_{i=2}^{K-1} \frac{(b_i-b_1)^2}{2}+ 
C\sum_{q=1}^{K-1} \sum_{k=\max(1,q-s+1)}^{\min(K, q+s)} \sum_{i=1}^{\ell_k} \xi_{i,q}^{(k)}\\ 
\end{equation}
\end{footnotesize}
with the same set of constraints as \eqref{eqno2}.

This formulation for the high-dimensional dataset matches the proposed formulation for ordinal data up to an additional regularization member in the objective function. This additional member is responsible for the unique determination of the biases.\footnote{
Different regulation members could be obtained by different extensions of the dataset. For example, if $\textbf{e}_q$ had been defined as the sequence $h, \cdots, h, 0, \cdots, 0$, with $q$ $h$'s and $(K-2-q)$ $0$'s, the regularization member would be $\frac{1}{2}\sum_{i=2}^{i=K-1}\frac{(b_i-b_{i-1})^2}{2}$.}


It is important to stress that the complexity of the SVM model does not depend on the dimensionality of the data.  
So, the only increase in the complexity of the problem is due to the duplication of the data 
(more generally, for a $K$-class problem, the dataset is increased at most {\small$(K-1)$} times).
As such, it compares favourably with the formulation in \cite{Herbrich1999A}, which squares the dataset.

\subsubsection*{Nonlinear boundaries}
As explained before, the search for nonlinear level curves can be pursued in the extended feature space by searching for a partially linear function $\overline{G}(\overline{\textbf{x}}) = G(\textbf{x}) + \underline{\textbf{w}}^t\textbf{e}_i$.
Since nonlinear boundaries are handled in the SVM context making use of the well known kernel trick, a specified kernel
$K(\textbf{x}_i, \textbf{x}_j)$ in the original feature space can be easily modified to 
$\overline{K}(\overline{\textbf{x}}_i, \overline{\textbf{x}}_j) = K(\textbf{x}_i, \textbf{x}_j) + \textbf{e}_{\textbf{x}_i}^t \textbf{e}_{\textbf{x}_j} $ in the extended space.

\begin{figure}
\begin{center}
\includegraphics[width=0.5\linewidth]{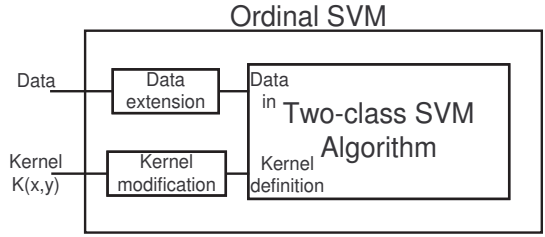}
\caption{oSVM interpretation of an ordinal multiclass problem as a two-class problem.}
\label{fig:mySVMModel}
\end{center}
\end{figure}

Summarizing, the nonlinear ordinal problem can be solved by extending the feature set and modifying the kernel function, figure \ref{fig:mySVMModel}.
As we see, the extension to nonlinear decision boundaries
follows the same reasoning as with the standard SVM \cite{Vapnik1998}.

\subsubsection*{Independent boundaries}
Considering now the setup for independent boundaries, as presented in \eqref{moreGeneralExt}, the linear, binary 
SVM formulation yields

\begin{footnotesize}
\begin{equation}
\label{eqno99}
\begin{split}
&\min_{\textbf{w}, b_i, \xi_i}	\quad \sum_{k=1}^{K-1} \frac{1}{2}\textbf{w}^t(kp-p+1:kp)\textbf{w}(kp-p+1:kp) + \frac{1}{h^2}\sum_{i=2}^{K-1} \frac{(b_i-b_1)^2}{2}+ 
C\sum_{q=1}^{K-1} \sum_{k=\max(1,q-s+1)}^{\min(K, q+s)} \sum_{i=1}^{\ell_k} \xi_{i,q}^{(k)}\\ 
&\begin{array}{ll}
s.t. &
\begin{array}{lll}
								-(\textbf{w}^t(1:p)\textbf{x}_i^{(k)}+ b_{1})& \geq +1-\xi_{i,1}^{(k)} & k = 1\\
								+(\textbf{w}^t(1:p)\textbf{x}_i^{(k)}+ b_{1})& \geq +1-\xi_{i,1}^{(k)} & k = 2, \cdots, \min(K, 1+s)\\
								 \vdots&&\\
								-(\textbf{w}^t(qp-p+1:qp)\textbf{x}_i^{(k)}+ b_{q})& \geq +1-\xi_{i,q}^{(k)} & k = \max(1,q-s+1), \cdots, q\\
								+(\textbf{w}^t(qp-p+1:qp)\textbf{x}_i^{(k)}+ b_{q})& \geq +1-\xi_{i,q}^{(k)} & k = q+1, \cdots, \min(K, q+s)\\
								 \vdots&&\\								
								-(\textbf{w}^t((K-1)p-p+1:(K-1)p)\textbf{x}_i^{(k)}+ b_{K-1})& \geq +1-\xi_{i,K-1}^{(k)} & k = \max(1,K-s), \cdots, K-1\\
								+(\textbf{w}^t((K-1)p-p+1:(K-1)p)\textbf{x}_i^{(k)}+ b_{K-1})& \geq +1-\xi_{i,K-1}^{(k)} & k = K\\
								 \xi_{i,q}^{(k)} \geq 0 &&
\end{array}\end{array}
\end{split}
\end{equation}
\end{footnotesize}

We see that if the regularization term $\frac{1}{h^2}\sum_{i=2}^{K-1} \frac{(b_i-b_1)^2}{2}$ is zero (in practice, sufficiently small), the optimization problem could be broken in $K-1$ \emph{independent} optimization problems, reverting to the procedure of Frank and Hall \cite{Frank1999}.

\section{Mapping the data replication method to NNs}

\begin{figure}
\begin{center}
\includegraphics[width=.9\linewidth]{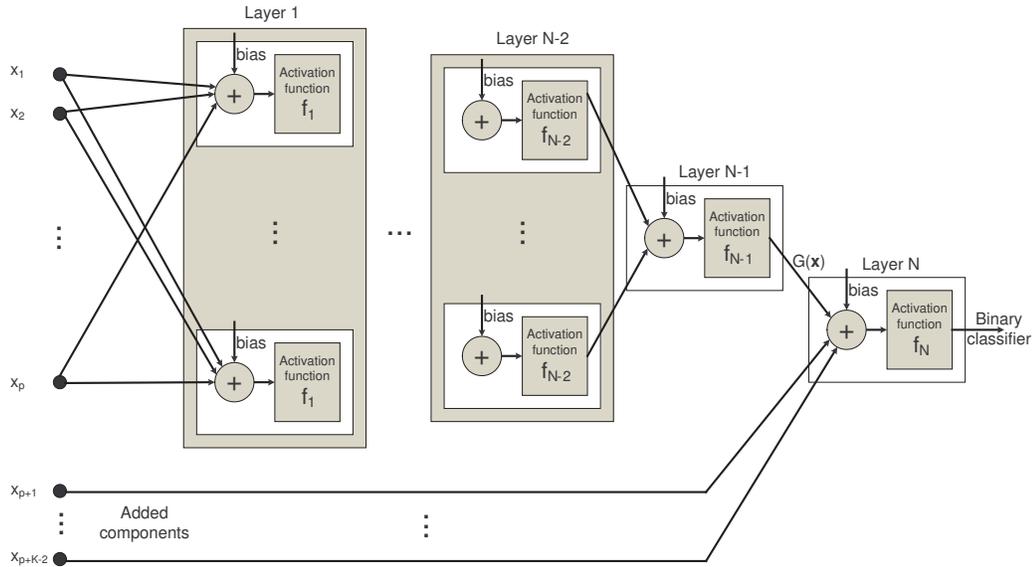}
\caption{Data replication method for neural networks (oNN).}
\label{fig:oNNmodel}
\end{center}
\end{figure}

By letting $G (\textbf{x})$ be the output of a neural network, a flexible architecture for ordinal data can be devised as represented diagrammatically in figure \ref{fig:oNNmodel}. 
$G (\textbf{x})$ is the output of a generic feedforward network (in fact, it could be any neural network, with a single output), which is then linearly combined with the added $(K-2)$ components. 

\begin{figure}
\begin{center}
\includegraphics[width=.6\linewidth]{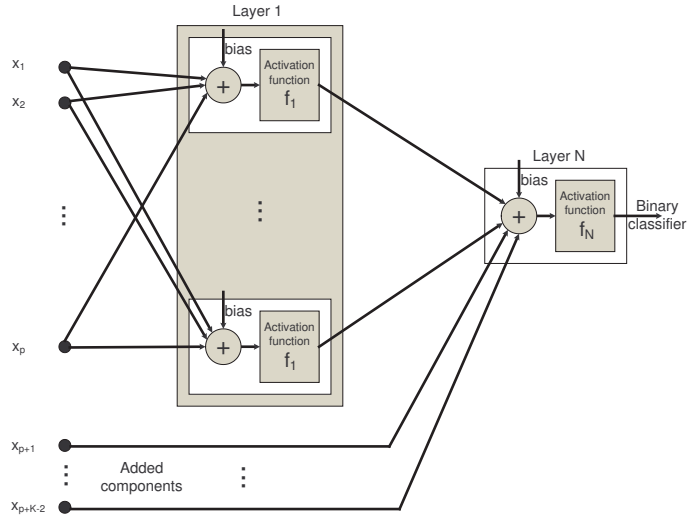}
\caption{Simplified oNN model for neural networks.}
\label{fig:oNNmodelsimp}
\end{center}
\end{figure}

For the simple case of searching for linear boundaries, the overall network simplifies to a single neuron with $p+K-2$ inputs.
A less simplified model, also used in the conducted experiments, is to consider a single hidden layer, as depicted in figure \ref{fig:oNNmodelsimp}.

Interestingly, it is possible to provide a probabilistic interpretation to this neural network model.
\subsection{Ordinal logistic regression model}
The traditional statistical approach for ordinal classification models the cumulative class probability $P_k = p(C \leq k | \textbf{x})$ by 
\begin{equation}
\label{eq:logistic}
\mbox{logit} (P_k) = \Phi_k - G(\textbf{x}) \Leftrightarrow P_k = \mbox{logsig} (\Phi_k - G(\textbf{x})), \quad k=1,\cdots, K-1
\end{equation}
Remember that
$\mbox{logit}(y) = \ln \frac{y}{1-y}$, $\mbox{logsig(y)} = \frac{1}{1+e^{-y}}$ and $\mbox{logsig}(\mbox{logit}(y)) = y$.

For the linear version (\cite{McCullagh1980,McCullagh1989}) we take $G(\textbf{x}) = \textbf{w}^t\textbf{x}$. Mathieson \cite{Mathieson1995A} presents a nonlinear version by letting $G(\textbf{x})$ be the output of a neural network. 
However other setups can be devised.
Start by observing that in \eqref{eq:logistic} we can always assume $\Phi_1 = 0$ by incorporating an appropriate additive constant in $G(\textbf{x})$. 
We are left with the estimation of $G(\textbf{x})$ and $(K-2)$ cut points.
By fixing $f_N () = \mbox{logsig}()$ as the activation function in the output layer of our oNN network, we can train the network to predict the values $P_k (\textbf{x})$, when fed with $\overline{\textbf{x}} = \left[\begin{smallmatrix}\textbf{x} \\ \textbf{e}_{k-1}\end{smallmatrix}\right]$, $k=1,\cdots,K-1$ .
By setting $\overline{{\cal C}}_1 = 1$ and $\overline{{\cal C}}_2 = 0$ we see that the extended dataset as defined in \eqref{generalExt} can be used to train the oNN network. 
The predicted cut points are simply the weights of the connection of the added $K-2$ components, scaled by $h$.

\begin{figure}
\begin{center}
\includegraphics[width=.43\linewidth]{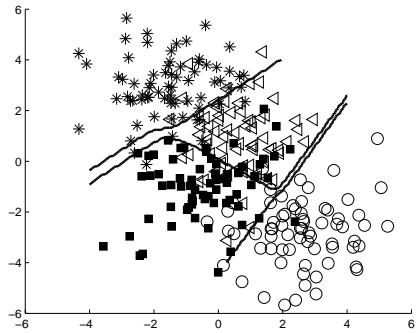}
\caption{Decision boundaries for the oNN with 3 units in the hidden layer, for a synthetic dataset from Mathieson \cite{Mathieson1995A}.
${\cal C}_1 = \circ$, ${\cal C}_2 = \blacksquare$, ${\cal C}_3 = \triangleleft$, ${\cal C}_4 = *$}
\label{fig:oNNmodelLogistic}
\end{center}
\end{figure}

Illustrating this model with the synthetic dataset from Mathieson \cite{Mathieson1995A}, we attained the decision boundaries depicted in figure \ref{fig:oNNmodelLogistic}.


\chapter[The unimodal method for NNs]{The unimodal method for NNs\footnotemark[4]}\footnotetext[4]{The text and idea presented in this section is thanks to Prof. Joaquim F. Pinto da Costa. 
Some portions of this chapter appeared in \cite{JFPCostaECML2005}.}
\setcounter{footnote}{1}
\label{chap:unimodal}
Given a new query point $\textbf{x}$, Bayes decision theory suggests to classify $\textbf{x}$ in the class which maximizes the {\it a posteriori} probability $P({\cal C}_k|\textbf{x})$. To do so, one usually has to estimate these probabilities, either implicitly or explicitly. Suppose for instance that we have 7 classes and, for a given point $\textbf{x}_0$, the highest probability is $P({\cal C}_5|\textbf{x}_0)$; we then assign class ${\cal C}_5$ to the given point. If there is not an order relation between the classes, it is perfectly natural that the second highest {\it a posteriori} probability is, for instance, $P({\cal C}_2|\textbf{x})$. However, if the classes are ordered, ${\cal C}_1 < {\cal C}_2 <, \ldots, <{\cal C}_7$, classes ${\cal C}_4$ and ${\cal C}_6$ are closer to class ${\cal C}_5$ and therefore the second and third highest  {\it a posteriori} probabilities should be attained in these classes.  This argument extends easily to the  classes, ${\cal C}_3$ and ${\cal C}_7$, and so on. 
This is the main idea behind the method proposed here, which is now detailed.

\begin{figure}
\begin{center}
\includegraphics[width=.95\linewidth]{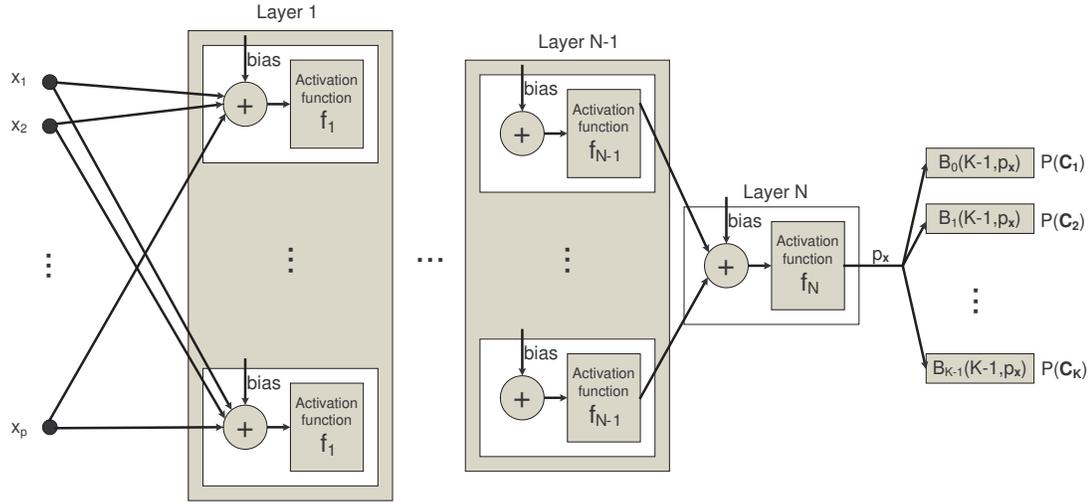}
\caption{unimodal neural network architecture.}
\label{fig:uNNarchitecture}
\end{center}
\end{figure}

Our method assumes that in a supervised classification problem with ordered classes, the random variable class associated with a given query $\textbf{x}$ should be unimodal. That is to say that if we plot the 
{\it a posteriori} probabilities $P({\cal C}_k|\textbf{x})$, from the first ${\cal C}_1$ to the last ${\cal C}_K$, there should be only one mode in this graphic. Here, we apply this idea in the context of neural networks. Usually in neural networks, the output layer has as many units as there are classes, $K$. We will use the same order for these units and the classes. In order to force the output values (which represent the {\it a posteriori} probabilities) to have just one mode, we will use a parametric model for these output units. This model consists in assuming that the output values come from a binomial distribution, $B (K-1,p)$. This distribution  is unimodal in most cases and when it has two modes, these are for contiguous values, which makes sense in our case, since we can have exactly the same probability for two classes. This binomial distribution takes integer values in the set $\{ 0,1,\ldots,K-1\}$; value $0$ corresponds to class ${\cal C}_1$, value 1 to class ${\cal C}_2$ and so on until value $K-1$ to class ${\cal C}_{K}$. As $K$ is known, the only parameter left to be estimated from this model is the probability $p$. We will therefore use a different architecture for the neural network; that is, the output layer will have just one output unit, corresponding to the value of $p$ -- figure \ref{fig:uNNarchitecture}. For a given query $\textbf{x}$, the output of the network will be a single numerical value in the range [0,1], which we call $p_\textbf{x}$. Then, the probabilities $P({\cal C}_k|\textbf{x})$ are calculated from the binomial model: $$P({\cal C}_k|\textbf{x})\;=\; \frac{(K-1)!p_\textbf{x}^{k-1}(1-p_\textbf{x})^{K-k}}{(k-1)!(K-k)!},\,\, k=1,2,\ldots,K$$

In fact these probabilities can be calculated recursively, to save computing time:  $$\frac{P({\cal C}_k|\textbf{x})}{P({\cal C}_{k-1}|\textbf{x})}\; = \; 
\frac{p_\textbf{x}(K-k+1)}{(k-1)(1-p_\textbf{x})},$$ and so 
$$P({\cal C}_k|\textbf{x})\; = \; P({\cal C}_{k-1}|\textbf{x})
\frac{p_\textbf{x}(K-k+1)}{(k-1)(1-p_\textbf{x})}.$$
 We start with $P({\cal C}_1|\textbf{x})\;=\; (1-p_\textbf{x})^{K-1}$ and compute the other probabilities, $P({\cal C}_k|\textbf{x}),\;\;k=2,3,\ldots,K,$ using the above formula. 
 
When the training case $\textbf{x}$ is presented, the error is defined as 
\begin{equation}
\mbox{error } = \sum_{k=1}^K \left|P({\cal C}_k|\textbf{x}) - \delta(k-{\cal C}_\textbf{x})\right|^2
\end{equation}
where $\delta (n) = \begin{cases}1 \mbox{ if } n=0 \\ 0 \mbox{ otherwise} \end{cases}$ and ${\cal C}_\textbf{x}$ the true class of $\textbf{x}$.
The network is trained to minimize the average value over all training cases of such error.
Finally, in the test phase, we choose the class $k$ which maximizes the probability $P({\cal C}_k)$.
As trivially confirmed, that simplifies to the rounding of $1+(K-1)p_\textbf{x}$ to the nearest integer, where $p_\textbf{x}$ is the network output.

\section{Connection with regression models}
Consider the following equivalences:

$
\min_{p_\textbf{x}} \sum_{k=1}^K \left|P({\cal C}_k|\textbf{x}) - \delta(k-{\cal C}_\textbf{x})\right| \Leftrightarrow
\min_{p_\textbf{x}} 1 - P({\cal C}_\textbf{x}|\textbf{x}) + \sum_{k \not = {\cal C}_\textbf{x}}  P({\cal C}_k|\textbf{x}) \Leftrightarrow \\
\min_{p_\textbf{x}}  2 - 2P({\cal C}_\textbf{x}|\textbf{x})\Leftrightarrow 
\max_{p_\textbf{x}}  P({\cal C}_\textbf{x}|\textbf{x})
$

Let $p_\textbf{x}^{\mbox{\scriptsize opt}}$ the parameter that maximizes $P({\cal C}_\textbf{x}|\textbf{x})$. For the binomial case $p_\textbf{x}^{\mbox{\scriptsize opt}} = \frac{{\cal C}_\textbf{x} - 1}{K-1}$.
Then 
\begin{equation}\max_{p_\textbf{x}}  P({\cal C}_\textbf{x}|\textbf{x})\end{equation}
and  
\begin{equation}\label{eq:binomialRegression} \min_{p_\textbf{x}} |p_\textbf{x} - p_\textbf{x}^{\mbox{\scriptsize opt}}|\end{equation}
both attain the global optimal value at the same $p_\textbf{x}$ value. 
That is to say that the training of the network could be also performed by minimizing the error of the network output to the optimal parameter: a simple case of regression of the parameter of the binomial distribution. 

Note that both approaches are not mathematically equivalent. Although they share the same global optimum, the error surface is different and is natural that practical optimization algorithms stop at different values, maybe trapped at some local optimum value. 
Another way of looking to the problem is to say that both are a regression of the parameter $p_\textbf{x}$, using different error measures.
The advantage of minimizing directly $\min_{p_\textbf{x}} |p_\textbf{x} - \frac{{\cal C}_\textbf{x} - 1}{K-1}|$, or the squared version of it, is that it fits directly in existing software packages.
However, both impose a unimodal distribution of the output probabilities.

As the above formulation suggests, the adjustment of any probability distribution, dependent on a single parameter resumes to a regression of that parameter against its optimal value.
This approach is then part of a larger set of techniques to estimate by regression any ordered \emph{scores} $s_1 \leq \cdots \leq s_K$ -- the simplest case would be the set of integers $1, \cdots, K$. \cite{Mathieson1995A, Moody1995, Anurag2001}

Using a neural network with not one but two outputs, it is natural to extend the former reasoning to unimodal distribution with two parameters, as a greater flexibility should bring a better fitting to the data. The training could be performed directly with some of the regression errors discussed above and the test phase would be just the selection of the mode class dictated by the network output.

\chapter{Experimental Results}
\setcounter{footnote}{1}
\label{chap:experimental}
Next we present experimental results for several models based on SVMs and NNs, when applied to several datasets, ranging from synthetic datasets, real ordinal data, to quantized data from regression problems.

\section{SVM based algorithms}
We compare the following algorithms:

\begin{itemize}
 
\item A conventional multiclass SVM formulation (cSVM), based on the one-against-one decomposition. 
The one-against-one decomposition transforms the multiclass problem into a series of $K(K - 1)/2$ binary subtasks that can be trained by a binary SVM.
Classification is carried out by a voting scheme.

\item Pairwise SVM (pSVM): Frank and Hall \cite{Frank1999} introduced a simple algorithm that enables standard classification algorithms to exploit the ordering information in ordinal prediction problems. First, the data is transformed from a $K$-class ordinal problem to $K-1$ binary class problems. To predict the class value of an unseen instance the probabilities of the $K$ original classes are estimated using the outputs from the $K-1$ binary classifiers.

\item Herbrich \cite{Herbrich1999A} model (hSVM), based on the correspondence of the ordinal regression task and the task of learning a preference relation on pairs of objects. 
A function loss was defined on pairs of objects and the classification task formulated in this space. 
The size of the new training set, derived from an $\ell$-sized training set, can be as high as $\ell^2$.
Only the direction $\textbf{w}$ was computed directly from this model. Scalars $b_i$ were obtained in a second step, performing a 1-dimensional SVM.
Due to limitations of the implementation of this method and its excessively long training time, some results are not available (NA).

\item Proposed ordinal method (oSVM), based on the data extension technique, as previously introduced.

\end{itemize}

Experiments were carried out in Matlab 7.0 (R14), using the Support Vector Machine toolbox, version 2.51, by Anton Schwaighofer.
This toolbox was used to construct the oSVM classifier, the Herbrich \cite{Herbrich1999A} model and the pairwise SVM.
It was also used the STPRtool, version 2.01, for the implementation of the generic multiclass SVM.
The $C$ and $h$ parameters were experimentally tuned for the best performance.


\section{Neural network based algorithms}
We compare the following algorithms:

\begin{itemize} 
\item Conventional neural network (cNN). To test the hypothesis that methods specifically targeted for ordinal data improve the performance of a standard classifier, 
we tested a conventional feed forward network, fully connected, with a single hidden layer, trained with the traditional least square approach
and with the special activation function \emph{softmax}. For each case study, the result presented is the best of the two configurations.

\item Pairwise NN (pNN): mapping in neural networks the strategy of \cite{Frank1999} mentioned above for pSVM.

\item Costa \cite{Costa1996}, following a probabilistic approach, proposes a neural network architecture (iNN) that exploits the ordinal nature of the data,  
by defining the classification task on a suitable space through a ``partitive approach''. It is proposed a feedforward neural network with $K-1$ outputs to
solve a $K$-class ordinal problem. The probabilistic meaning assigned to the network outputs is exploited to rank the elements of the dataset. 

\item Proposed unimodal model (uNN).
Several variants of the unimodal model were gauged, ranging from one-parameter distributions, such as the binomial and the poison, to two-parameter distributions, such as the hypergeometric and the gaussian distribution. Other ideas such as modifying a conventional neural network to penalise multimodal outputs were also considered. However, models whose optimization did not fit directly under a standard implementation of the backpropagation algorithm were optimized with generic optimization functions available in Matlab. Presumably due to that fact, the best results were obtained when performing direct regression of the binomial parameter, as in \eqref{eq:binomialRegression}, for which we present the results.

\item Proposed ordinal method (oNN), based on the data extension technique, as previously introduced.

\end{itemize}

Experiments were carried out in Matlab 7.0 (R14), making use of the Neural Network Toolbox.
All models were configured with a single hidden layer and trained with Levenberg-Marquardt back propagation method, over 2000 epochs.

The number of neurons in the hidden layer was experimentally tuned for the best performance.

\section{Measuring classifier performance}
Having built a classifier, the obvious question is ``how good is it?''. 
This begs the question of what we mean by good.
The obvious answer is to treat every misclassification as equally likely, adopting the misclassification error rate (MER) criterion to measure the performance of the classifier.
However, for ordered classes, losses that increase with the absolute difference between the class numbers are more natural choices in the absence of better information \cite{Mathieson1995A}. 

The mean absolute error (MAE) criterion takes into account the degree of misclassification and is thus a richer criterion than MER. The loss function corresponding to this criterion is $l(f(\textbf{x}), y) = |f(\textbf{x})- y|$.

A variant of the above MAE measure is the mean square error (MSE), where the absolute difference is replaced with the square of the difference, $l(f(\textbf{x}), y) = (f(\textbf{x})- y)^2$.

Finally, the performance of the classifiers was also assessed with the Spearman ($r_s$) and Kendall's tau-b ($\tau$) coefficients, nonparametric rank-order correlation coefficients well established in the literature \cite{NumericalRecipes}.
A proposal for yet another coefficient, $o_c$, was also implemented.\footnote{The idea for this coefficient is thanks to Prof. Joaquim F. Pinto da Costa.} 
To define $o_c$, we start with the $N$ data points $(x_i, y_i)$ and consider all $\frac{1}{2}N(N-1)$ pairs of data points. 
Following the notation in \cite{NumericalRecipes}, we call a pair \emph{concordant} if the relative ordering of the ranks of the two $x$'s is the same as the relative ordering of the ranks of the two $y$'s. 
We call a pair \emph{discordant} if the relative ordering of the ranks of the $x$'s is opposite from the relative ordering of the ranks of the two $y$'s. 
If there is a tie in either the ranks of the two $x$'s or the ranks of the two $y$'s, then we do not call the pair either concordant or discordant. If the tie is in the $x$'s, we will call the pair an ``extra $x$ pair'', $e_x$. If the tie is in the $y$'s, we will call the pair an ``extra $y$ pair'', $e_y$. If the tie is both on the $x$'s and the $y$'s, we ignore the pair.

Inspired by the work of Lerman \cite{Lerman1992A,Lerman1992B}, a simplified coefficient was conceived from a set theoretic representation of the two variables to be compared. After a straight forward mathematical manipulation, the $o_c$ coefficient can be computed as 
$$o_c = -1+2\frac{concordant}{\sqrt{concordant+discordant+e_x}\sqrt{ concordant+discordant+e_y }}$$
where the scale factor and bias are used just to set the parameter between $1$ and $-1$.

This expression shows a striking resemblance with the formula for Kendall's $\tau$:
$$\tau = \frac{concordant-discordant}{\sqrt{concordant+discordant+e_x}\sqrt{ concordant+discordant+e_y }}$$

\chapter{Results for a synthetic dataset}
\setcounter{footnote}{1}
\section{Results for neural networks methods}

In a first comparative study we generated a synthetic dataset in a similar way to Herbrich \cite{Herbrich1999A}.

\begin{figure}
\begin{center}
\subfigure[Classes' boundaries.]{
        \label{fig:syntheticA}
        \includegraphics[width=0.32\linewidth]{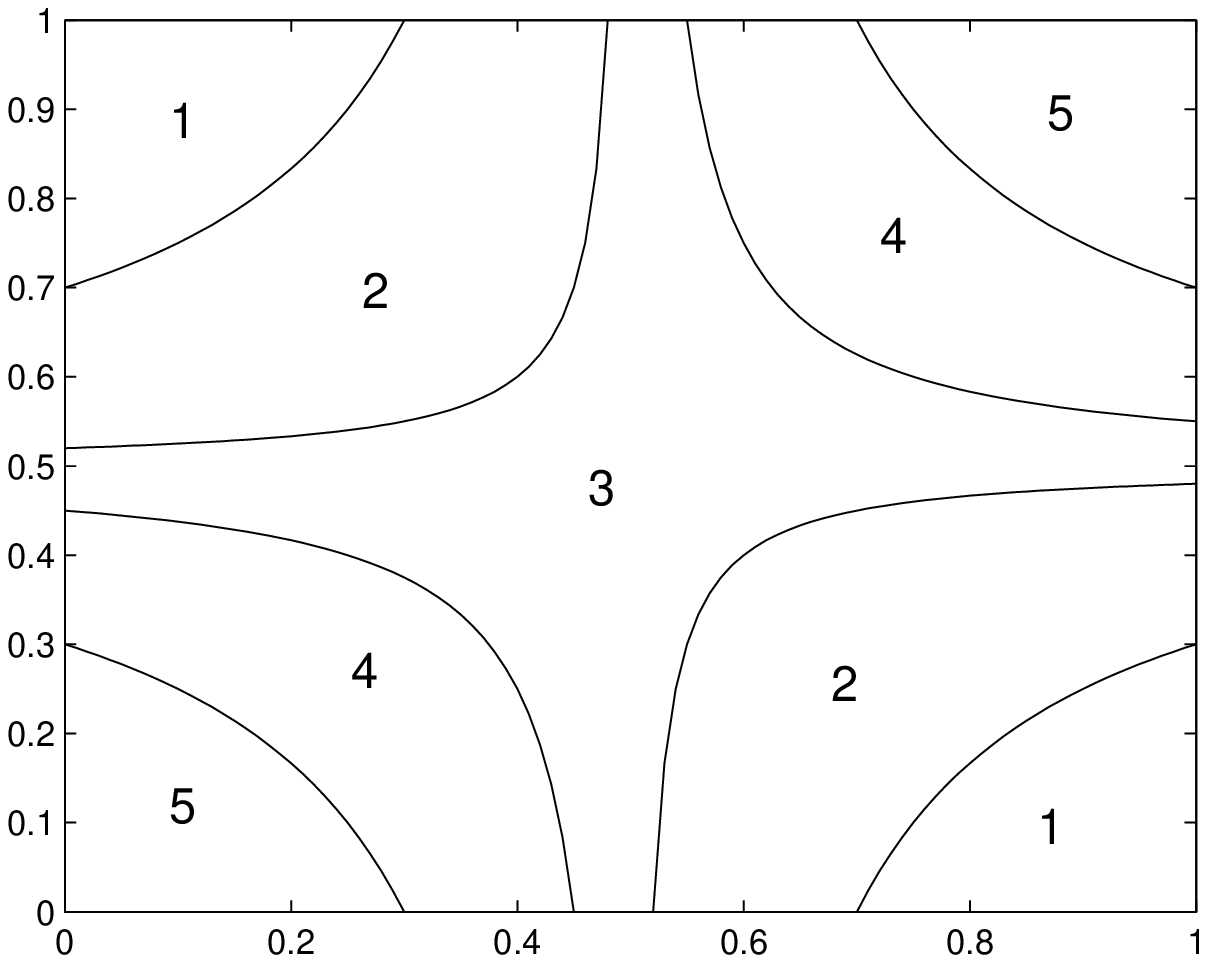}}
\subfigure[Scatter plot of the data points wrongly ranked. Number of wrong points: 14.2\%.]{
        \label{fig:syntheticB}
        \includegraphics[width=0.32\linewidth]{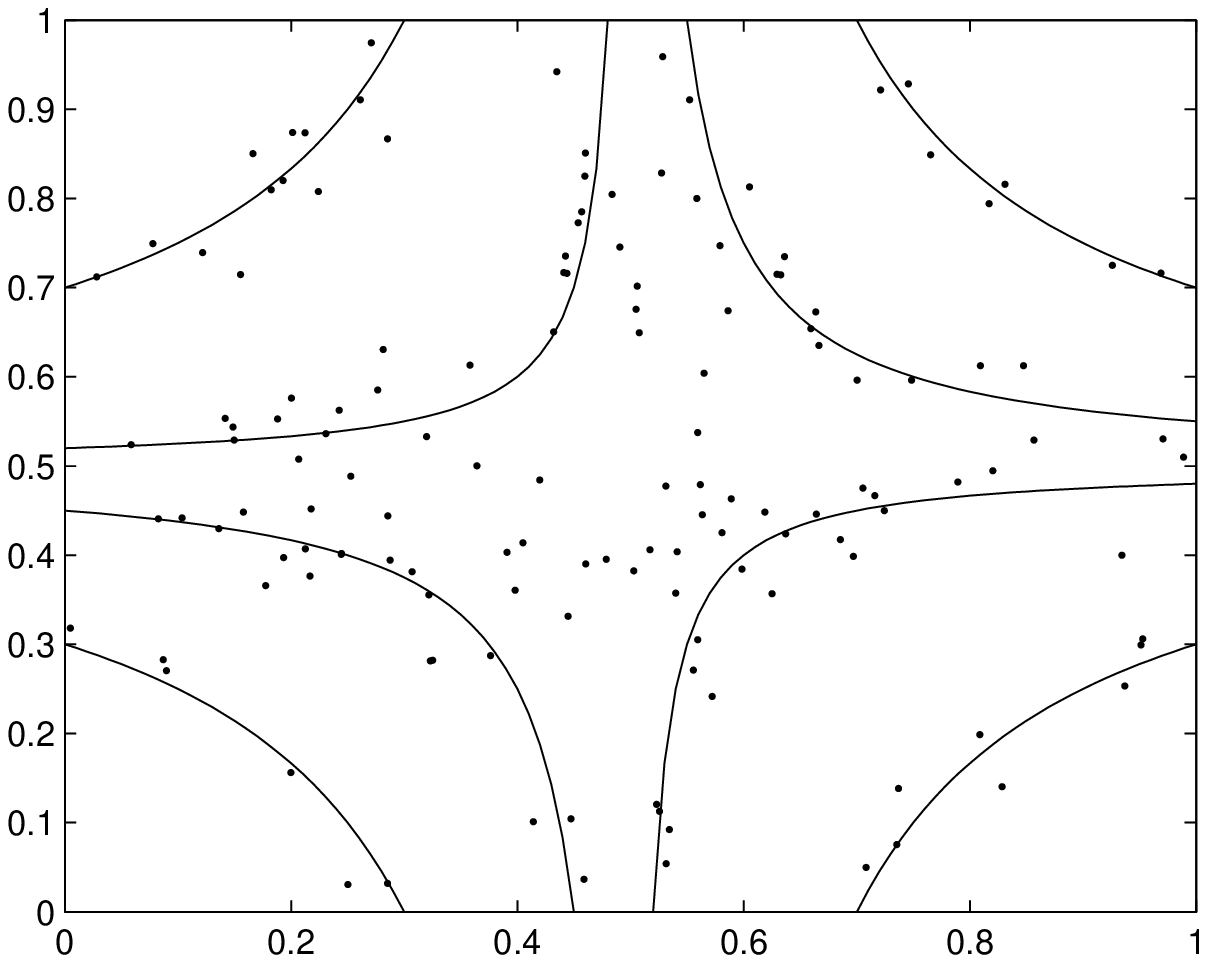}}
\subfigure[Class distribution.]{
        \label{fig:syntheticC}
        \includegraphics[width=0.32\linewidth]{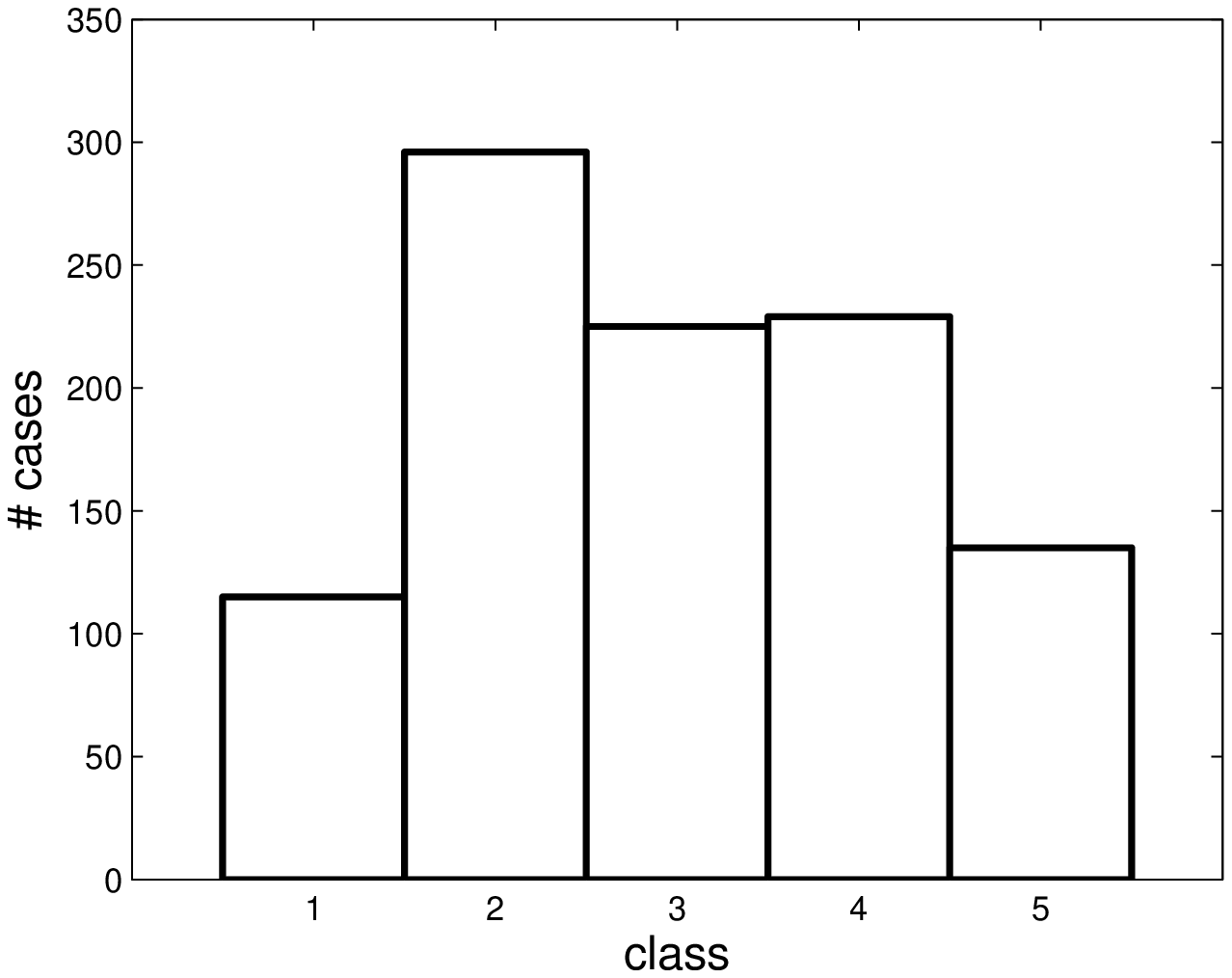}}        
\caption{Test setup for $5$ classes in $\bbbr^2$.}
\label{fig:synthetic}
\end{center}
\end{figure}

We generated $1000$ example points $\textbf{x}=[x_1 \ x_2]^t$ uniformly at random in the unit square $[0, 1]\times [0, 1] \subset \bbbr^2$.
Each point was assigned a rank $y$ from the set $\{1, 2, 3, 4, 5\}$, according to \\
\[y= \min_{r\in\{1,2,3,4,5\}}\{r: b_{r-1} < 10(x_1-0.5)(x_2-0.5) + \varepsilon\ < b_r\}\]
\[(b_0, b_1, b_2, b_3,b_4,b_5)=(-\infty, -1, -0.1, 0.25, 1, +\infty)\]
where $\varepsilon$ is a random value, normally distributed with zero mean and standard deviation $\sigma = 0.125$.
Figure \ref{fig:syntheticA} shows the five regions and figure \ref{fig:syntheticB} the points which were assigned to a different rank after the corruption with the normally distributed noise.

\begin{figure}
\begin{center}
\subfigure[MER criterion.]{
        \label{fig:synthetic0205ErrorRate}
        \includegraphics[width=0.31\linewidth]{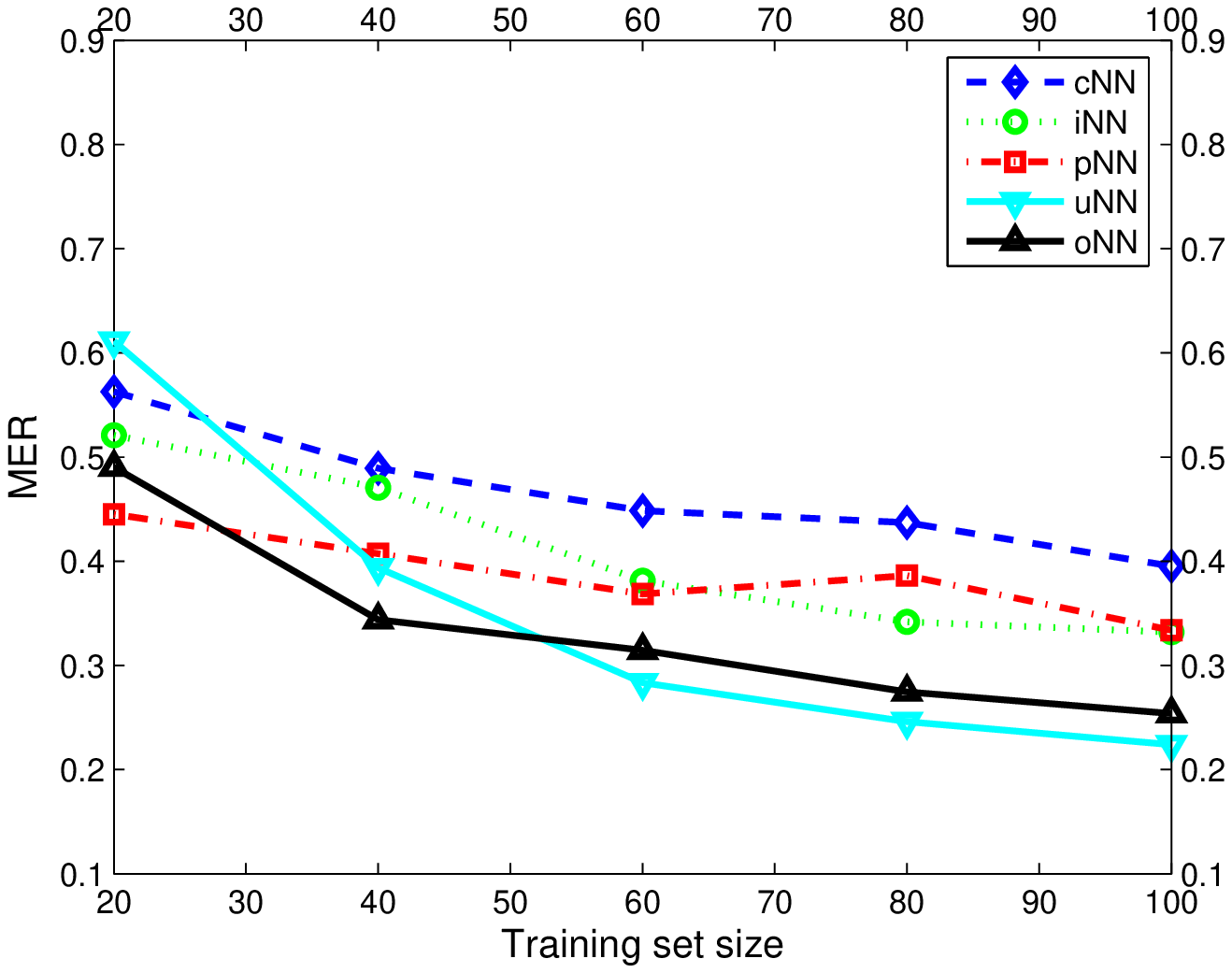}}    
\subfigure[MAE criterion.]{
        \label{fig:synthetic0205MAE}
        \includegraphics[width=0.31\linewidth]{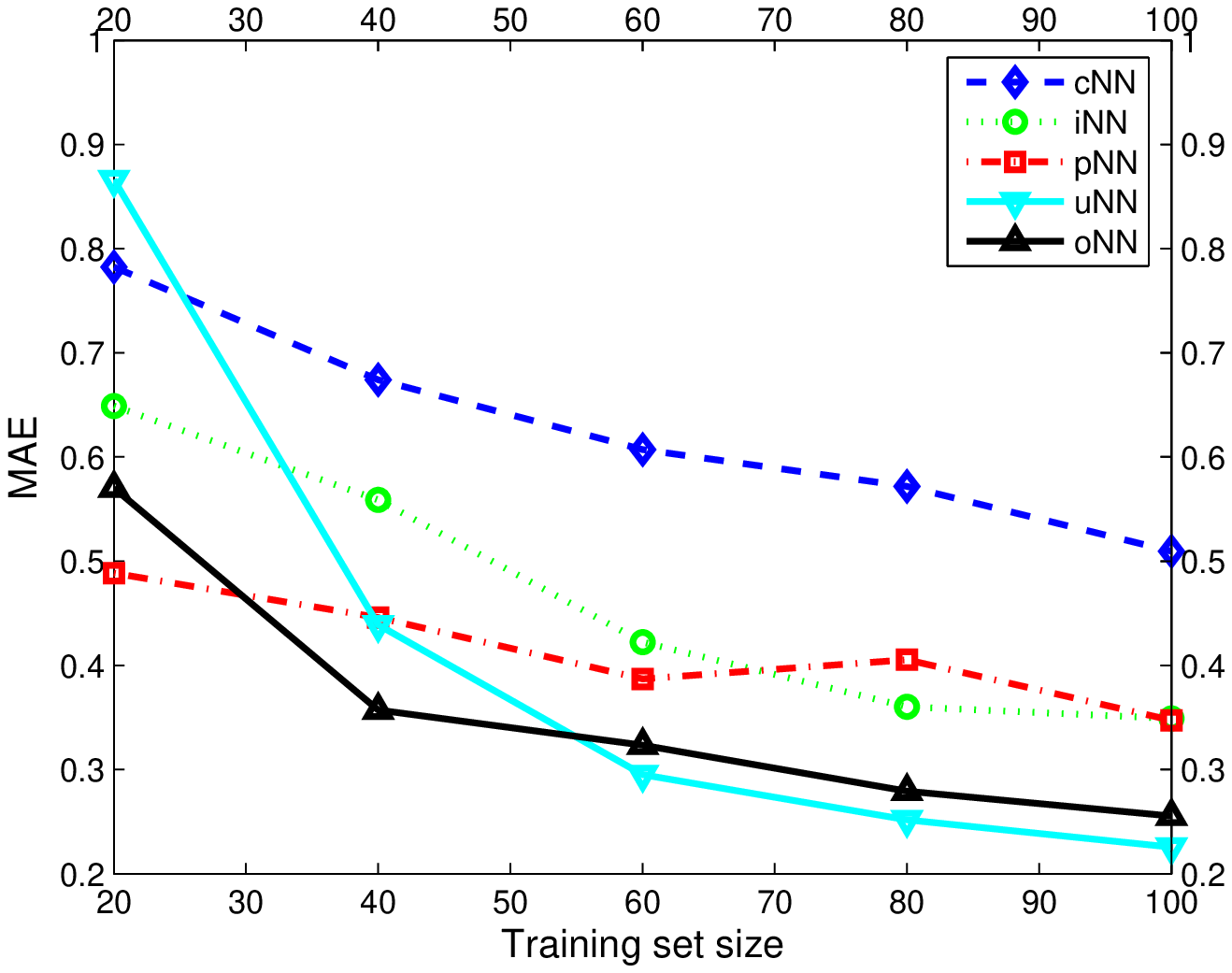}} 
\subfigure[MSE criterion.]{
        \label{fig:synthetic0205RMSE}
        \includegraphics[width=0.31\linewidth]{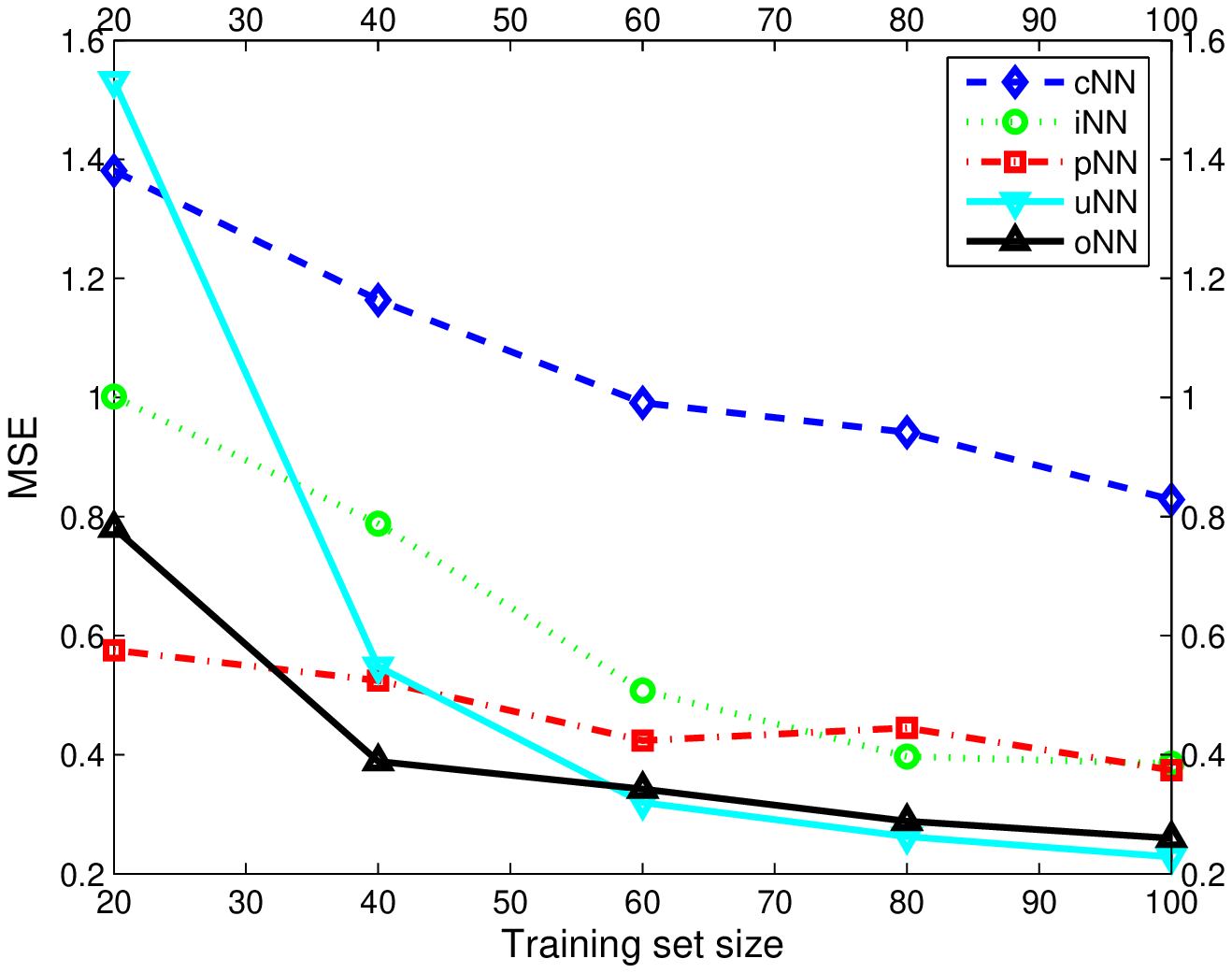}}               
\subfigure[Spearman coefficient.]{
        \label{fig:synthetic0205SpearmanCoef}
        \includegraphics[width=0.31\linewidth]{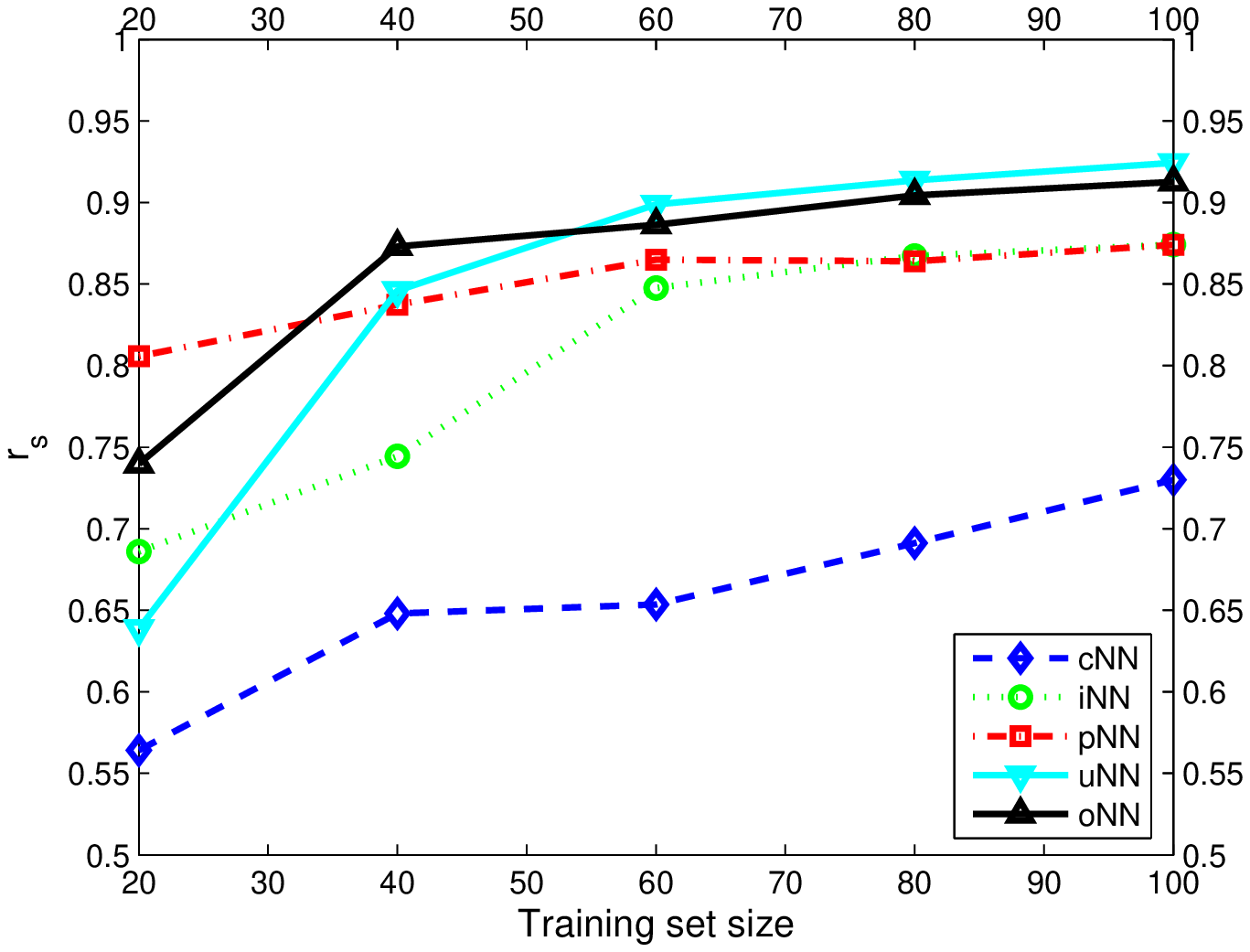}}  
\subfigure[Kendall's tau-b criterion.]{
        \label{fig:synthetic0205Kendall}
        \includegraphics[width=0.31\linewidth]{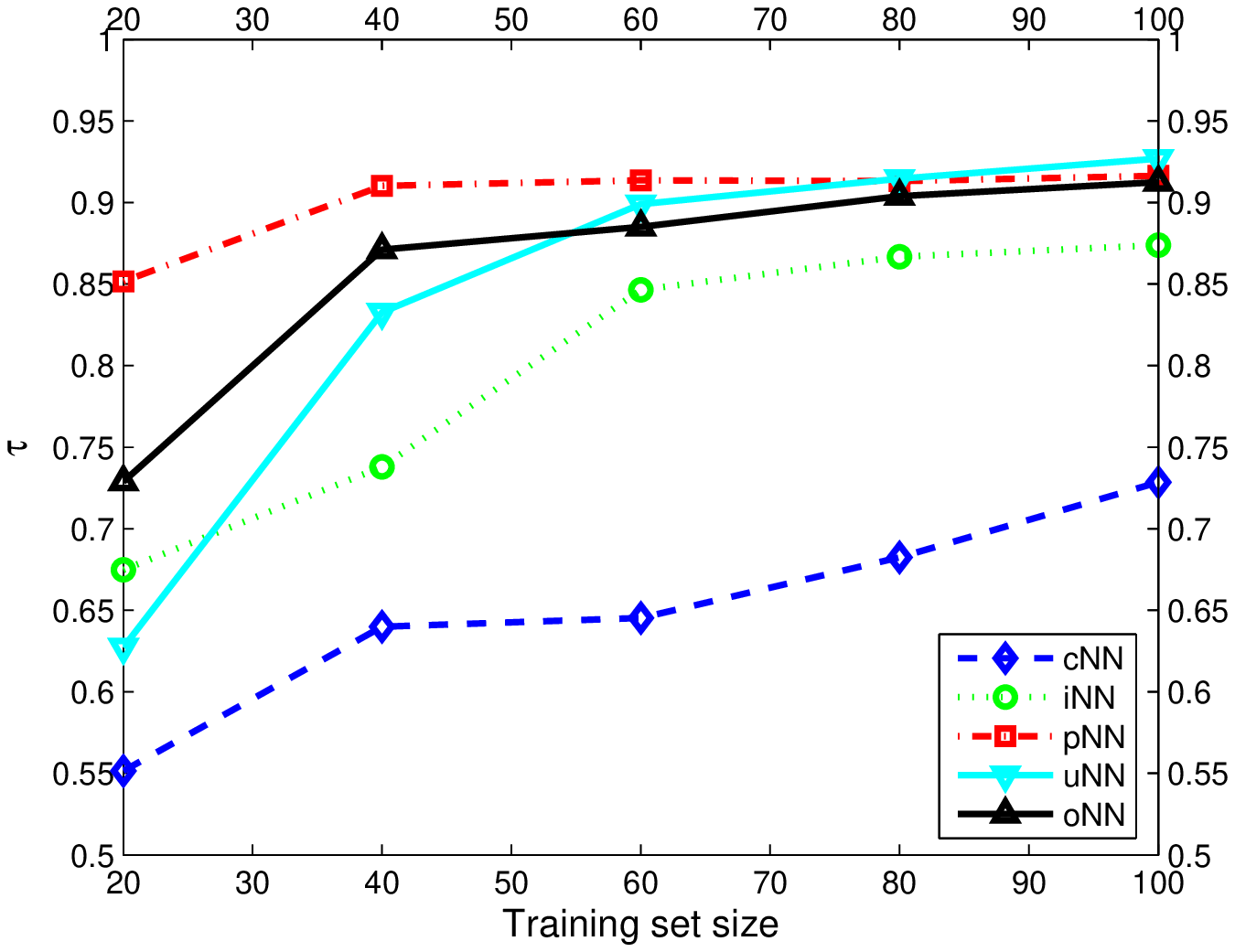}}                 
\subfigure[$o_c$ criterion.]{
        \label{fig:synthetic0205Prof}
        \includegraphics[width=0.31\linewidth]{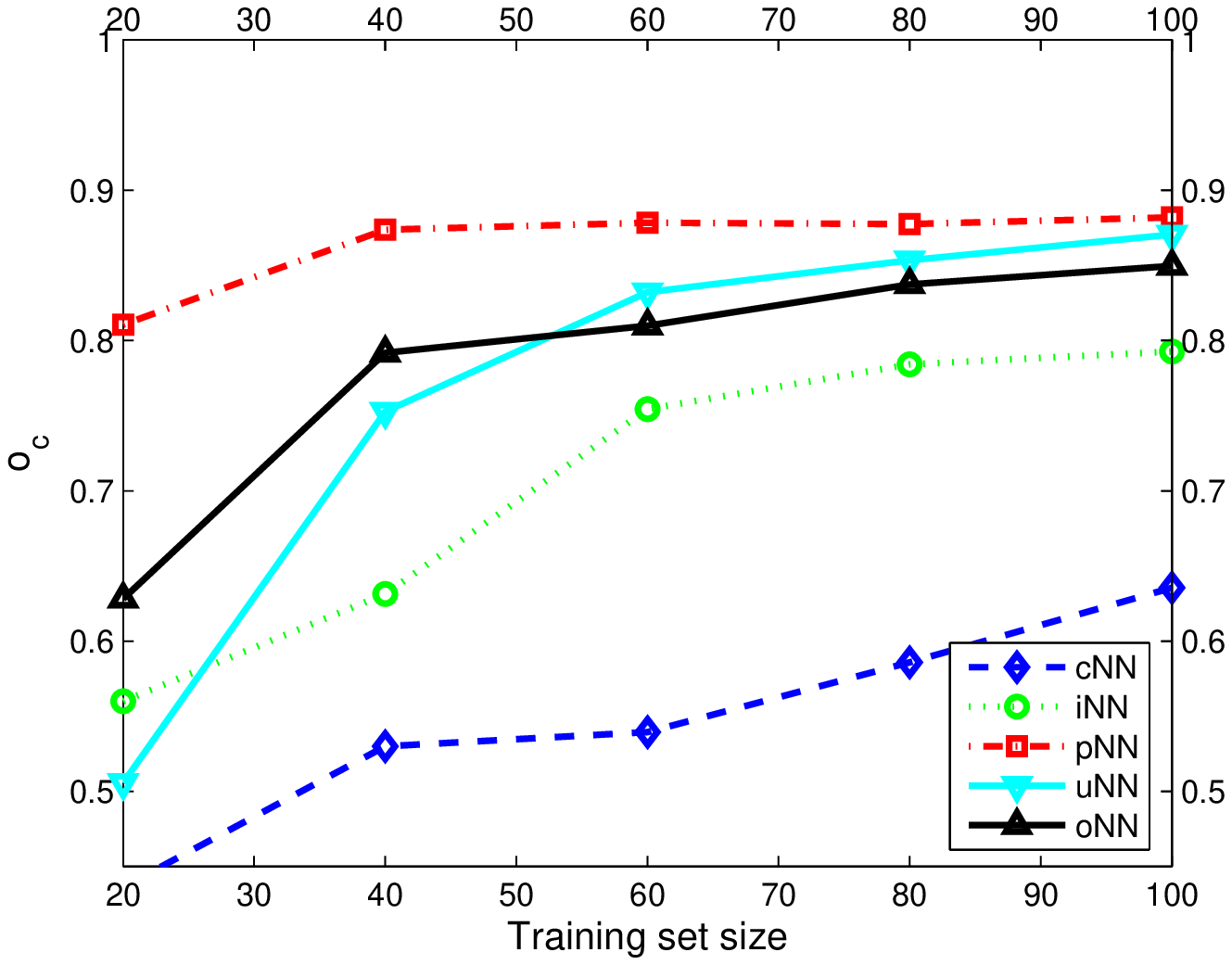}}           
\caption{NN results for $5$ classes in $\bbbr^2$, with 5 hidden units.}
\label{fig:synthetic0205NN}
\end{center}
\end{figure}

In order to compare the different algorithms, and similarly to \cite{Herbrich1999A}, we randomly selected 
training sequences of point-rank pairs of length $\ell$ ranging from $20$ to $100$.
The remaining points were used to estimate the classification error, which were averaged over $100$
runs of the algorithms for each size of the training sequence. 
Thus we obtained the learning curves shown in figure \ref{fig:synthetic0205NN}, for 5 neurons in the hidden layer.

\subsection{Accuracy dependence on the number of classes}
To investigate the relation between the number of classes and the performance of the evaluated algorithms, we also ran all models on the same dataset but with $10$ classes.

\begin{figure}
\begin{center}
\subfigure[Classes' boundaries.]{
        \label{fig:syntheticClasses10}
        \includegraphics[width=0.32\linewidth]{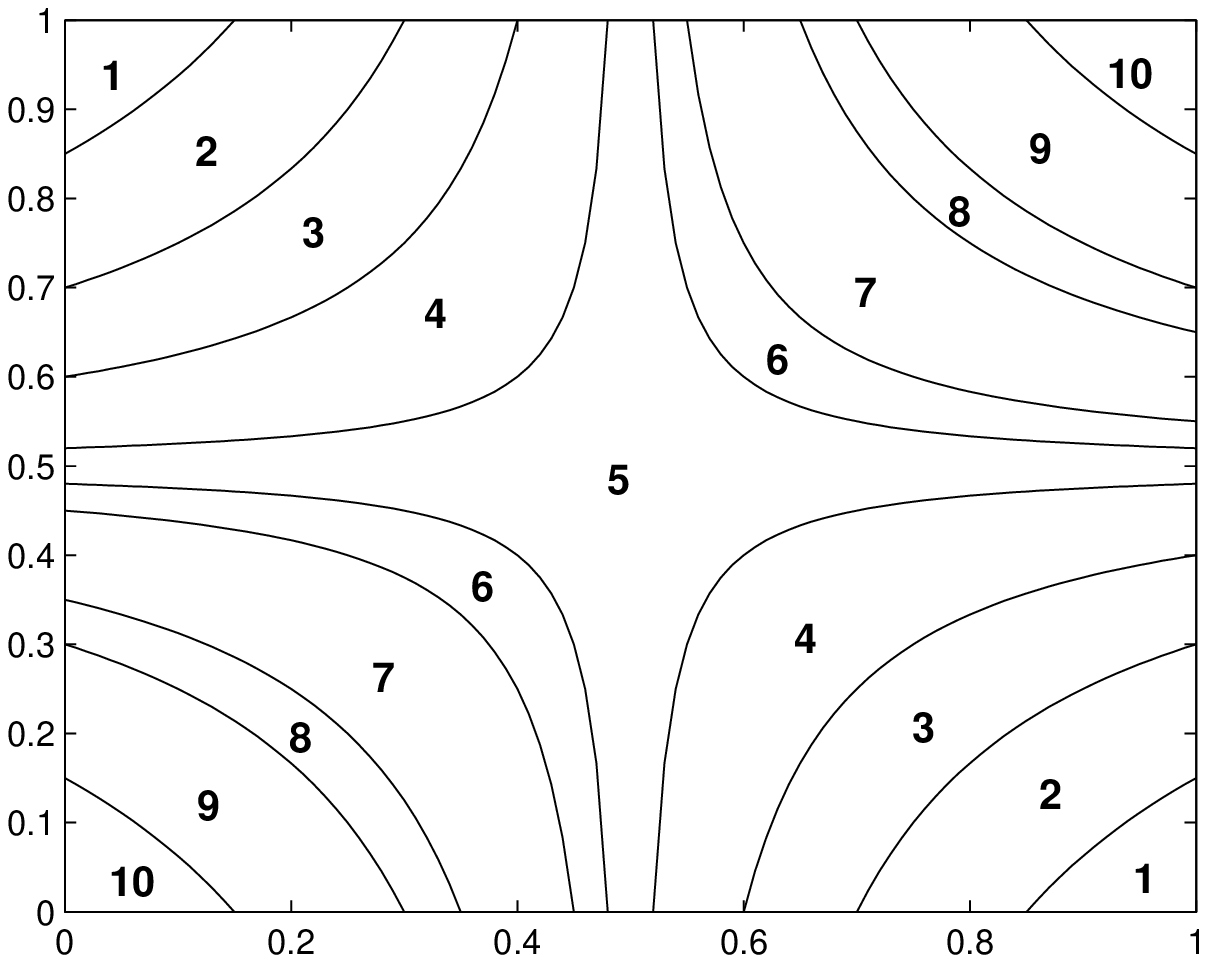}}
\subfigure[Scatter plot of the data points wrongly ranked. Number of wrong points: 13.9\%.]{
        \label{fig:syntheticData10}
        \includegraphics[width=0.32\linewidth]{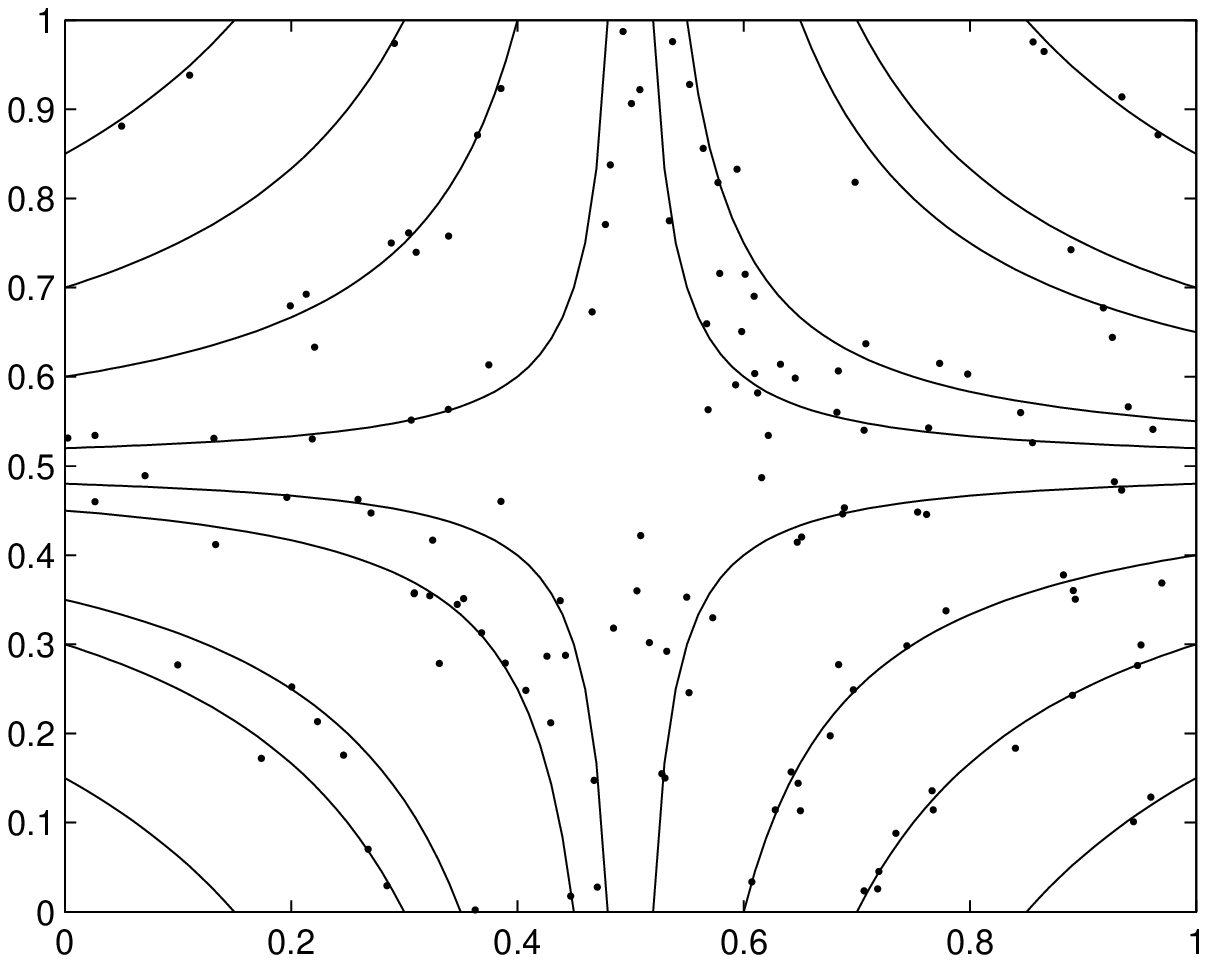}}
\subfigure[Class distribution.]{
        \label{fig:syntheticDataDist10}
        \includegraphics[width=0.32\linewidth]{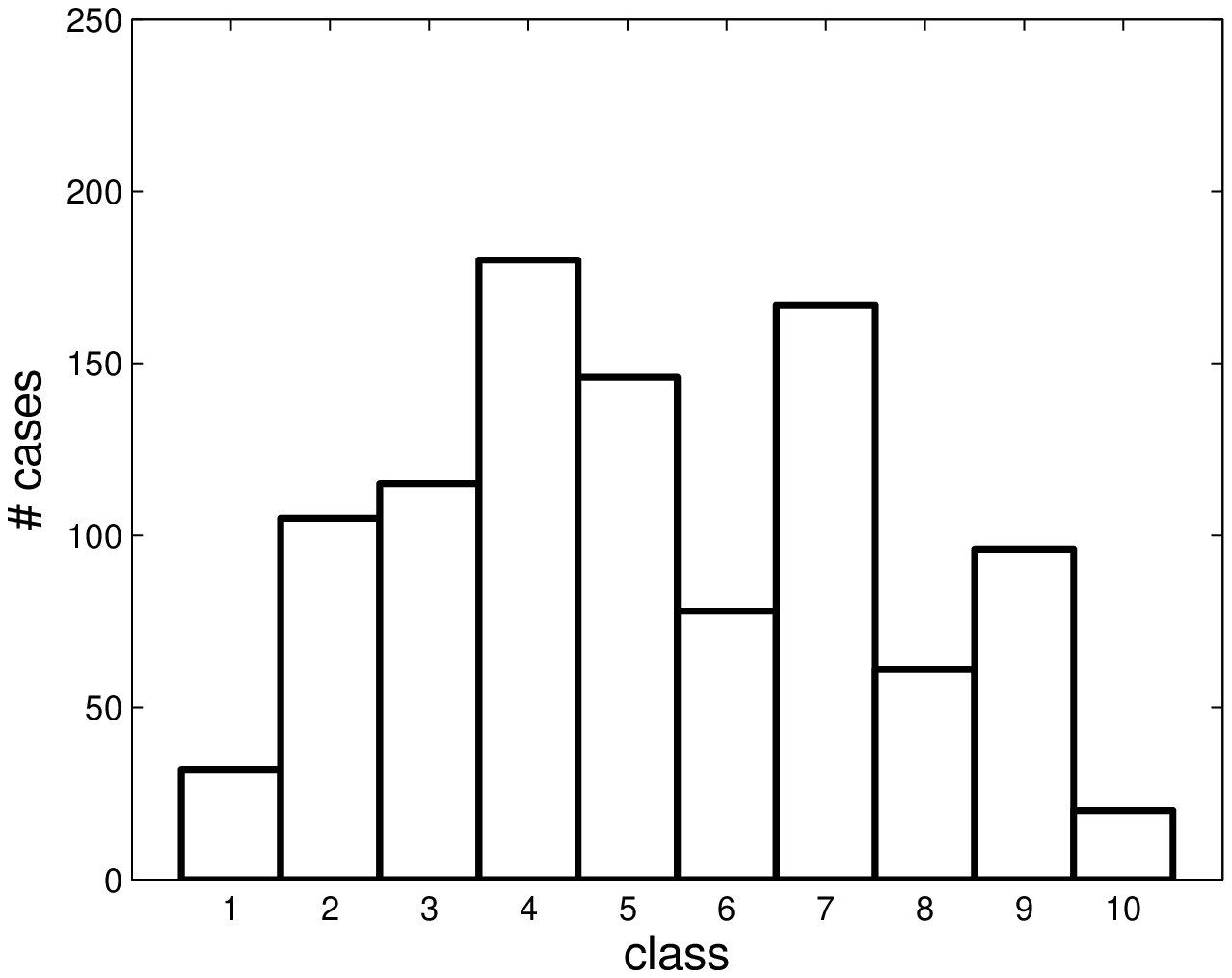}}              
\caption{Test setup for $10$ classes in $\bbbr^2$.}
\label{fig:synthetic0210}
\end{center}
\end{figure}

\begin{figure}
\begin{center}
\subfigure[MER criterion.]{
        \label{fig:synthetic0210MER}
        \includegraphics[width=0.31\linewidth]{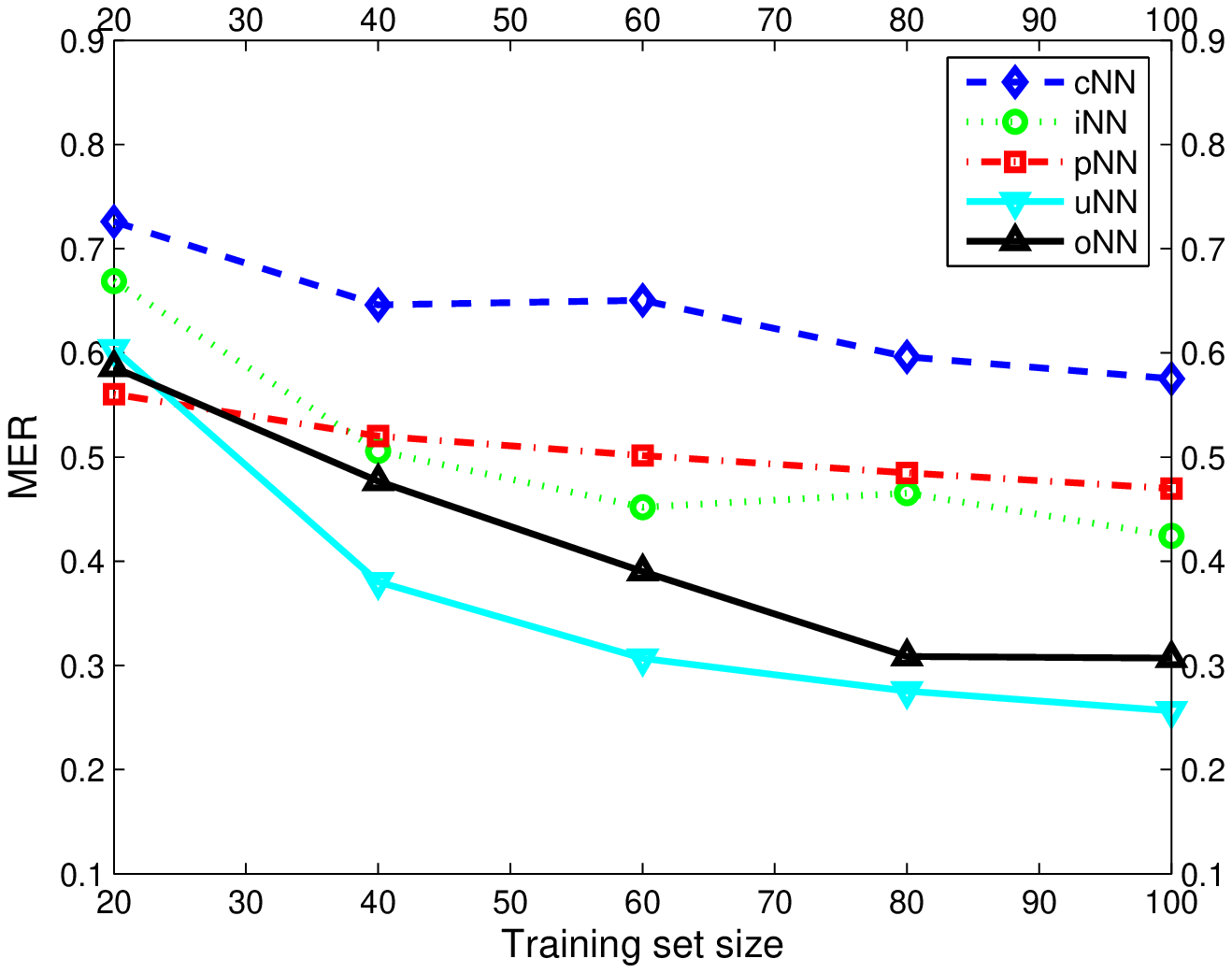}} 
\subfigure[MAE criterion.]{
        \label{fig:synthetic0210MAE}
        \includegraphics[width=0.31\linewidth]{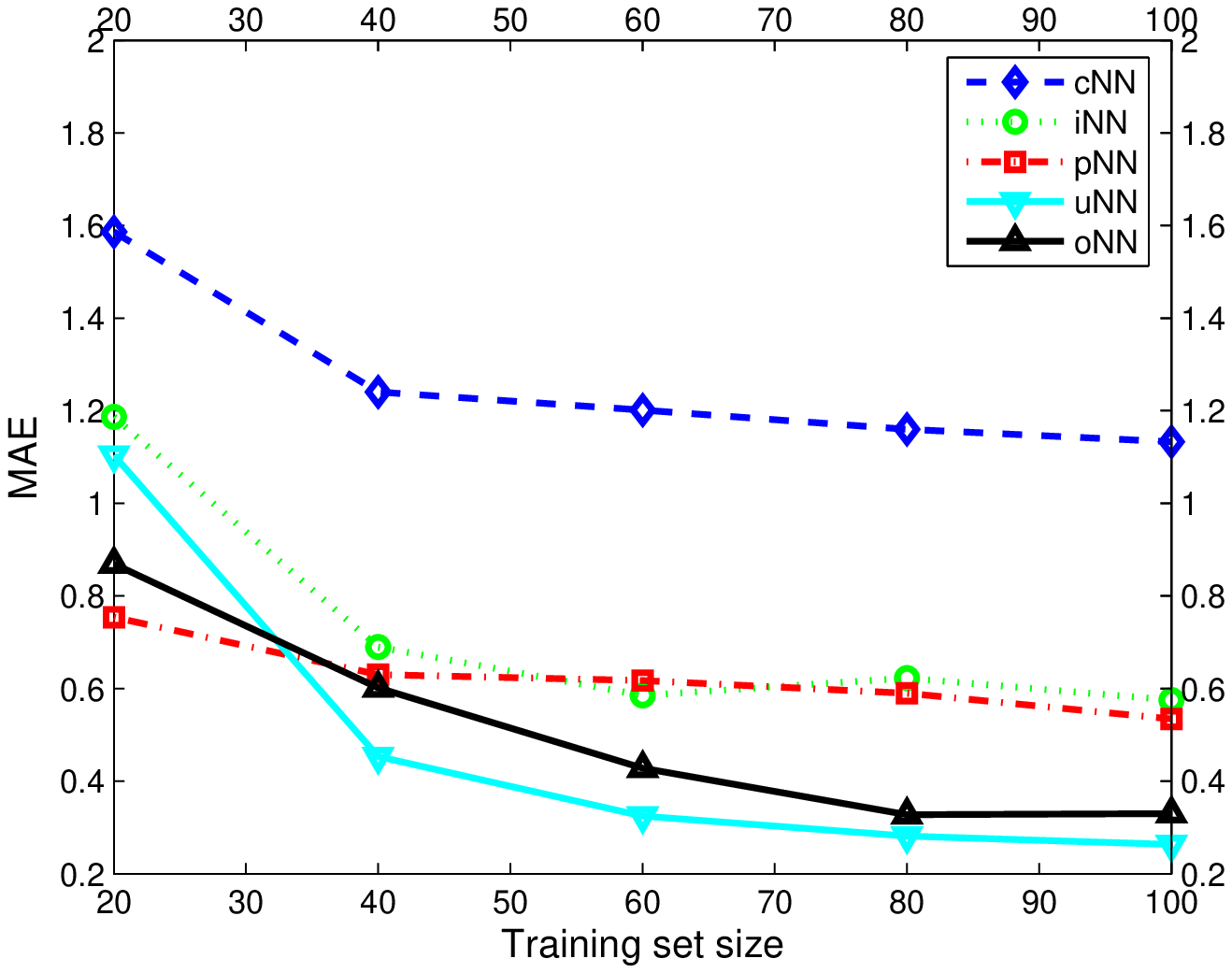}}         
\subfigure[MSE criterion.]{
        \label{fig:synthetic0210RMSE}
        \includegraphics[width=0.31\linewidth]{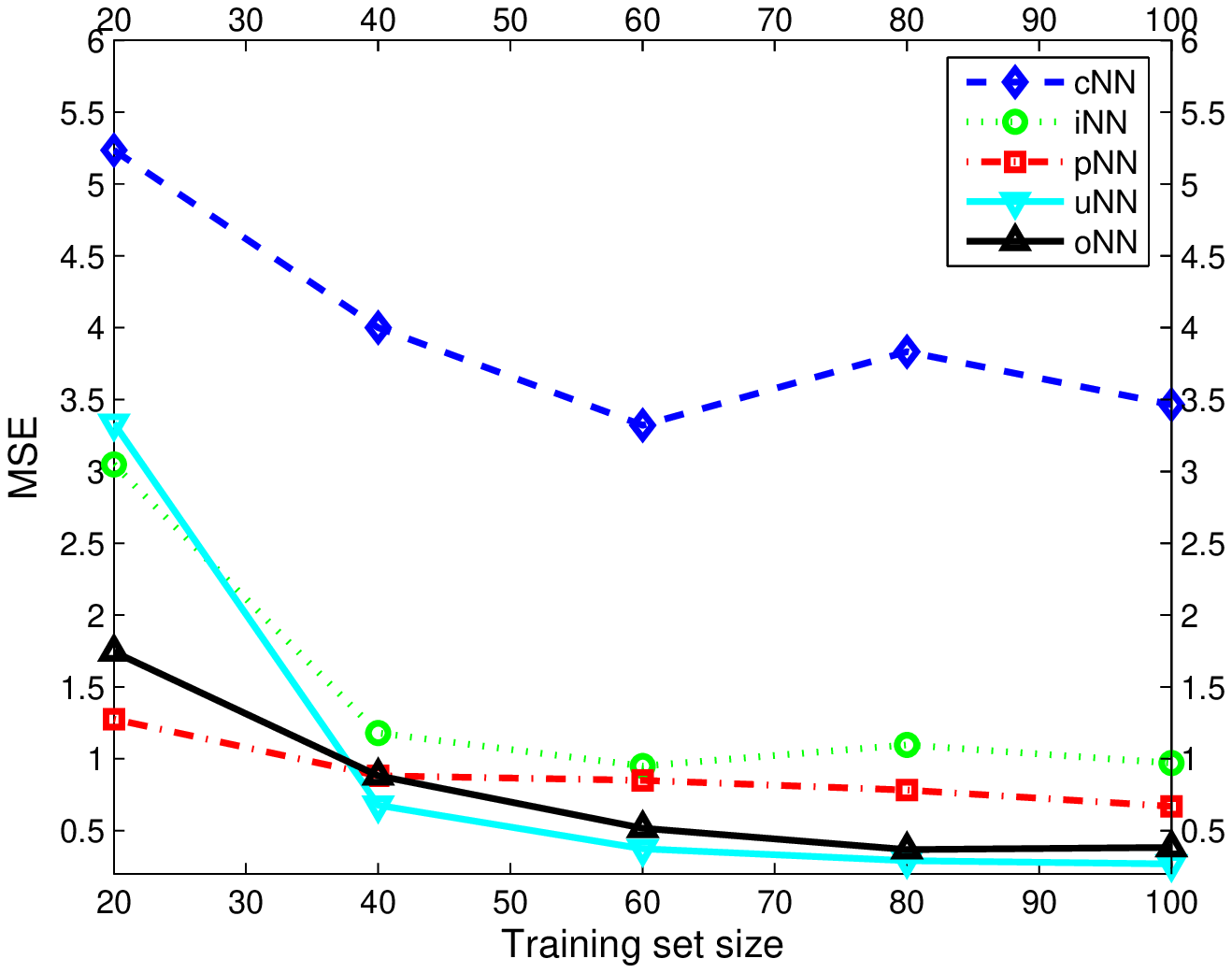}}    
\subfigure[Spearman coefficient.]{
        \label{fig:synthetic0210SpearmanCoef}
        \includegraphics[width=0.31\linewidth]{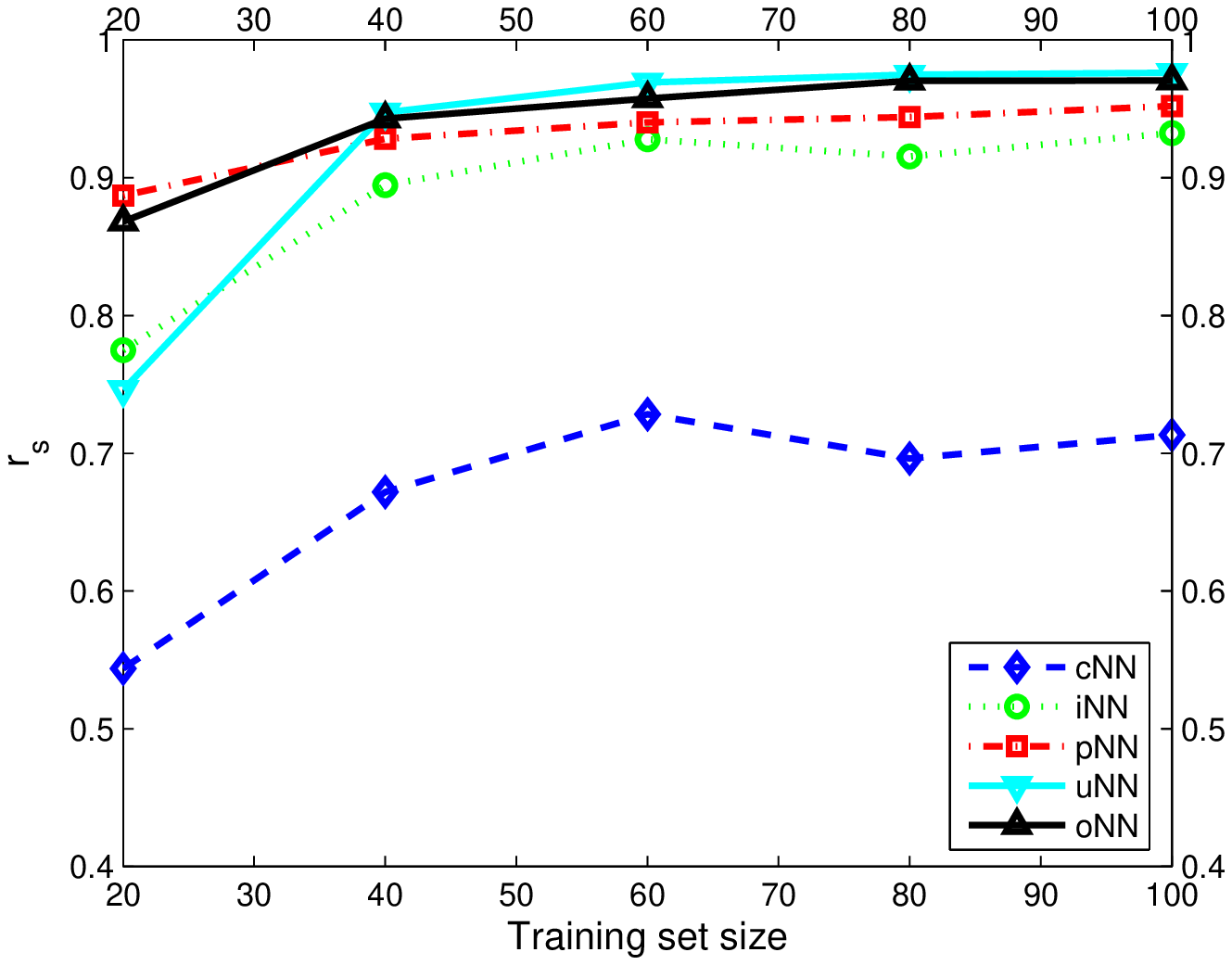}}   
\subfigure[Kendall's tau-b coefficient.]{
        \label{fig:synthetic0210Kendall}
        \includegraphics[width=0.31\linewidth]{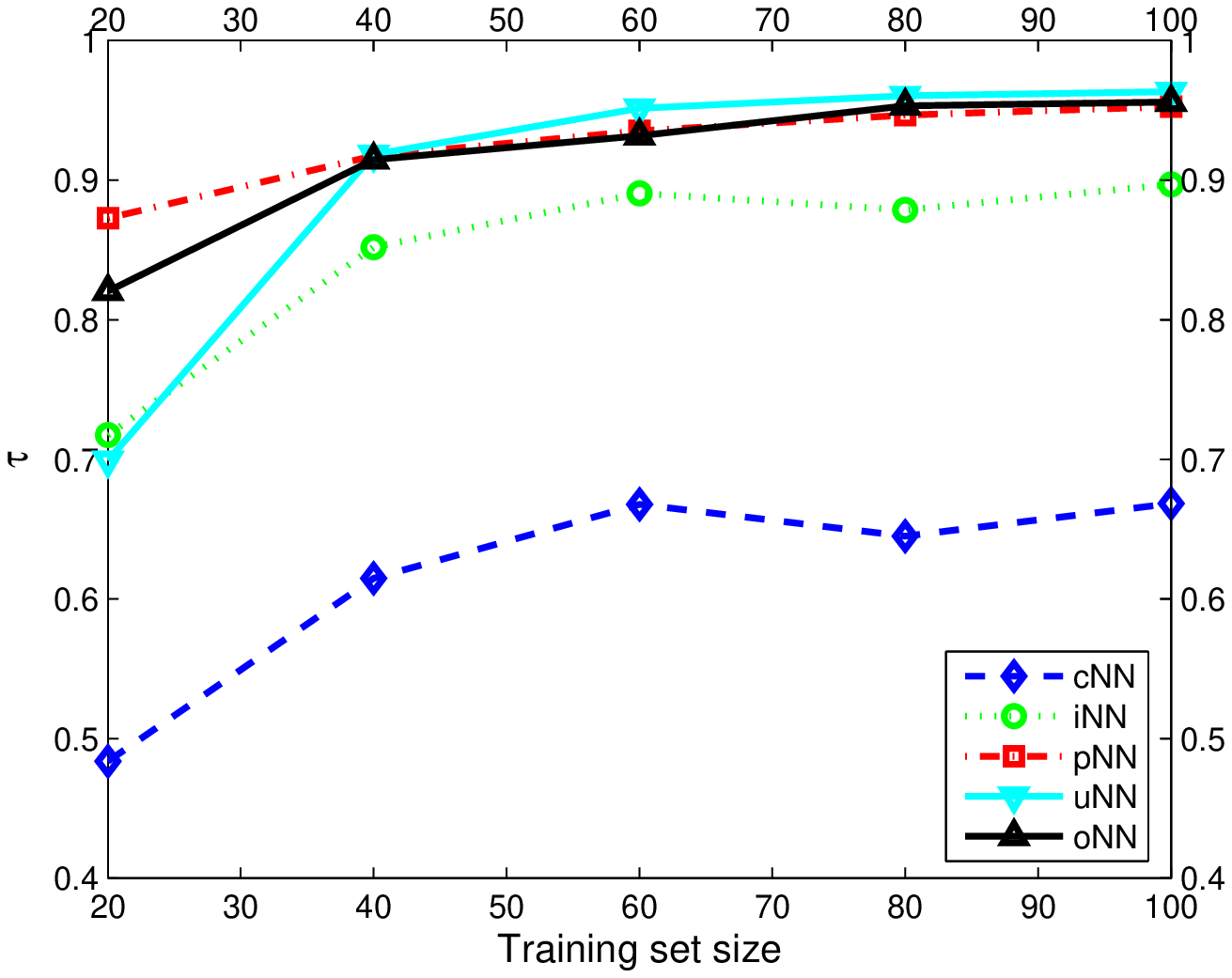}}   
\subfigure[$o_c$ coefficient.]{
        \label{fig:synthetic0210Prof}
        \includegraphics[width=0.31\linewidth]{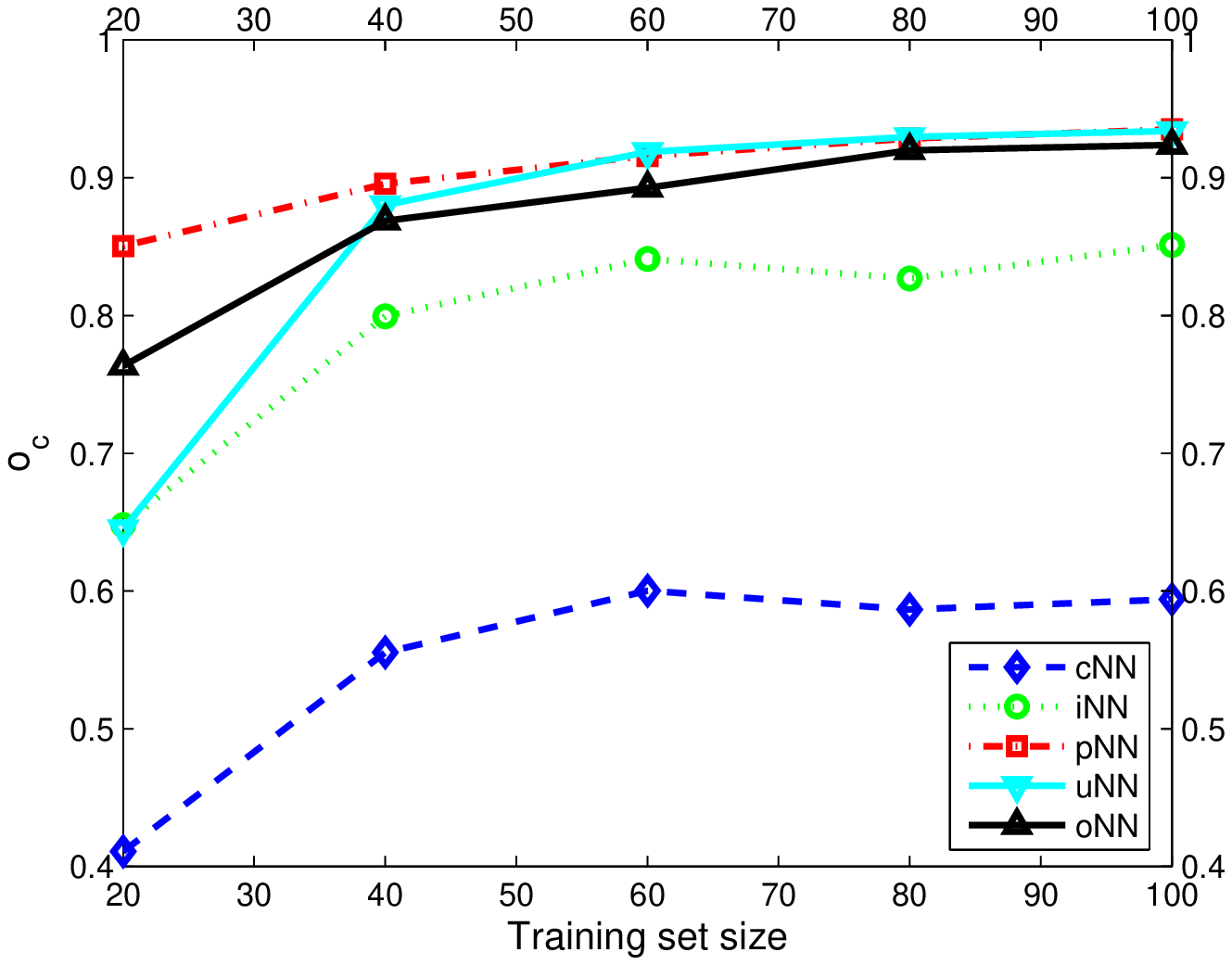}}                
\caption{NN results for $10$ classes in $\bbbr^2$, with 5 hidden units.}
\label{fig:synthetic0210NN}
\end{center}
\end{figure}

This time each point was assigned a rank $y$ from the set $\{1, 2, 3, 4, 5, 6, 7, 8, 9, 10\}$, according to \\
\[y= \min_{r\in\{1, 2, 3, 4, 5, 6, 7, 8, 9, 10\}}\{r: b_{r-1} < 10(x_1-0.5)(x_2-0.5) + \varepsilon\ < b_r\}\]
\begin{multline*}(b_0, b_1, b_2, b_3,b_4,b_5, b_6, b_7, b_8, b_9, b_{10})=(-\infty, -1.75, -1, -0.5, -0.1, 0.1, 0.25, 0.75, 1, 1.75, +\infty)\end{multline*}
where $\varepsilon$ is a random value, normally distributed with zero mean and standard deviation $\sigma = 0.125/2$.
Figure \ref{fig:syntheticClasses10} shows the ten regions and figure \ref{fig:syntheticData10} the points which were assigned to a different rank after the corruption with the normally distributed noise. 
The learning curves obtained for this arrangement are shown in figure \ref{fig:synthetic0210NN} (again, for 5 neurons in the hidden layer).

\subsection{Accuracy dependence on the data dimension}
The described experiments in $\bbbr^2$ were repeated for data points in $\bbbr^4$, to evaluate the influence of data dimension on models' relative performance.
 
We generated $2000$ example points $\textbf{x}=[x_1 \ x_2 \ x_3 \ x_4]^t$ uniformly at random in the unit square in $\bbbr^4$.

For $5$ classes, each point was assigned a rank $y$ from the set $\{1, 2, 3, 4, 5\}$, according to \\
\[y= \min_{r\in\{1, 2, 3, 4, 5\}}\{r: b_{r-1} < 1000\prod_{i=1}^4 (x_i-0.5) + \varepsilon\ < b_r\}\]
\[(b_0, b_1, b_2, b_3,b_4)=(-\infty, -2.5, -0.5, 0.5, 3, +\infty)\]
where $\varepsilon$ is a random value, normally distributed with zero mean and standard deviation $\sigma = 0.25$.

\begin{figure}
\begin{center}
\subfigure[K=5.]{
        \includegraphics[width=0.32\linewidth]{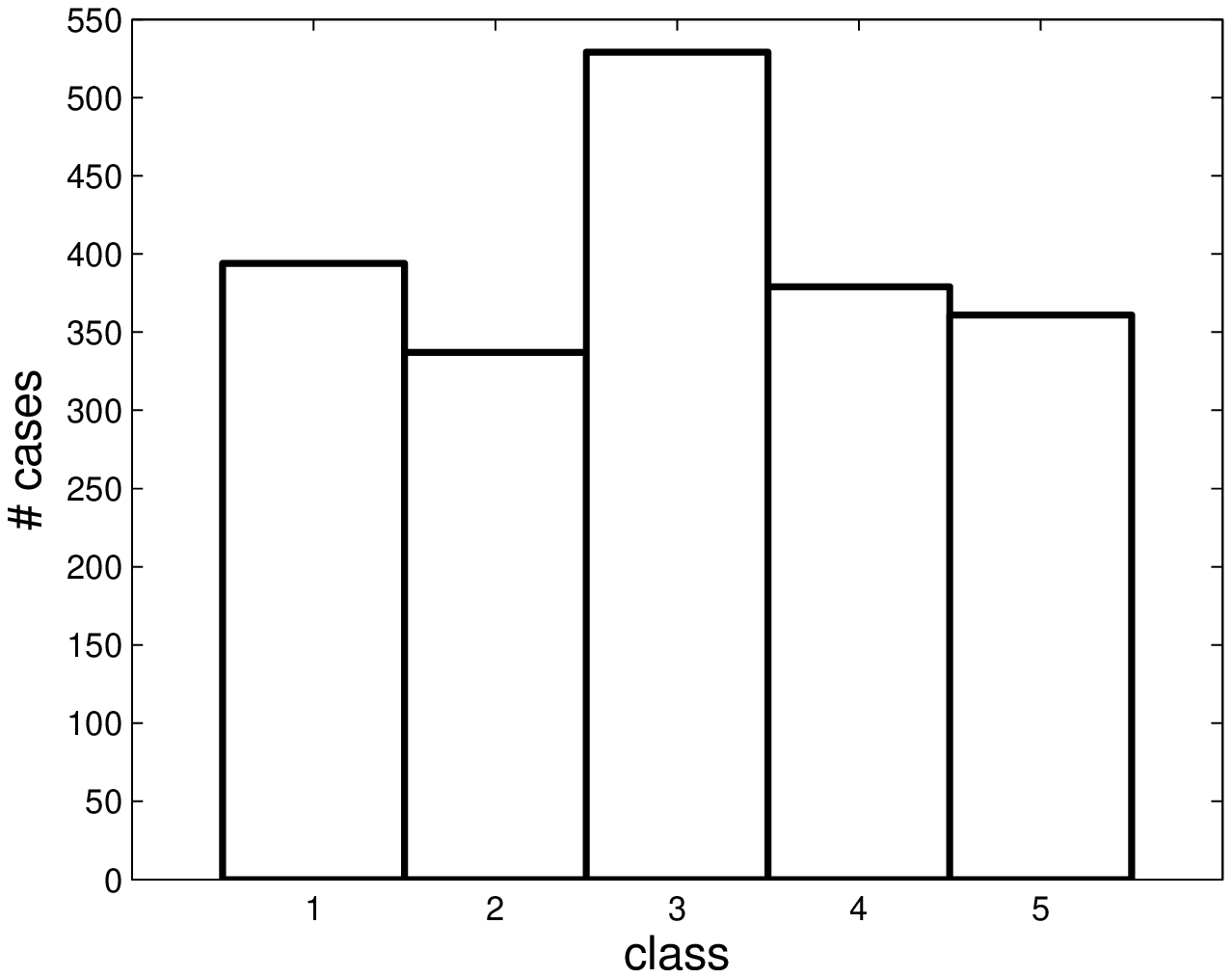}}
\subfigure[K=10.]{
        \includegraphics[width=0.32\linewidth]{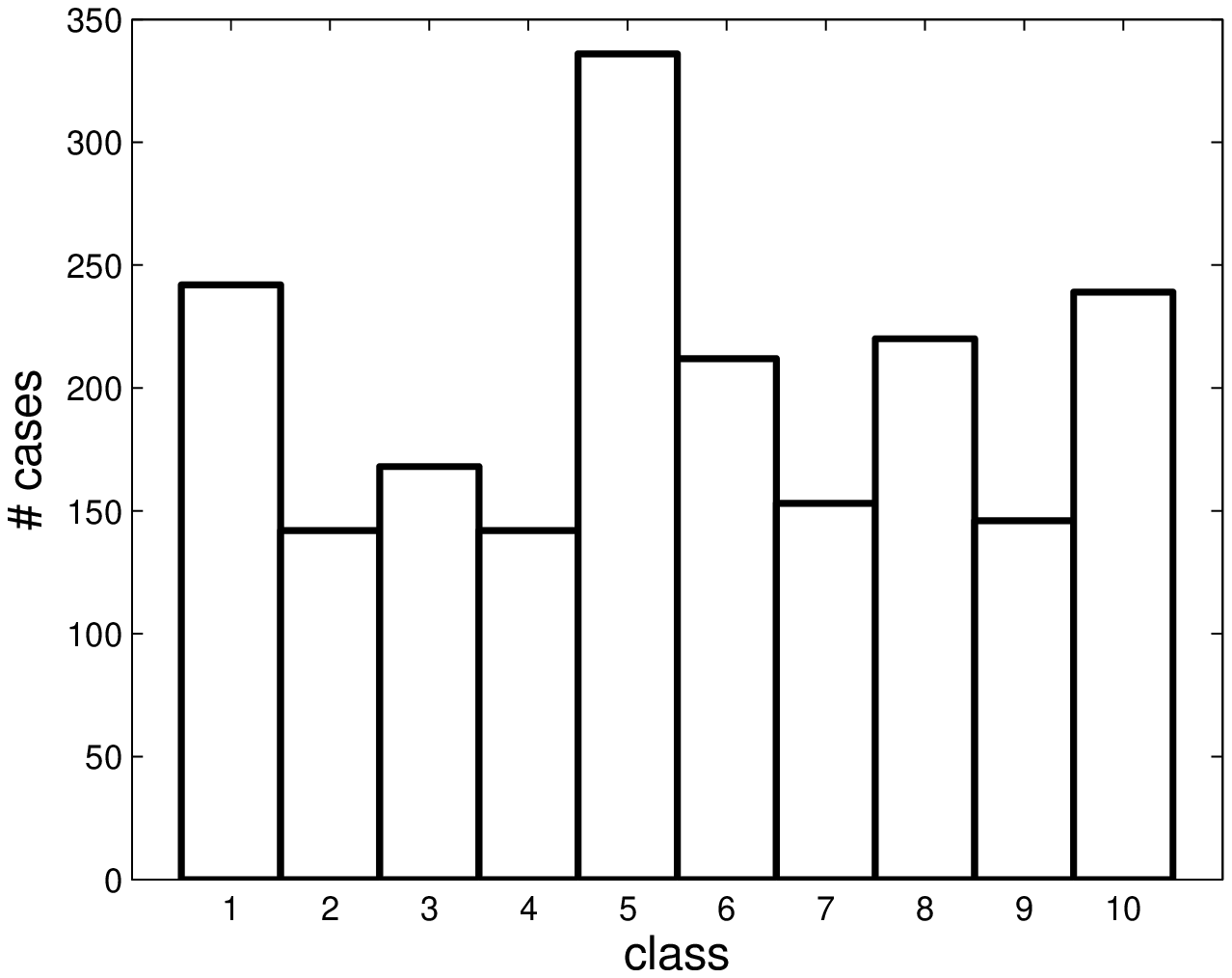}}        
\caption{Class distribution in $\bbbr^4$.}
\label{fig:synthetic04}
\end{center}
\end{figure}

\begin{figure}
\begin{center}
\subfigure[MER criterion.]{
        \label{fig:synthetic0405MER}
        \includegraphics[width=0.31\linewidth]{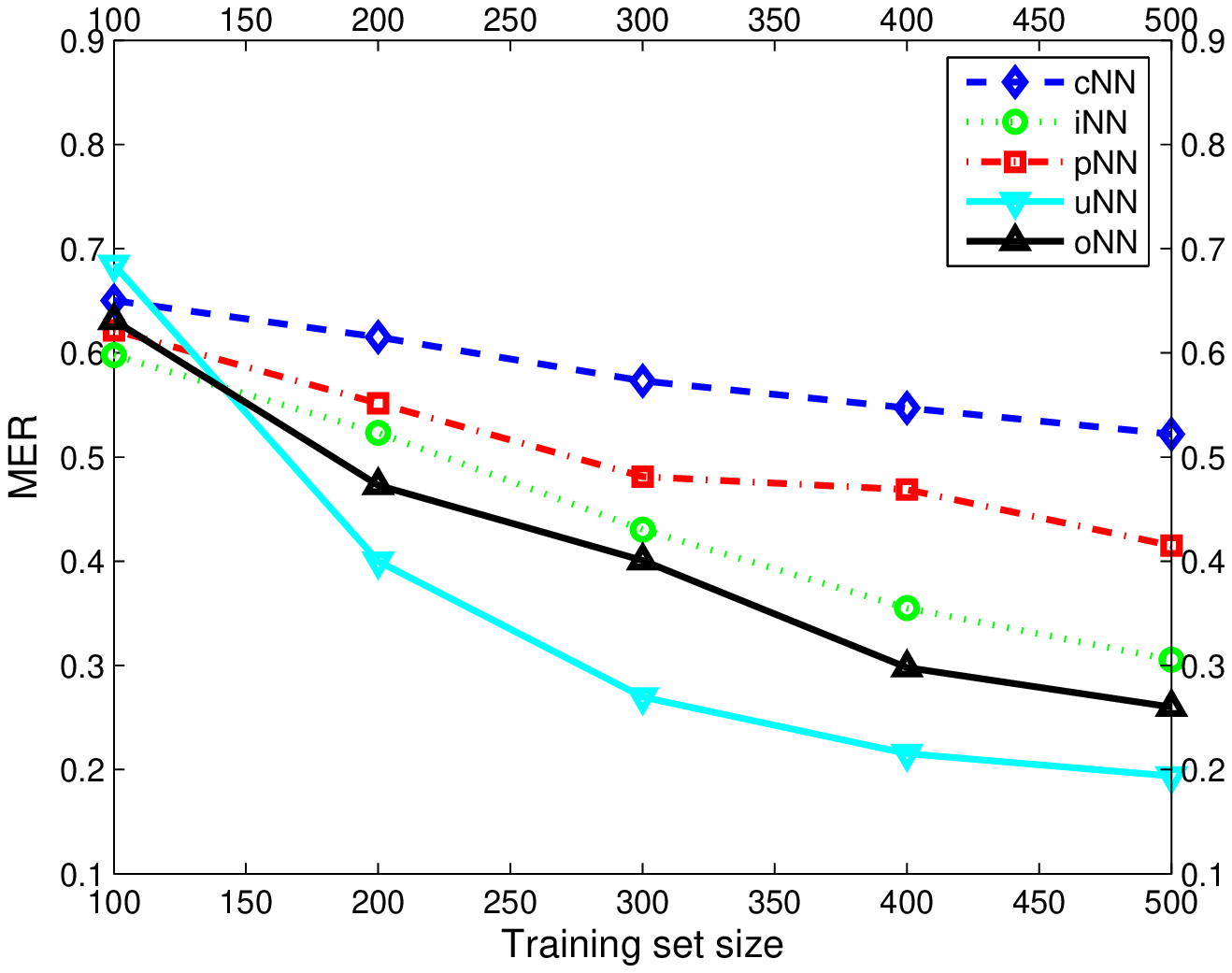}} 
\subfigure[MAE criterion.]{
        \label{fig:synthetic0405MAE}
        \includegraphics[width=0.31\linewidth]{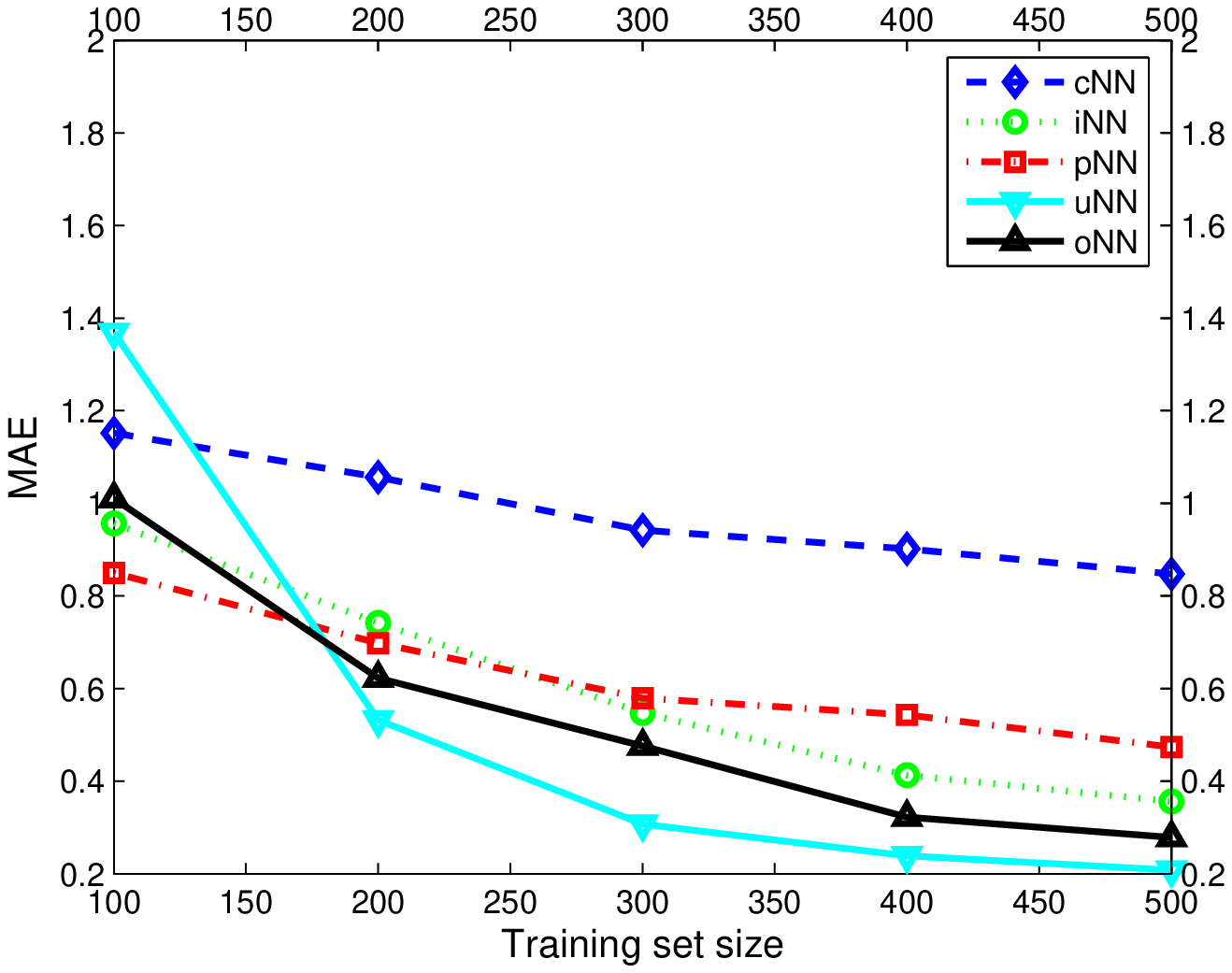}}         
\subfigure[MSE criterion.]{
        \label{fig:synthetic0405RMSE}
        \includegraphics[width=0.31\linewidth]{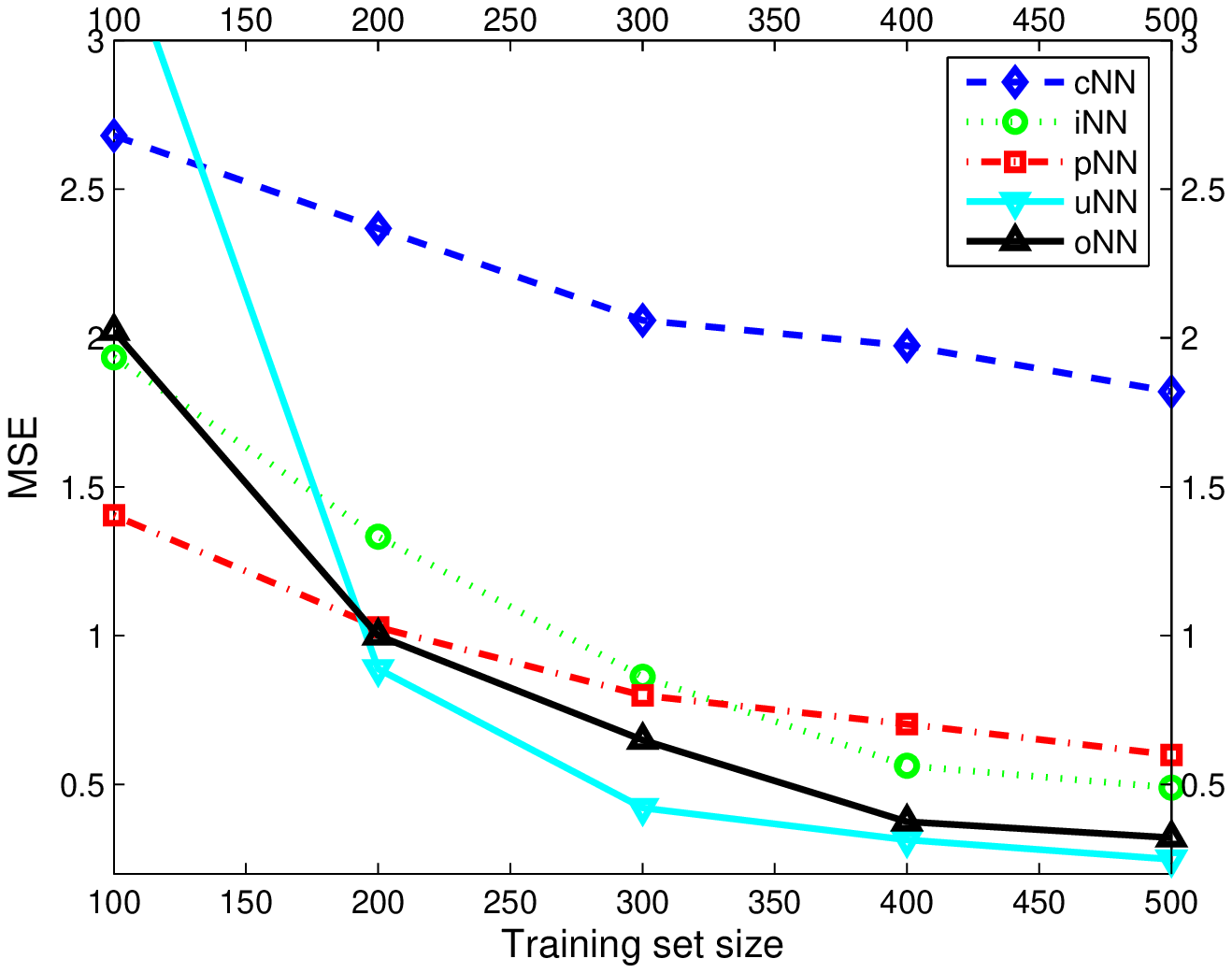}}    
\subfigure[Spearman coefficient.]{
        \label{fig:synthetic0405SpearmanCoef}
        \includegraphics[width=0.31\linewidth]{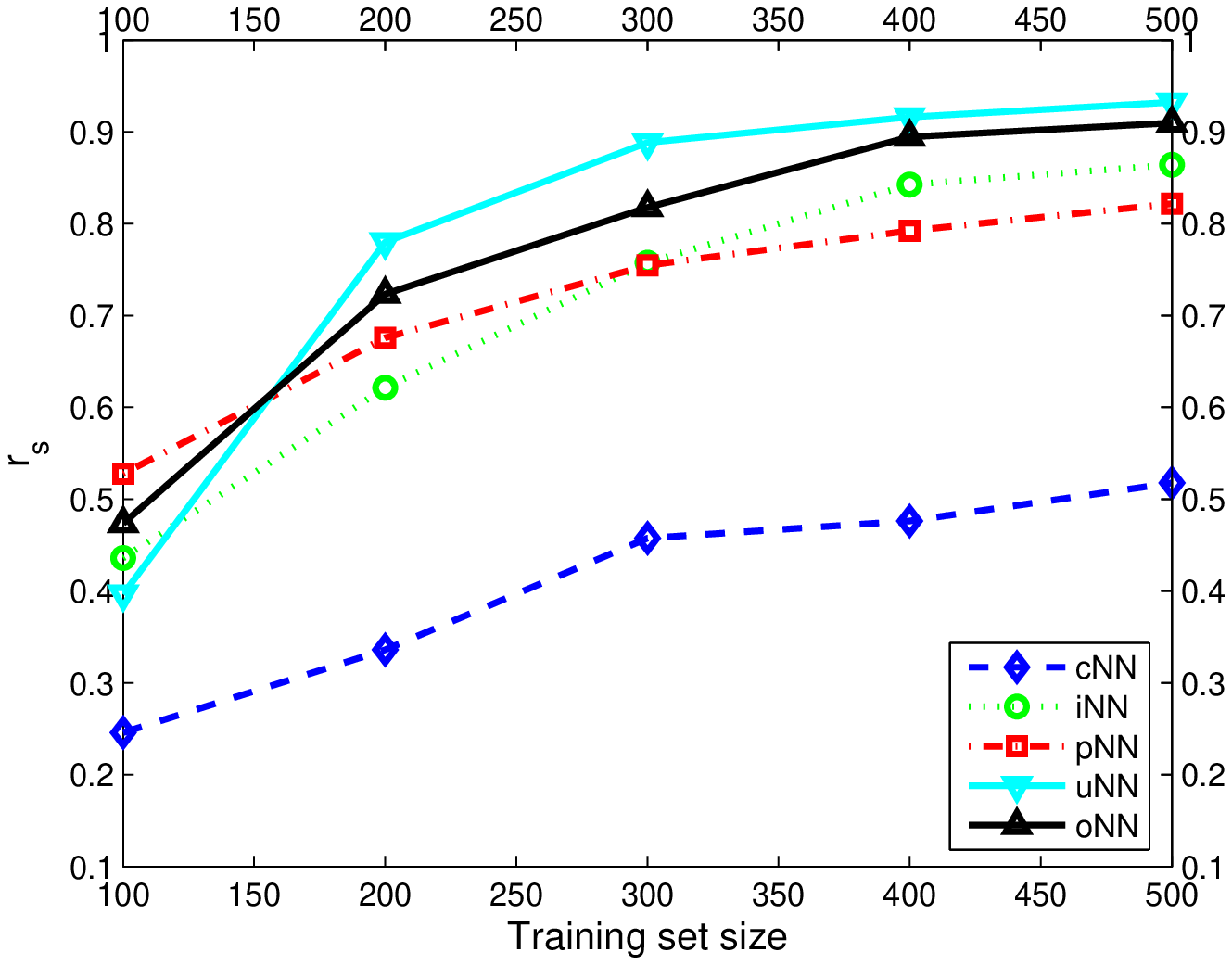}}    
\subfigure[Kendall's tau-b coefficient.]{
        \label{fig:synthetic0405Kendall}
        \includegraphics[width=0.31\linewidth]{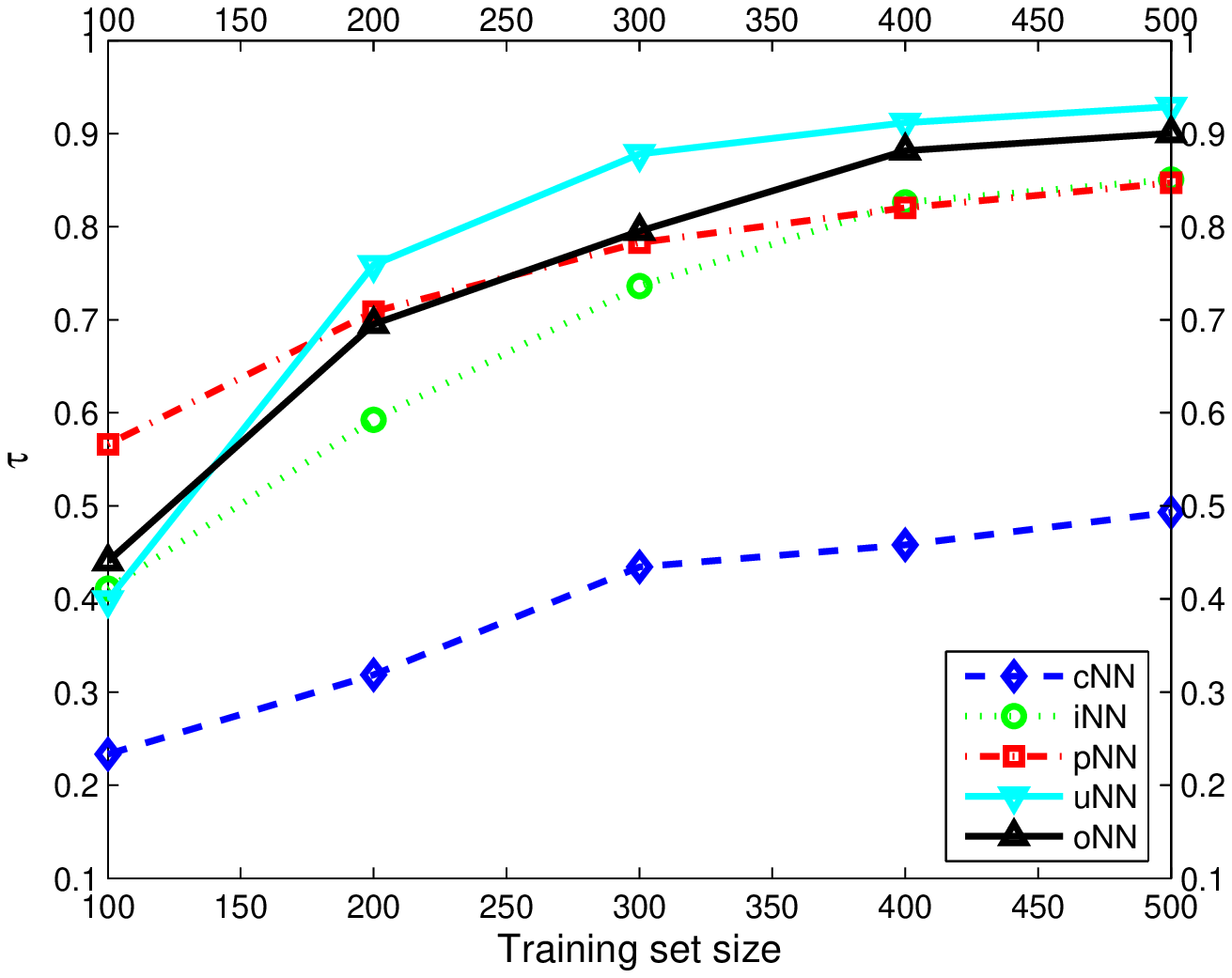}}    
\subfigure[$o_c$ coefficient.]{
        \label{fig:synthetic0405Prof}
        \includegraphics[width=0.31\linewidth]{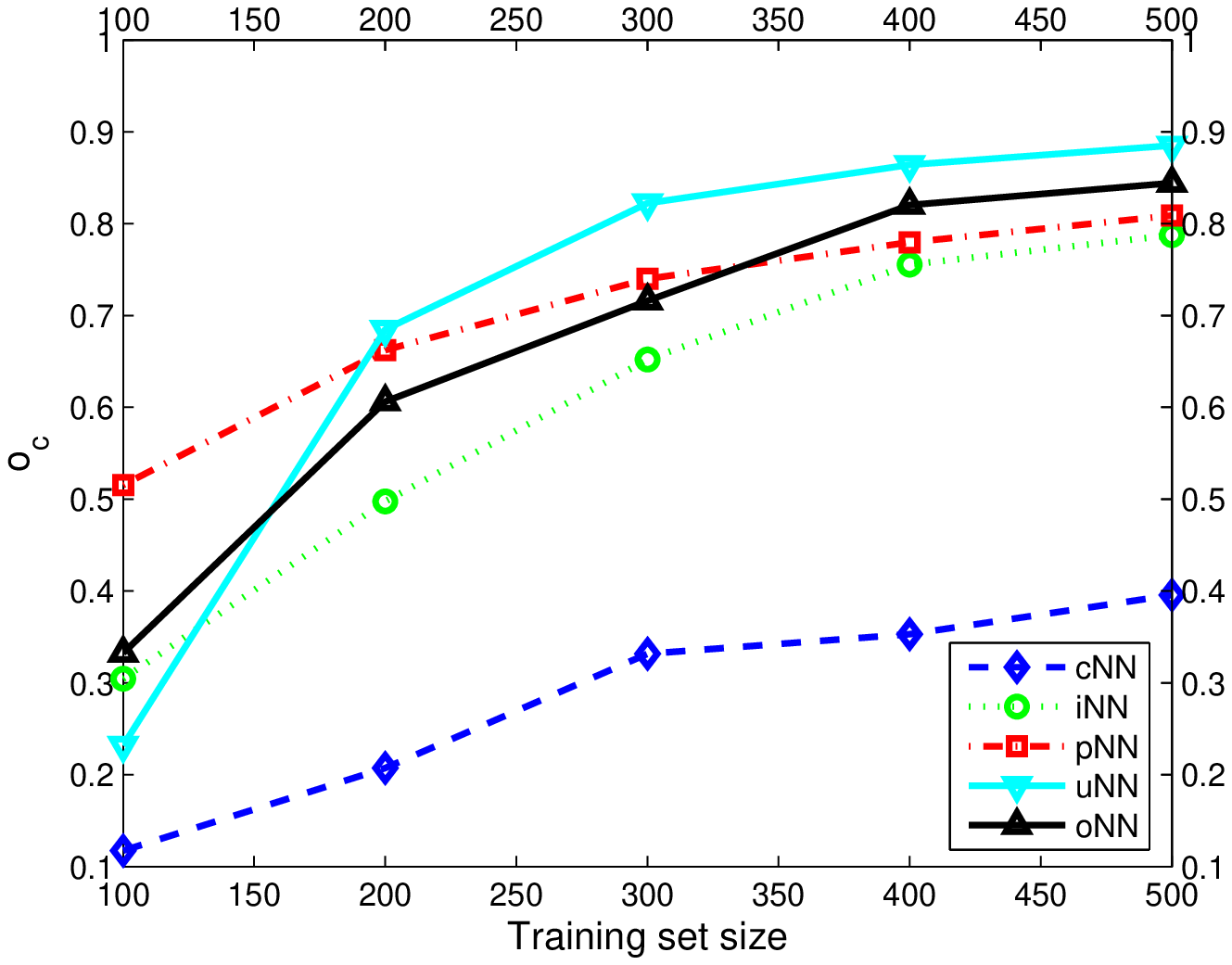}}                      
\caption{NN results for $5$ classes in $\bbbr^4$, with 16 hidden units.}
\label{fig:synthetic0405NN}
\end{center}
\end{figure}

\begin{figure}
\begin{center}
\subfigure[MER criterion.]{
        \label{fig:synthetic0410MER}
        \includegraphics[width=0.31\linewidth]{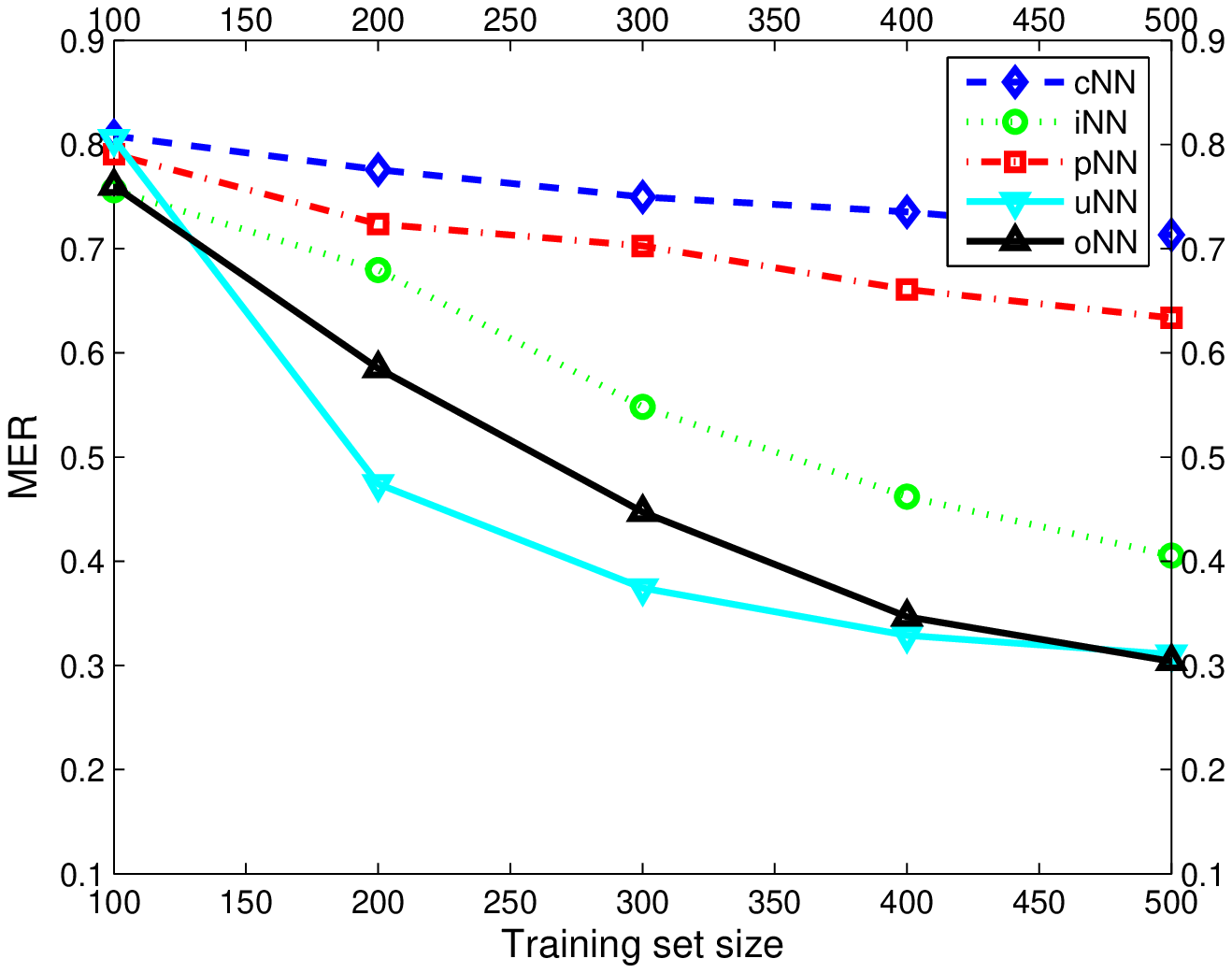}} 
\subfigure[MAE criterion.]{
        \label{fig:synthetic0410MAE}
        \includegraphics[width=0.31\linewidth]{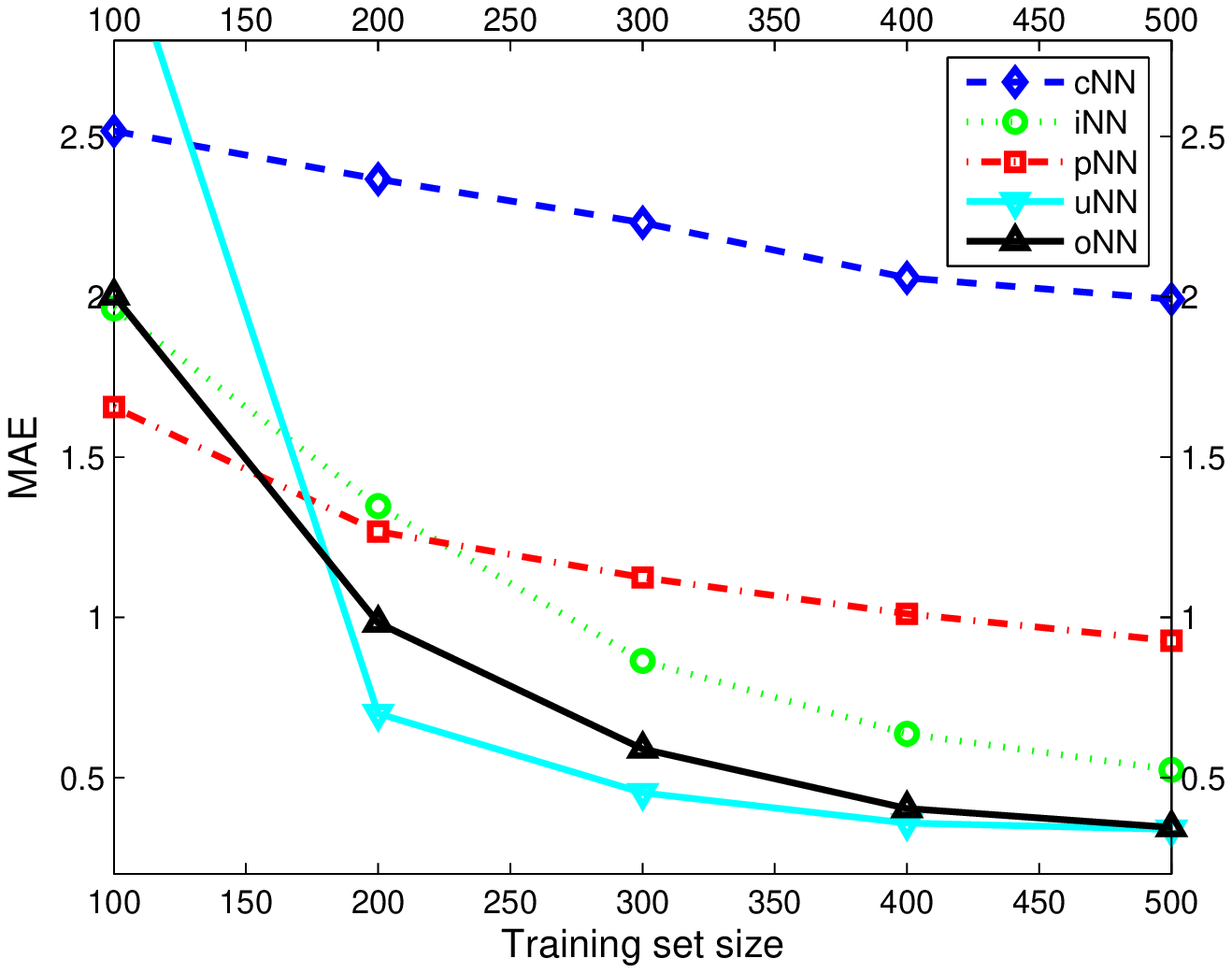}} 
\subfigure[MSE criterion.]{
        \label{fig:synthetic0410RMSE}
        \includegraphics[width=0.31\linewidth]{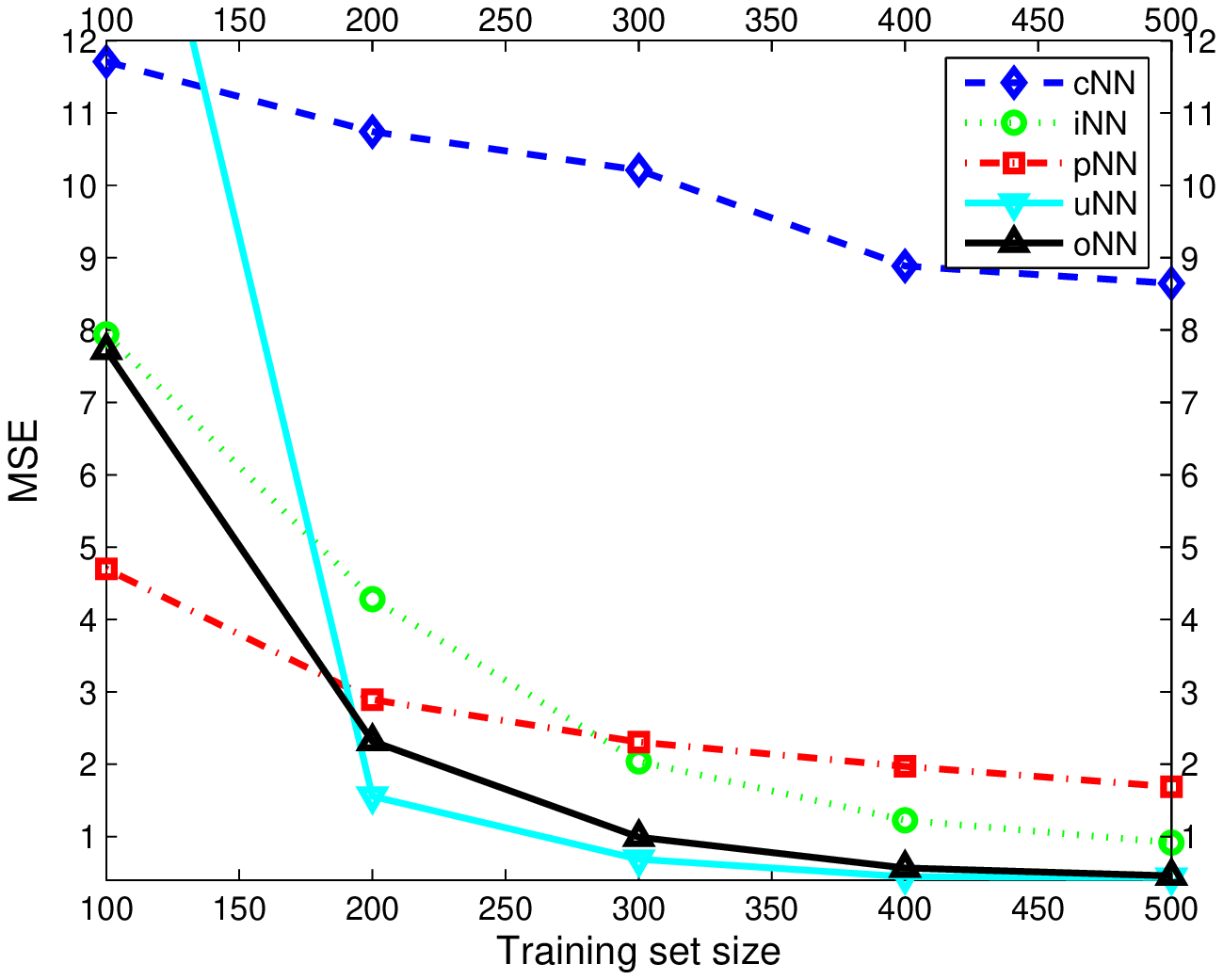}}    
\subfigure[Spearman coefficient.]{
        \label{fig:synthetic0410SpearmanCoef}
        \includegraphics[width=0.31\linewidth]{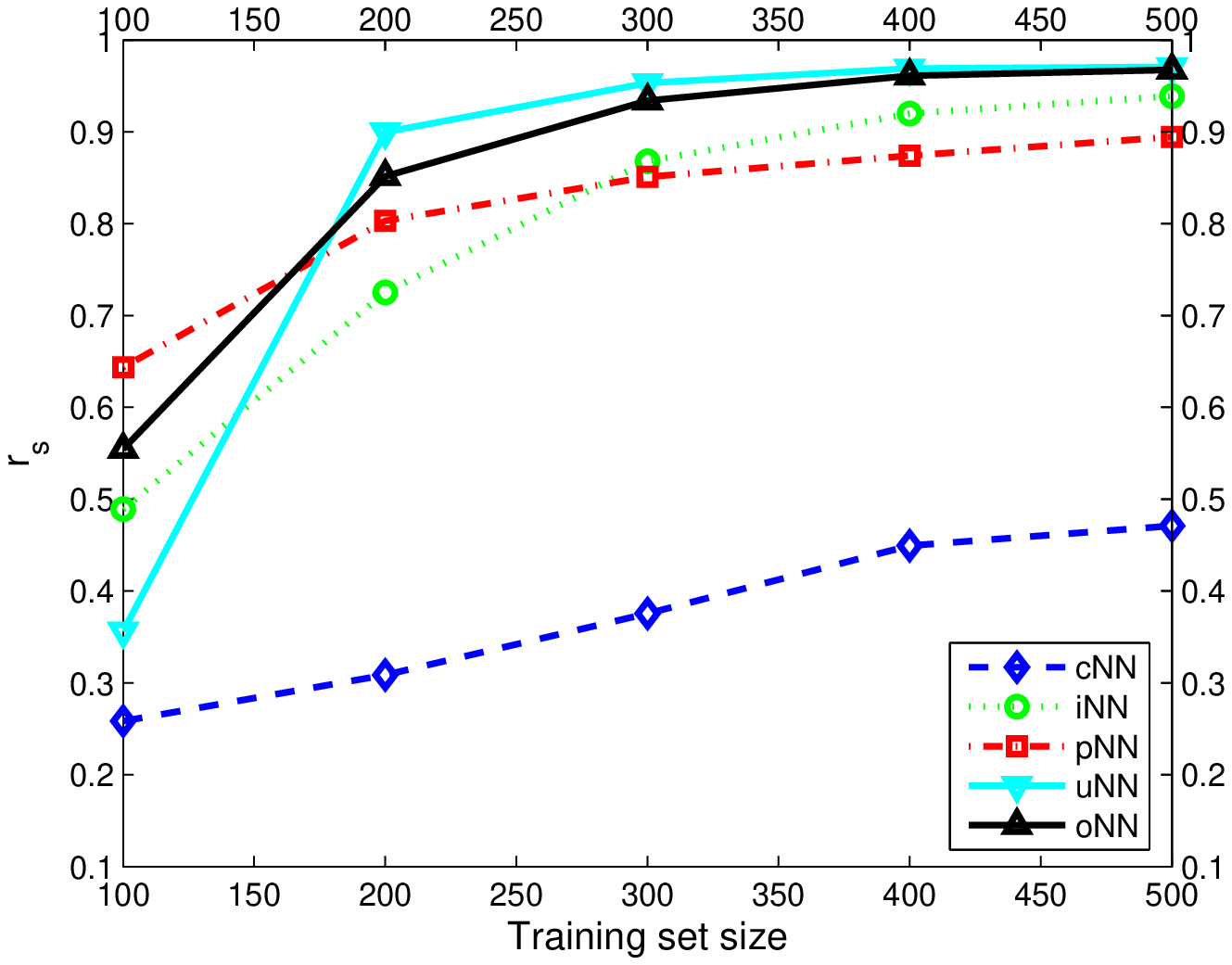}}  
\subfigure[Kendall's tau-b coefficient.]{
        \label{fig:synthetic0410Kendall}
        \includegraphics[width=0.31\linewidth]{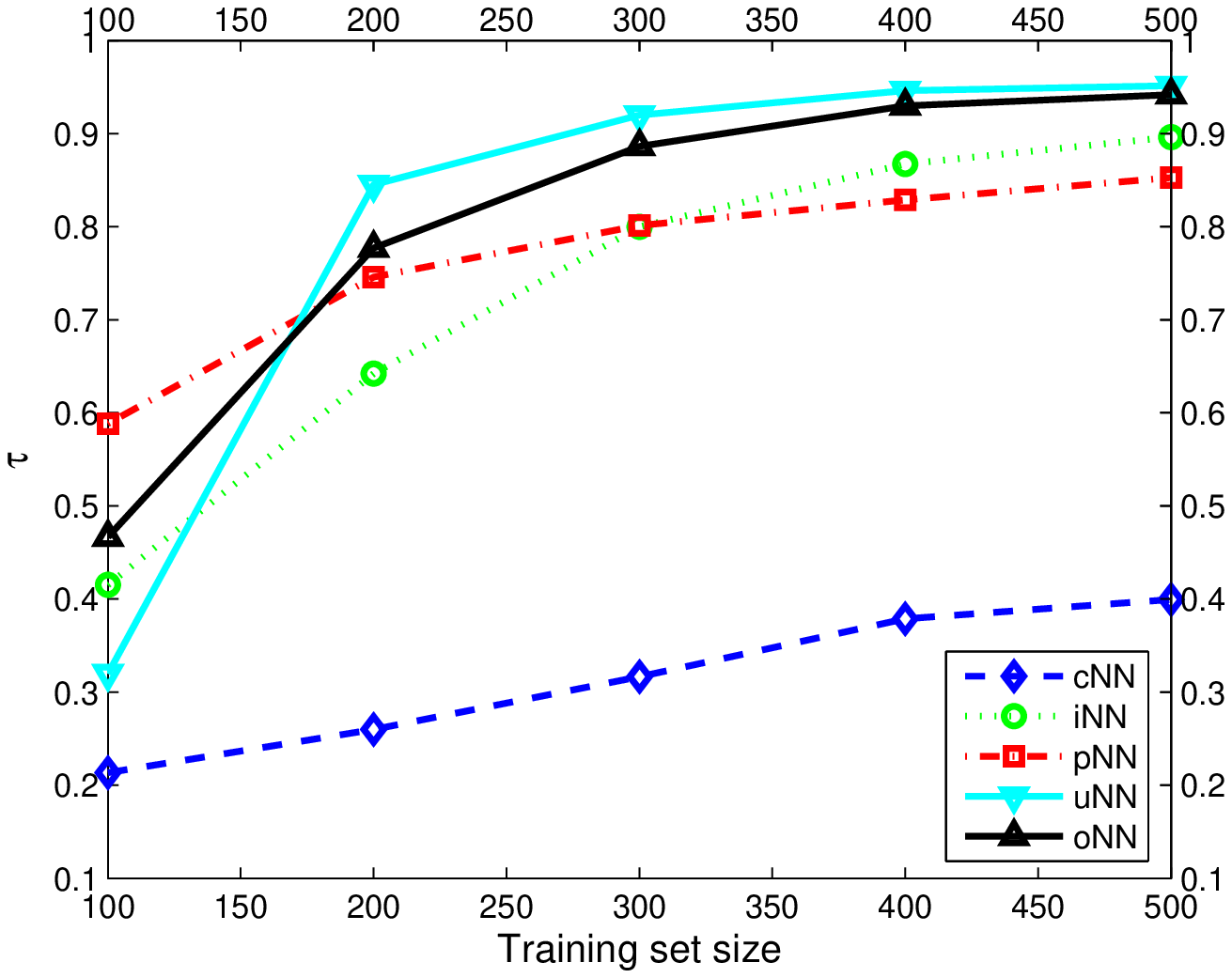}}  
\subfigure[$o_c$ coefficient.]{
        \label{fig:synthetic0410Prof}
        \includegraphics[width=0.31\linewidth]{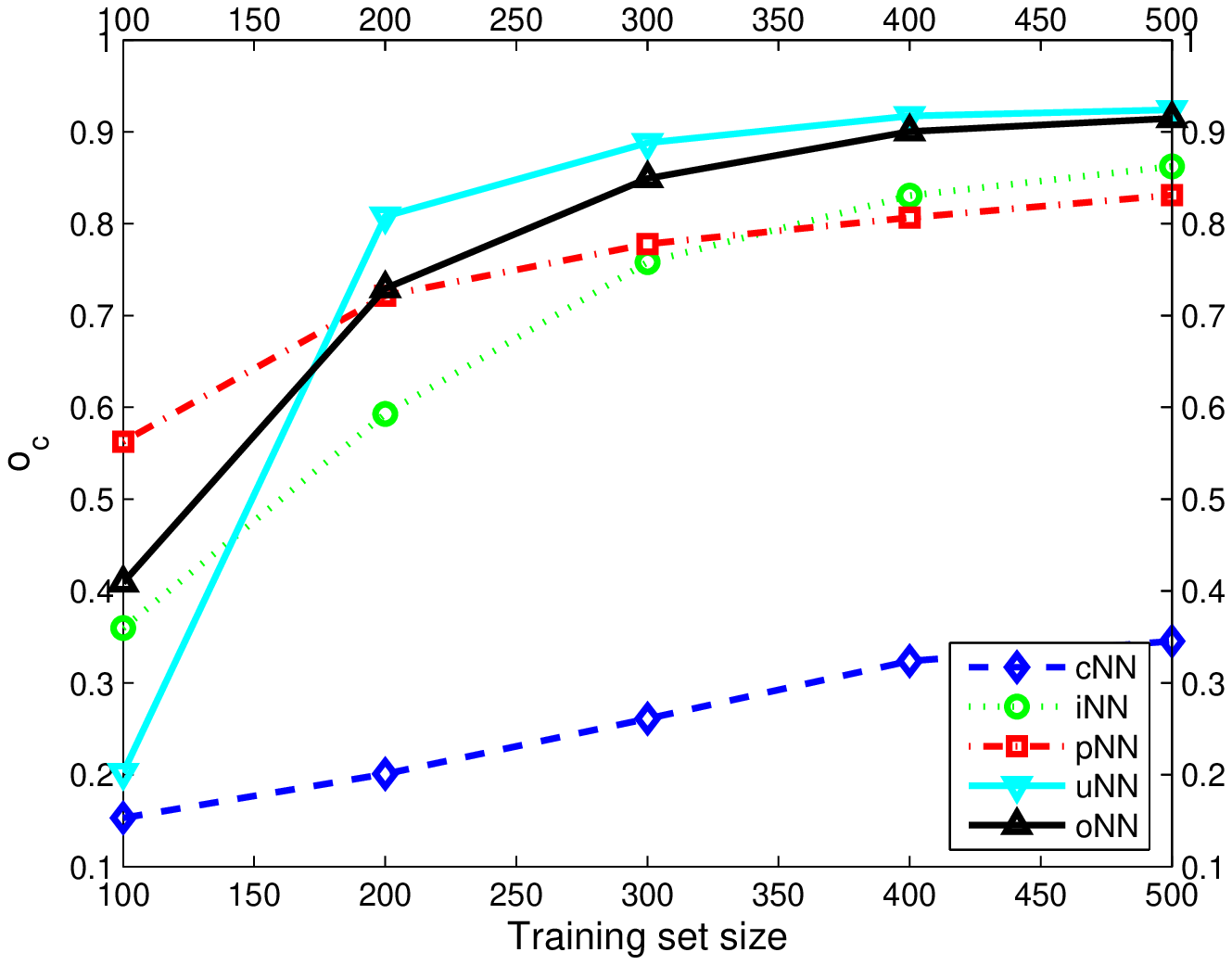}}                
\caption{NN results for $10$ classes in $\bbbr^4$, with 16 hidden units.}
\label{fig:synthetic0410NN}
\end{center}
\end{figure}

Finally, for $10$ classes the rank was assigned according to the rule 
$$y= \min_{r\in\{1, 2, 3, 4, 5, 6, 7, 8, 9, 10\}}\{r: b_{r-1} < 1000\prod_{i=1}^4 (x_i-0.5) + \varepsilon\ < b_r\}$$
\begin{multline*}(b_0, b_1, b_2, b_3,b_4, b_5, b_6, b_7, b_8, b_9, b_{10})=(-\infty, -5, -2.5, -1, -0.4, 0.1, 0.5, 1.1, 3, 6, +\infty)\end{multline*}
where $\varepsilon$ is a random value, normally distributed with zero mean and standard deviation $\sigma = 0.125$.
Class distributions are presented in figure \ref{fig:synthetic04};
the learning curves are shown in figures \ref{fig:synthetic0405NN} and \ref{fig:synthetic0410NN}, for 16 neurons in the hidden layer.

\subsection{Network complexity}
One final point to make in any comparison of methods regards complexity.
The number of learnable parameters for each model is presented in table \ref{tab:parameters}.

\begin{table}
\begin{center}
\begin{footnotesize}
\begin{tabular}{|l|c|c|c|c|c|}
\hline
Model & cNN & pNN & iNN & uNN & oNN\\
\hline
\hline
$\bbbr^2$, $K=5$ & 45 & $21\times 4$ & 39 & 21 & 23\\
\hline
$\bbbr^2$, $K=10$ & 75 & $21\times 9$ & 69 & 21 & 28\\
\hline
$\bbbr^4$, $K=5$  & 165 & $97\times 4$ & 148 & 97 & 100\\
\hline
$\bbbr^4$, $K=10$  & 250 & $97\times 9$ & 233 & 97 & 105\\
\hline
\end{tabular}
\end{footnotesize}
\end{center}
\caption{Number of parameters for each neural network model.}
\label{tab:parameters}
\end{table}

\section{Results for SVM methods}

\begin{figure}
\begin{center}
\subfigure[$5$ classes in $\bbbr^2$. $C=10000$, $h=10$, K(\textbf{x}, \textbf{y}) = $(1+\textbf{x}^t\textbf{y})^2$.]{
        \label{fig:SVMsynthetic0205ErrorRate}
        \includegraphics[width=0.35\linewidth]{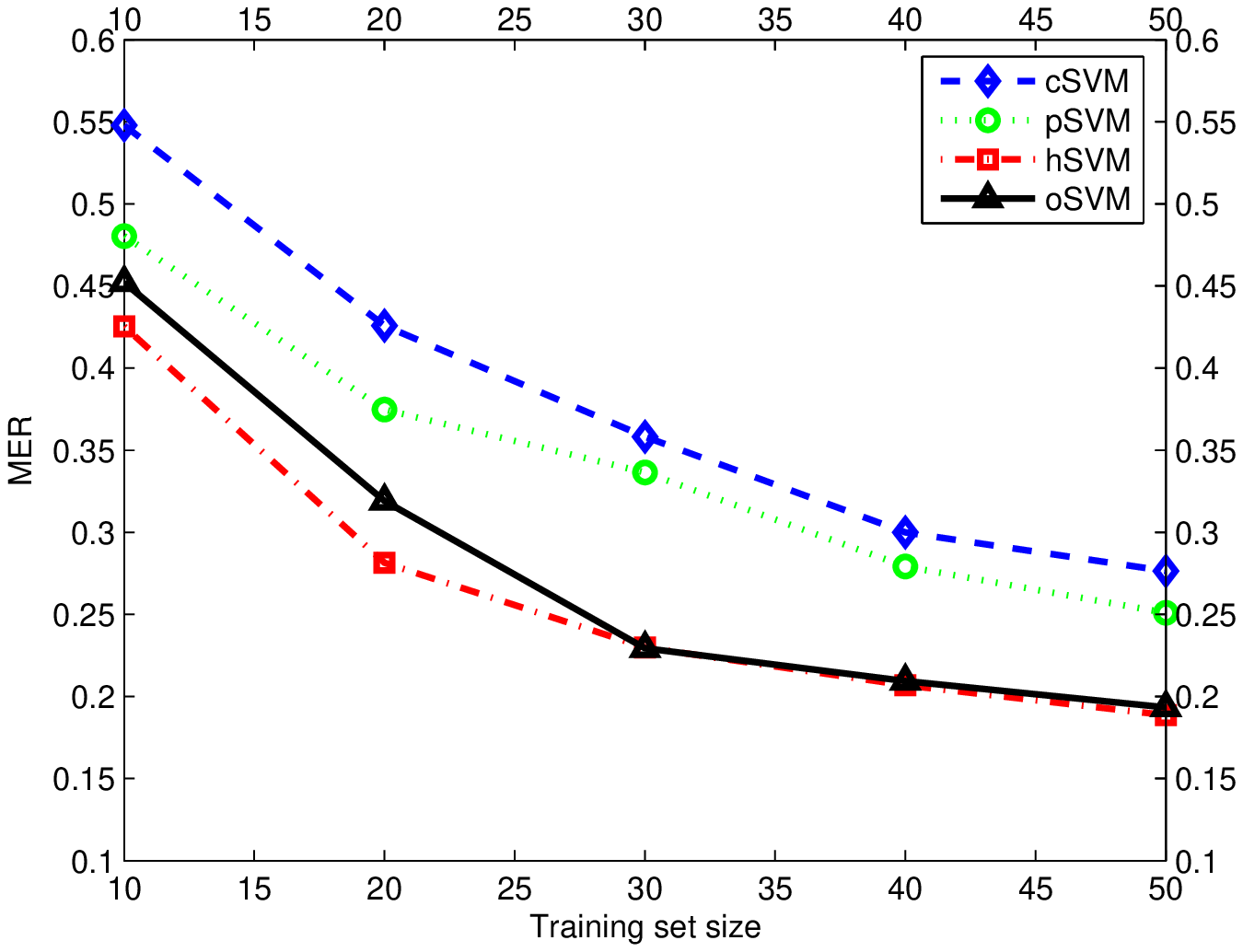}}
\subfigure[$10$ classes in $\bbbr^2$. $C=10000$, $h=10$, K(\textbf{x}, \textbf{y}) = $(1+\textbf{x}^t\textbf{y})^2$.]{
        \label{fig:SVMsynthetic0210ErrorRate}
        \includegraphics[width=0.35\linewidth]{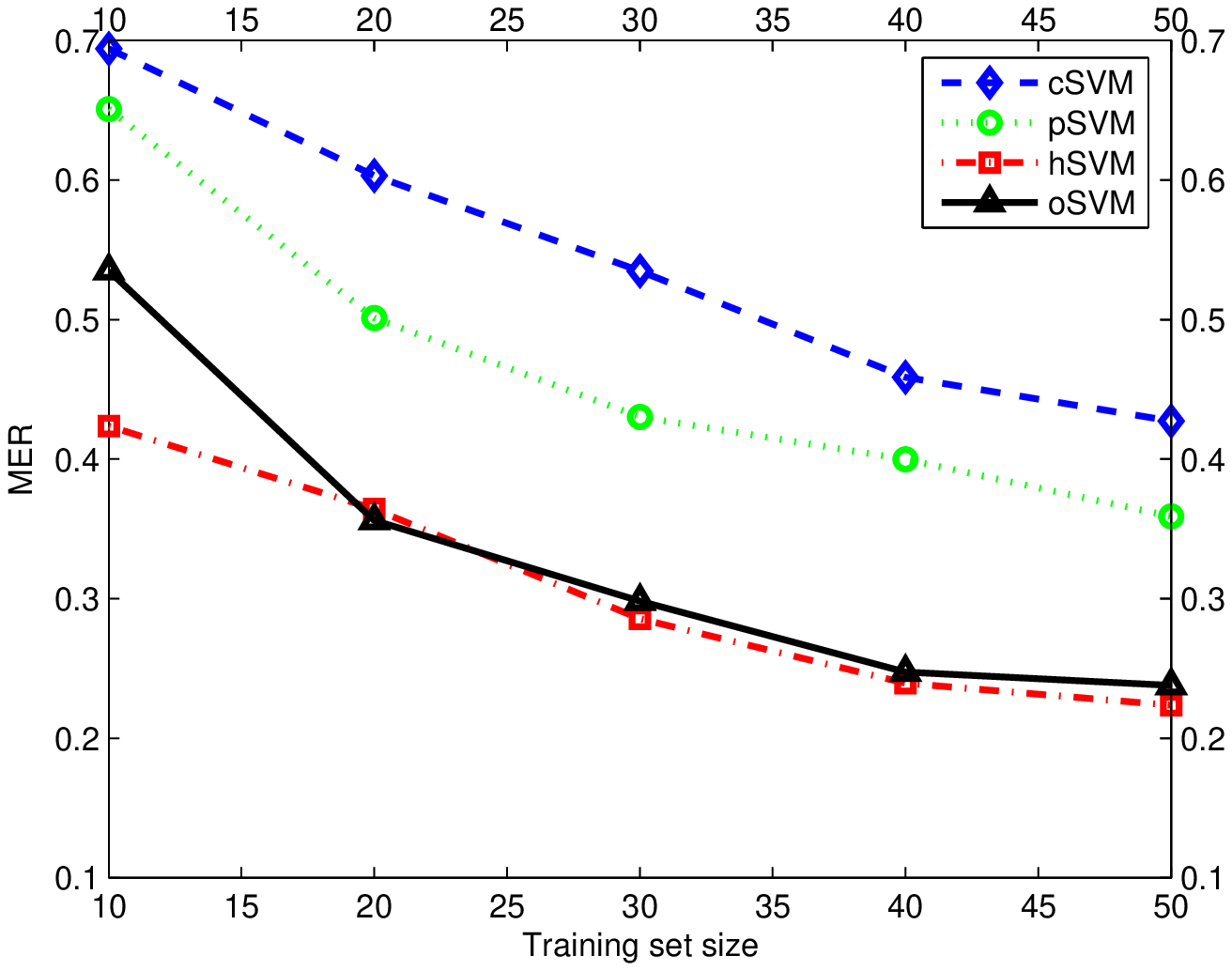}} 
\subfigure[$5$ classes in $\bbbr^4$. $C=10000$, $h=10$, K(\textbf{x}, \textbf{y}) = $(1+\textbf{x}^t\textbf{y})^4$.]{
        \label{fig:SVMsynthetic0405ErrorRate}
        \includegraphics[width=0.35\linewidth]{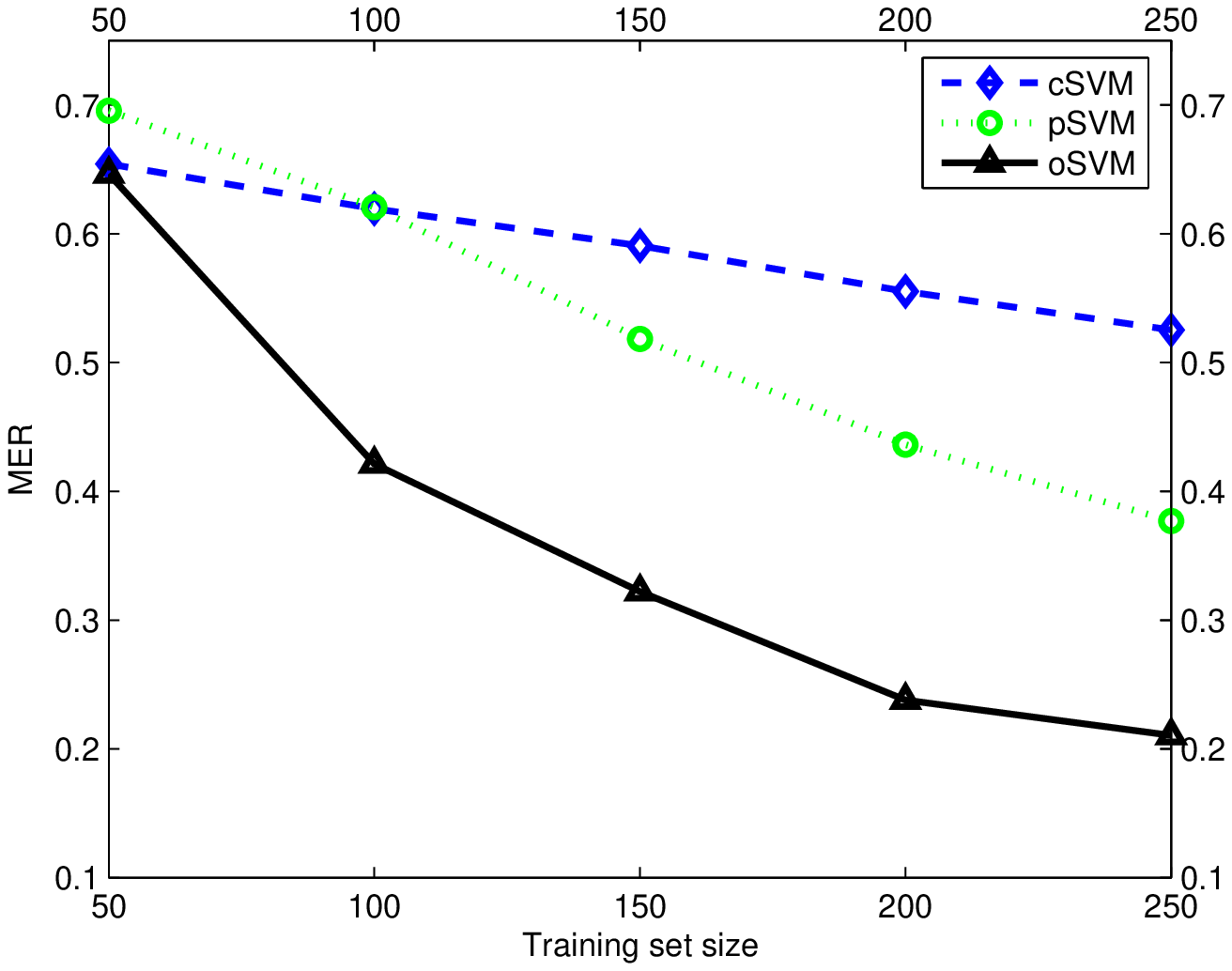}}
\subfigure[$10$ classes in $\bbbr^4$. $C=10000$, $h=10$, K(\textbf{x}, \textbf{y}) = $(1+\textbf{x}^t\textbf{y})^4$.]{
        \label{fig:SVMsynthetic0410ErrorRate}
        \includegraphics[width=0.35\linewidth]{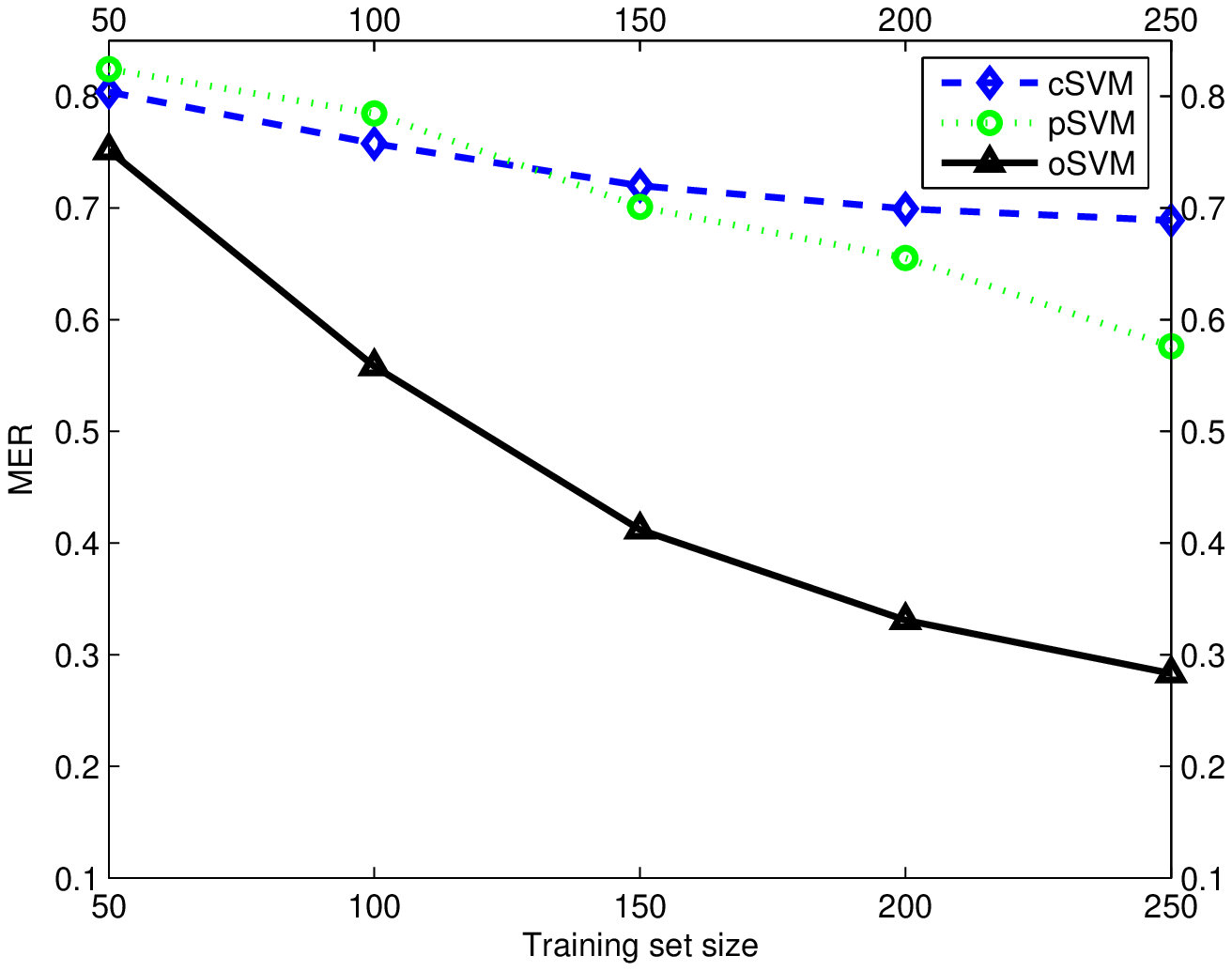}}                           
\caption{SVM results -- MER criterion.}
\label{fig:syntheticSVM}
\end{center}
\end{figure}

Because the comparative study for the SVM based methods followed the same reasoning as for the neural network methods, we restrict to present here the attained results in figures \ref{fig:SVMsynthetic0205ErrorRate}, \ref{fig:SVMsynthetic0210ErrorRate}, \ref{fig:SVMsynthetic0405ErrorRate} and \ref{fig:SVMsynthetic0410ErrorRate}. Because all classification indices portrayed essentially the same relative performance, and to facilitate the comparison with results previously reported in the literature, we will restrict here and in the future to the MER criterion.

\section{Discussion}
A first comment relates to the unfairness of comparing SVM to NN based methods since the kernel parameters were illegally tuned to the datasets. 
The main assertions concerns the superiority of all algorithms specific to ordinal data over conventional methods, both for SVMs and NNs; the proposed method, in spite of being the simplest model, performs as good or better than the other models under comparison.

\chapter{Results for practical ordinal datasets}
\setcounter{footnote}{1}
The next sections present results for datasets with real data. 

\section{Pasture production}
The next experiment is based on a publicly available dataset with real-life data, 
available at the WEKA website\footnote{\url{http://www.cs.waikato.ac.nz/ml/weka/}

The information is a replica of the notes made available with the data.}.
The objective was to predict pasture production from a variety of biophysical
factors. Vegetation and soil variables from areas of grazed North Island 
hill country with different management (fertilizer application/stocking rate)
histories (1973-1994) were measured and subdivided into 36 paddocks. Nineteen 
vegetation (including herbage production); soil chemical, physical and 
biological; and soil water variables were selected as potentially useful 
biophysical indicators -- table \ref{tab:INFOPASTURE}.
The target feature, the pasture production, has been categorized in three classes (Low, Medium, High), evenly distributed in the dataset of 36 instances.

\begin{table}[!ht]
\begin{center}
\begin{footnotesize}
\begin{tabular}{|c|c|c|}
\hline
Name & Data Type 	& Description\\
\hline
\hline
	fertiliser		&	enumerated	(LL, LN, HN, HH) 	& fertiliser used\\
	slope 				& integer			 									& slope of the paddock\\
	aspect-dev-NW	& integer			 									& the deviation from the north-west\\
	OlsenP				& integer												&\\
	MinN					& integer			 									&\\
	TS						& integer			 									&\\
	Ca-Mg					& real													& calcium magnesium ration\\
	LOM						& real 													& soil lom (g/100g)\\
	NFIX-mean			& real 				 									& a mean calculation\\
	Eworms-main-3 & real 													& main 3 spp earth worms per g/m2\\
	Eworms-No-species&integer											&number of spp\\
	KUnSat 				&  real													& mm/hr\\
	OM 						& real													& \\	
	Air-Perm 			& real & \\
	
 	Porosity 			& real &\\
 	HFRG-pct-mean & real &mean percent\\
 	legume-yield 	& real & kgDM/ha\\
	OSPP-pct-mean & real & mean percent\\
	Jan-Mar-mean-TDR & real & \\
	Annual-Mean-Runoff & real & mm\\
	root-surface-area & real & m2/m3\\
	Leaf-P 				& real & ppm \\
\hline
\end{tabular}
\end{footnotesize}
\end{center}
\caption{Characteristics of the 22 features of the Pasture dataset.}
\label{tab:INFOPASTURE}
\end{table}

The results attained are summarized in table \ref{tab:PASTURE}. Before training, the data was scaled to fall always within the range $[0, 1]$, 
using the transformation $x'=\frac{x-x_{\min}}{x_{\max}-x_{\min}}$.
The fertiliser attribute was represented using 4 variables: LL = (1, 0, 0, 0), LN = (0, 1, 0, 0), HL = (0, 0, 1, 0) and HH = (0, 0, 0, 1).

\begin{table}[!ht]
\begin{center}
\begin{footnotesize}
\subtable[SVMs' results.  $h = 100$, $s=1$, leave-one-out.]{
\begin{tabular}{|c|c|c|c|c|}
\hline
kernel& cSVM & pSVM & hSVM & oSVM\\
\hline
\hline
$K(\textbf{x},\textbf{y})=\textbf{x}^t\textbf{y}$ &  27.8 (C=0.2) & 27.8 (C=1.0) & 27.8 (C=0.01) & 27.8 (C=0.5) \\
\hline
$K(\textbf{x},\textbf{y})=(1+\textbf{x}^t\textbf{y})^2$  & 25.0 (C=0.04) & 25.0 (C=0.2) & 25.0 (C=0.01) & 22.2 (C=0.02)\\
\hline
\end{tabular}}\hfill
\subtable[NNs' results. $h = 1$, $s=2$, leave-one-out.]{
\begin{tabular}{|c|c|c|c|c|c|}
\hline
 	hidden units & cNN  & pNN 	& iNN   & uNN 	& oNN\\
\hline
\hline
0 						& 	35.6	& 48.1 & 34.2 & 56.7	& 55.0\\
\hline
4							& 	36.1 	& 37.5 & 33.6	& 35.3  & 38.3\\
\hline
\end{tabular}}
\end{footnotesize}
\end{center}
\caption{MER (\%) for the Pasture dataset.}
\label{tab:PASTURE}
\end{table}
The lack of motivation to impose an ordered relation in the fertiliser attribute, suggests a good scenario to apply the general version of the data replication method, where only 21 attributes ($j=21$) are constrained to have the same direction, with the fertiliser attribute left free. Using a linear kernel with $C=0.5$ ($h=100$, $s=1$) emerges a classifier with expected MER of 22.2\%. This way, a very simple classifier was obtained at the best performance.

\section{Employee selection: the ESL dataset}
\begin{figure}[!ht]
\begin{center}
\includegraphics[width=.4\linewidth]{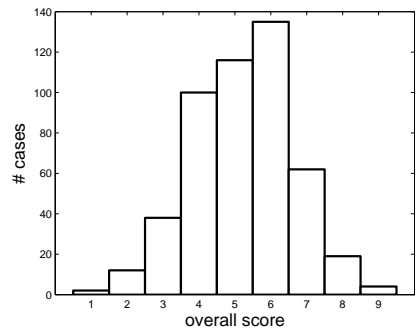}
\caption{Class distribution of 488 examples, for the ESL dataset.}
\label{fig:ESLhist}
\end{center}
\end{figure}

The next experiment is also based on a publicly dataset available at the WEKA website. 
The ESL dataset contains $488$ profiles of applicants for certain industrial jobs.  
Expert psychologists of a recruiting company, based upon psychometric test results and interviews with the candidates, determined the values of the input attributes ($4$ attributes, with integer values from $0$ to $9$). 
The output is an overall score ($1..9$) corresponding to the degree of fitness of the candidate to this type of job, distributed according to figure \ref{fig:ESLhist}.

\begin{figure}
\begin{center}
\subfigure[SVM results.  $K(\textbf{x}, \textbf{y}) = x^ty$, $C=100$, $h=0.5$.]{
        \label{fig:SVMwekaESLErrorRate}
        \includegraphics[width=0.40\linewidth]{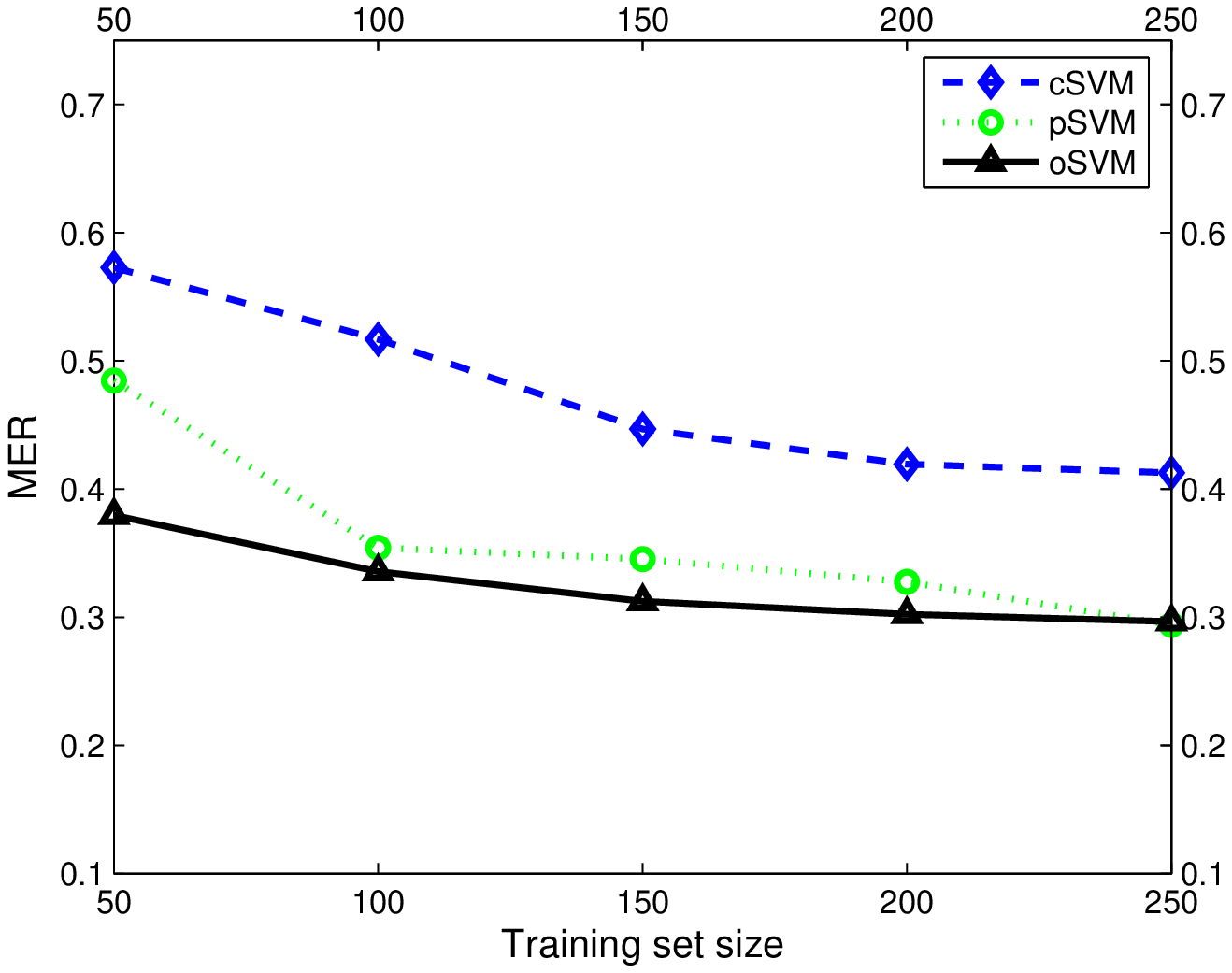}} 
\subfigure[NN results. $h=0.5$, no hidden layer.]{
        \label{fig:NNwekaESLErrorRate}
        \includegraphics[width=0.40\linewidth]{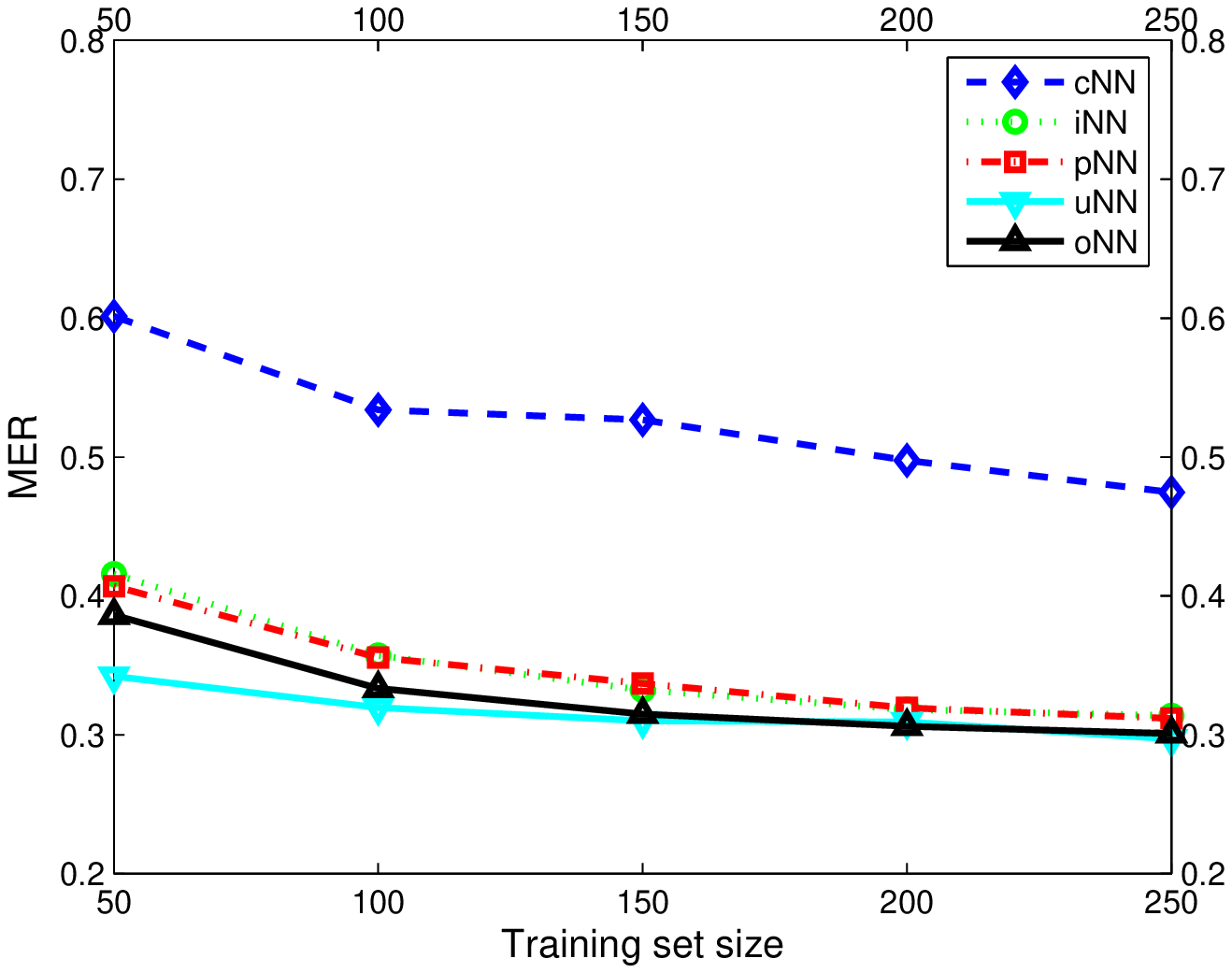}} 
\caption{Results for the ESL dataset, MER criterion.}
\label{figwekaESLResults}
\end{center}
\end{figure}

The comparative study of the learning algorithms followed the same reasoning as for the synthetic datasets; therefore we restrict to present here the attained results for the MER criterion -- figures \ref{fig:SVMwekaESLErrorRate} and \ref{fig:NNwekaESLErrorRate}. 

In the pasture dataset conventional methods performed as well as ordinal methods, while algorithms based on SVMs clearly outperformed NN based algorithms -- an expected result if we attend to the limited number of examples in the dataset. On the other side, for the ESL dataset, there is no discernible difference between SVM and NN based algorithms, but conventional methods are clearly behind specific methods for ordinal data.
\section[Aesthetic evaluation of breast cancer conservative treatment]{Aesthetic evaluation of breast cancer conservative treatment\footnotemark[4]}\footnotetext[4]{Some portions of this section appeared in \cite{JaimeNN2005,JaimeIJCNN2005}.}
\setcounter{footnote}{1}
\label{sec:bcct}
In this section we illustrate the application of the learning algorithms to the prediction of the cosmetic result of  breast cancer conservative treatment.

Breast cancer conservative treatment (BCCT) has been increasingly used over the last few years, as a consequence of its much more acceptable cosmetic outcome than traditional techniques, but with identical oncological results. 
Although considerable research has been put into BCCT techniques, diverse aesthetic results are common, highlighting the importance of this evaluation in institutions performing breast cancer treatment, so as to improve working practices. 

Traditionally, aesthetic evaluation has been performed subjectively by one or more observers \cite{Harris1979, Beadle1984Dec, Pierquin1991}. 
However, this form of assessment has been shown to be poorly reproducible \cite{Pezner1985Mar, Sacchini1991, Sneeuw1992, Christie1996}, which creates uncertainty when comparing results between studies. It has also been demonstrated that observers with different backgrounds evaluate cases in different ways \cite{MJCardoso2005A}.

Objective methods of evaluation have emerged as a way to overcome the poor reproducibility of subjective assessment and have until now consisted of measurements between identifiable points on patient photographs \cite{Pezner1985Mar, Limbergen1989, Christie1996}. 
The correlation of objective measurements with subjective overall evaluation has been reported by several authors 
\cite{Sacchini1991,  Sneeuw1992, Christie1996, Al-Ghazal1999Dec}.
Until now though, the overall cosmetic outcome was simply the sum of the individual scores of subjective and objective individual indices \cite{Noguchi1991, Sacchini1991, Sneeuw1992, Al-Ghazal1999Dec}.

\subsection{Data and method}
Instead of heuristically weighting the individual indices in an overall measure, 
we introduced pattern classification techniques to find the correct contribution of each individual 
feature in the final result and the scale intervals for each class, constructing this way an optimal rule to classify patients.

\subsubsection{Reference classification}

\begin{figure}[!ht]
\begin{center}
\includegraphics[width=\linewidth]{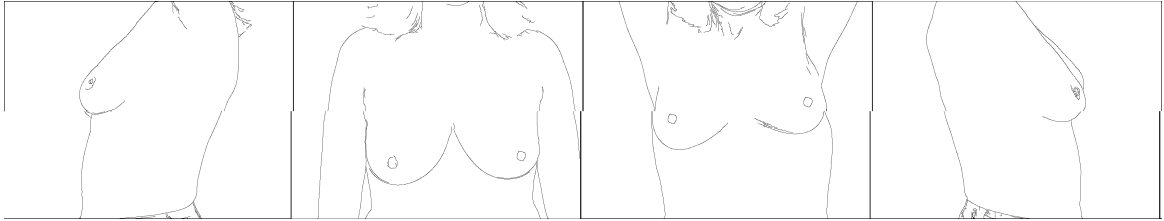}
\caption{Positions used in the photographs.}
\label{fig:four_positions}
\end{center}
\end{figure}

Twenty-four clinicians working in twelve different countries were selected, based on their experience in BCCT (number of cases seen per year and/or participation in published work on evaluation of aesthetic results). They were asked to evaluate individually a series of 240 photographs taken from 60 women submitted to BCCT (surgery and radiotherapy). 
Photographs were taken (with a 4M digital camera) in four positions with the patient standing on floor marks: facing, arms down; facing, arms up; left side, arms up; right side, arms up -- figure \ref{fig:four_positions}. 

Participants were asked to evaluate overall aesthetic results, classifying each case into one of four categories: \emph{excellent} -- treated breast nearly identical to untreated breast; \emph{good} -- treated breast slightly different from untreated; \emph{fair} -- treated breast clearly different from untreated but not seriously distorted; \emph{poor} -- treated breast seriously distorted \cite{Harris1979}. 

In order to obtain a consensus among observers, the Delphi process was used \cite{Jones1995, Hasson2000}. Evaluation of each case was considered consensual when more than 50\% of observers provided the same classification. When this did not occur, another round of agreement between observers was performed. 
By means of the Delphi process each and every patient was classified in 
one of the four categories (table \ref{tab:classes}): {\em poor}, {\em fair}, {\em good}, and {\em excellent}.

\begin{table}
\begin{center}
\begin{footnotesize}
\begin{tabular}{|c|c|}
\hline
Class & \# cases\\
\hline
\hline
Poor & 7\\
Fair & 12\\
Good & 32\\
Excellent & 9\\
\hline
\end{tabular}
\end{footnotesize}
\end{center}
\caption{Distribution of patients over the four classes.}
\label{tab:classes}
\end{table}

The evaluation of two individual aesthetic characteristics, scar visibility and colour dissimilarities between the breasts, were asked to the panel, using the same grading scale: \emph{excellent}; \emph{good}; \emph{fair}; \emph{poor}. 

\subsubsection{Feature Selection}
\begin{figure}[!ht]
\begin{center}
\includegraphics[width=0.4\linewidth]{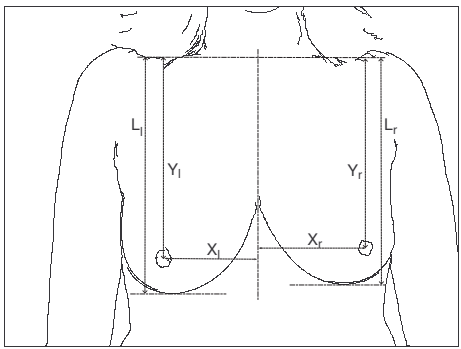}
\caption{LBC = $\abs{L_r - L_l}$, BRA = $\sqrt{(X_r-X_l)^2 + (Y_r-Y_l)^2}$, UNR = $\abs{Y_r - Y_l}$.}
\label{fig:measures}
\end{center}
\end{figure}

As possible objective features we considered those already identified by domain experts as relevant 
to the aesthetic evaluation of the surgical procedure \cite{Pezner1985Mar, Limbergen1989}.
The cosmetic result after breast conserving treatment is mainly determined by visible skin alterations or changes in breast volume or shape. Skin changes may consist of a disturbing surgical scar or radiation-induced pigmentation or telangiectasia \cite{Limbergen1989}. 
Breast asymmetry was assessed by Breast Retraction Assessment (BRA), Lower Breast Contour (LBC) or Upward Nipple Retraction (UNR) -- figure \ref{fig:measures}.
Because breast asymmetry was insufficient to discriminate among patients, we adopted  the mean of the scar visibility and skin colour change, as measured by the Delphi panel, as additional features to help in the separation task, as we had not yet established the evaluation of those features by quantitative methods \cite{JaimeIJCNN2005}.

\subsubsection{Classifier}
The {\em leave one out} method \cite{Duda2001} was selected for the validation of the classifiers: the 
classifier is trained in a round-robin fashion, each time using the available dataset from which a single 
patient has been deleted; each resulting classifier is then tested on the single deleted patient.

When in possession of a {\em nearly separable} dataset, a simple linear separator is bound to 
misclassify some points. 
But the real question is if the {\em non-linearly-separable} 
data indicates some intrinsic property of the problem (in which case a more complex classifier, 
allowing more general boundaries between classes may be more appropriate) or if 
it can be interpreted as the result of {\em noisy points} 
(measurement errors, uncertainty in class membership, etc), 
in which case keeping the linear separator and accept some errors is more natural.
Supported by Occam's razor principle 
(``{one should not increase, beyond what is necessary, the number of entities required to explain anything}''), 
the latter was the option taken in this research.

\subsection*{Datasets}
\begin{figure}
\begin{center}
\includegraphics[width=0.4\linewidth]{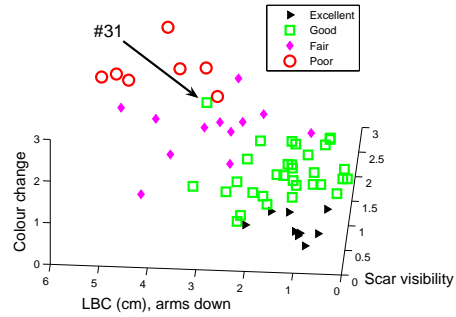}
\caption{Data points in a three-feature space.}
\label{fig:Triplet}
\end{center}
\end{figure}

A fast visual checking of the quality of the data (figure \ref{fig:Triplet}) shows that there is a data 
value that is logically inconsistent with the others: an individual (patient $\#31$) 
labeled as {\em good} when in fact it is placed between {\em fair} and {\em poor} in the feature space. 
The classifiers were evaluated using datasets with and without this outlier in order to assess the behaviour in the presence of noisy examples.
In summary, results are reported for six different datasets: 
\{\emph{LBC} (arms down); \emph{scar} visibility (mean); \emph{skin} colour change (mean)\}, 
\{\emph{BRA} (arms down); \emph{scar} visibility (mean); \emph{skin} colour change (mean)\}, 
\{\emph{UNR} (arms down); \emph{scar} visibility (mean); \emph{skin} colour change (mean)\}, 
each with $59$ and $60$ examples.
In \cite{JaimeIJCNN2005} other datasets were evaluated, showing similar behaviour. 

\subsection*{Results}

\begin{figure}
\begin{center}
\subfigure[Results for SVM methods. $C=10$, $h=100$, $K(\textbf{x}, \textbf{y})= \textbf{x}^t\textbf{y}$.]{
        \label{fig:bcctSVM}
        \includegraphics[width=0.35\linewidth]{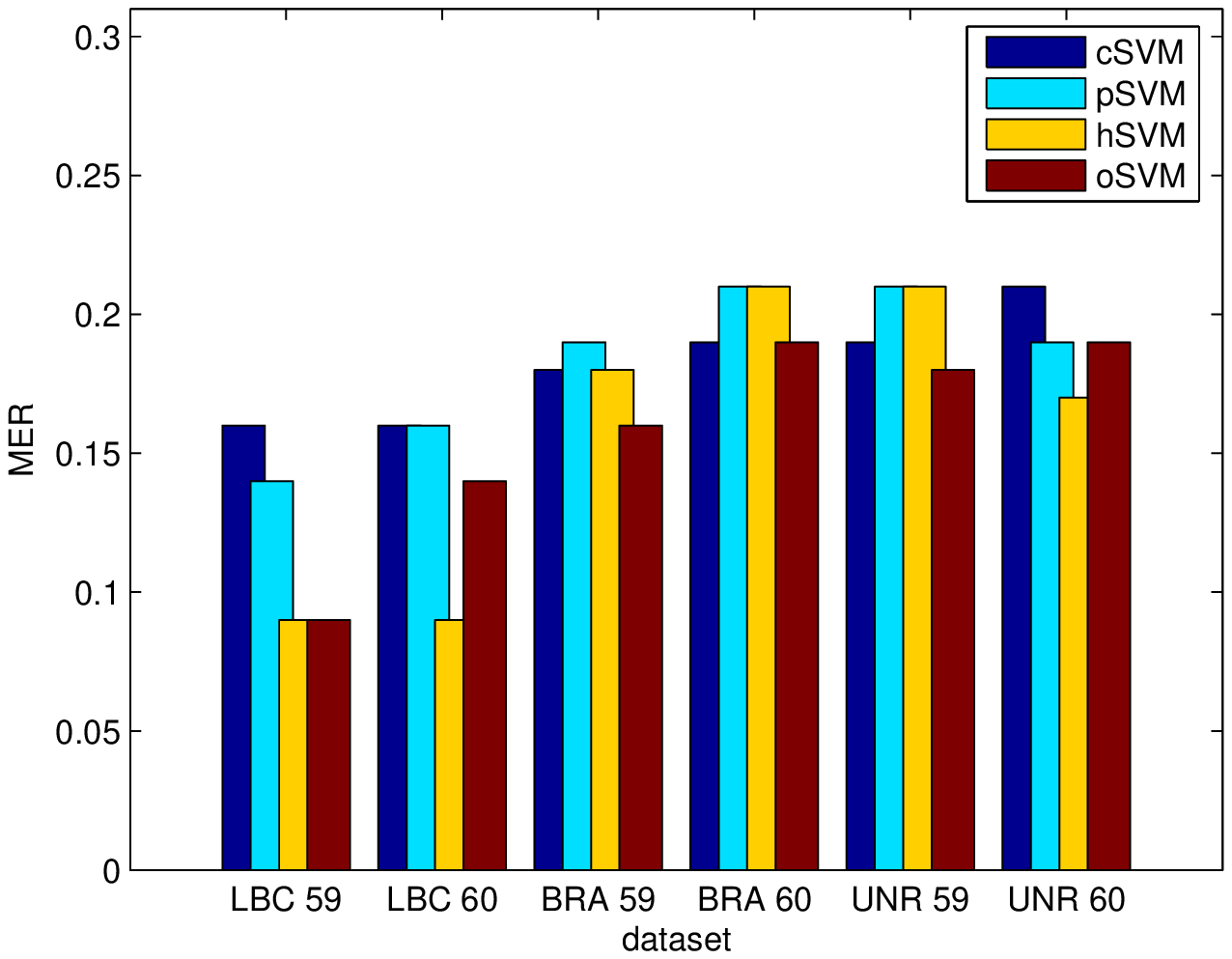}} 
\subfigure[Results for NN methods. $h=100$, no hidden units.]{
        \label{fig:bcctNN}
        \includegraphics[width=0.35\linewidth]{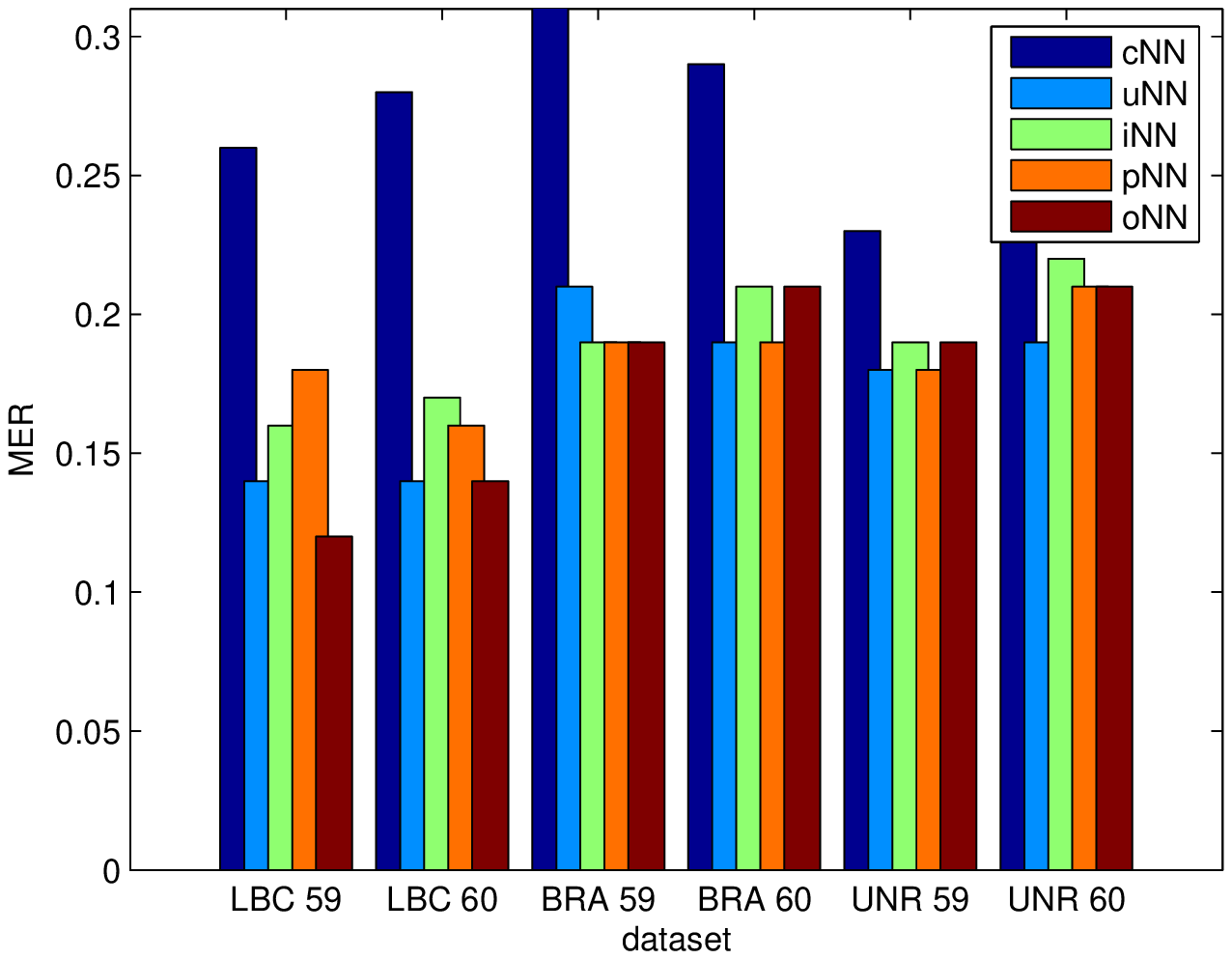}} 
\caption{Average of generalization error (MER).}
\label{fig:BCCT}
\end{center}
\end{figure}

The bar graph \ref{fig:BCCT} summarizes the generalization error estimated for each classifier. 
It is apparent that algorithms specially designed for ordinal data perform better than generic algorithms for nominal classes.
It is also noticeable the superiority of the LBC measure over the other asymmetry measures under study to discriminate classes.

\chapter{Results for datasets from regression problems}
\setcounter{footnote}{1}
Because of the general lack of benchmark datasets for ordinal classification, we also performed experiments with datasets from regression problems, by converting the target variable into an ordinal quantity.
The datasets were taken from a publicly available collection of regression problems\footnote{The datasets were selected from \url{http://www.liacc.up.pt/~ltorgo/Regression/DataSets.html}}. 

\section{Abalone dataset}
The goal is to predict the age of abalone from physical measurements.\footnote{The information is a replica of the notes for the abalone dataset from the UCI repository.}
The age of abalone is determined by cutting the shell through the cone, staining it,
   and counting the number of rings through a microscope -- a boring and
   time-consuming task.  Other measurements, which are easier to obtain, are
   used to predict the age.  Further information, such as weather patterns
   and location (hence food availability) may be required to solve the problem.

\begin{figure}[!ht]
\begin{center}
\includegraphics[width=.4\linewidth]{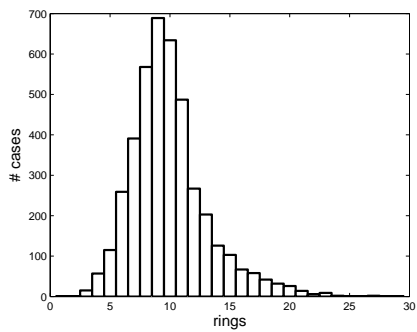}
\caption{Class distribution over 4177 examples, for the abalone dataset.}
\label{fig:alaboneHist}
\end{center}
\end{figure}

   Examples with missing values were removed from the original data  (the
   majority missing the predicted value), and the ranges of the
   continuous values have been scaled for the use with an artificial neural network (by dividing by 200).
The sex attribute was represented as $M=1$, $F=0$, $I=-1$.
The characteristics of the dataset are summarized in table \ref{tab:INFOABALONE}, where are listed the attribute name, attribute type, the measurement unit and a
   brief description; the class distribution is depicted in figure \ref{fig:alaboneHist}.
   
\begin{table}[t]
\begin{center}
\begin{footnotesize}
\begin{tabular}{|c|c|c|c|}
\hline
Name & Data Type & Meas.	& Description\\
\hline
\hline
	Sex	&	nominal		&	M, F, and I (infant)&\\
	Length	&	continuous&	mm&	Longest shell measurement\\
	Diameter&	continuous&	mm	&perpendicular to length\\
	Height	&	continuous	&mm	&with meat in shell\\
	Whole weight&	continuous&	grams	&whole abalone\\
	Shucked weight&	continuous&	grams	&weight of meat\\
	Viscera weight&	continuous&	grams	&gut weight (after bleeding)\\
	Shell weight	& continuous&	grams	&after being dried\\
	Rings	&	integer		&	&+1.5 gives the age in years\\
\hline
\end{tabular}
\end{footnotesize}
\end{center}
\caption{Characteristics of the abalone dataset.}
\label{tab:INFOABALONE}
\end{table}

The results obtained, table \ref{tab:ABALONE}, can be confronted with results reported in previous studies -- table \ref{tab:LeeABALONE}.

\begin{table}[!ht]
\begin{center}
\begin{footnotesize}
\subtable[Results reported in \cite{Frank1999}.]{
\begin{tabular}{|c|c|c|c|}
\hline
& C4.5 ORD & C 4.5 & C4.5-1 PC\\
\hline
\hline
3 classes  & 34.9 & 36.1 & 34.1\\
5 classes  & 51.9 & 53.7 & 50.5\\
10 classes & 70.6 & 73.3 & 72.6\\
\hline
\end{tabular}}\hfill
\subtable[Results reported in \cite{Lee2003}.]{
\begin{tabular}{|c|c|c|}
\hline
& CRT & MDT\\
\hline
\hline
3 classes  & 47.7 & 54.7\\
5 classes  & 62.1 & 57.8\\
10 classes & 78.4 & 70.7 \\
\hline
\end{tabular}}
\end{footnotesize}
\end{center}
\caption{MER (\%) for the Abalone dataset with decision trees, using equal frequency binning.}
\label{tab:LeeABALONE}
\end{table}

\begin{table}[!ht]
\begin{center}
\begin{footnotesize}
\subtable[SVMs' results.  $C=1000$, $h = 1$, $s=2$, $K(\textbf{x},\textbf{y})=\textbf{x}^t\textbf{y}$, training set size = 200.]{
\begin{tabular}{|c|c|c|c|c|}
\hline
& cSVM & pSVM & hSVM & oSVM\\
\hline
\hline
3 classes  & 37.4 & 36.8 & NA & 37.0\\
5 classes  & 53.3 & 54.2 & NA & 54.4\\
10 classes & 73.5 & 74.1 & NA & 73.7\\
\hline
\end{tabular}}\hfill
\subtable[NNs' results. $h = 1$, no hidden units, training set size = 200.]{
\begin{tabular}{|c|c|c|c|c|c|}
\hline
 					 & cNN  & pNN & iNN & uNN & oNN\\
\hline
\hline
3 classes  & 37.4 & 37.3 & 37.4 & 37.9 & 37.2\\
5 classes  & 53.0 & 53.9 & 54.3 & 60.5 & 55.5\\
10 classes & 73.3 & 74.0 & 74.7 & 80.8 & 75.9\\
\hline
\end{tabular}}
\end{footnotesize}
\end{center}
\caption{MER (\%) for the Abalone dataset, using equal frequency binning.}
\label{tab:ABALONE}
\end{table}
If it is true that generally we might prefer simpler models for explanation at the same performance -- a parsimonious representation of the observed data --, then the simple weighted sum of the attributes yielded by the data replication method is clearly in advantage.

\section{CPU performance dataset}
The goal is to predict the relative CPU performance.  From the 10 initial attributes 6 were used as predictive attributes and 1 as the goal field, discarding the vendor name, model name and estimated relative performance from the original article. The characteristics of the fields used are summarized in table \ref{tab:INFOCPU}, for the 209 instances. 
\begin{table}[!h]
\begin{center}
\begin{footnotesize}
\begin{tabular}{|c|c|c|c|c|}
\hline
Name & Data Type & Description & Min & Max\\
\hline
\hline
	MYCT	&	integer	&	machine cycle time in nanoseconds & 17 &  1500   \\
	MMIN	&	integer &	minimum main memory in kilobytes& 64  & 32000 \\
	MMAX  &	integer &	maximum main memory in kilobytes &64 &  64000 \\
	CACH	&	integer	& cache memory in kilobytes &0   & 256   \\
	CHMIN &	integer &	minimum channels in units& 0  &  52    \\
	CHMAX &	integer &	maximum channels in units &0  &  176   \\
	PRP   &	integer &	published relative performance &6  &  1150  \\
\hline
\end{tabular}
\end{footnotesize}
\end{center}
\caption{Characteristics of the CPU performance dataset.}
\label{tab:INFOCPU}
\end{table}

Before training, the predictive attributes were scaled to fall always within the range $[0, 1]$, using the transformation $x'=\frac{x-x_{\min}}{x_{\max}-x_{\min}}$.
The results obtained, table \ref{tab:MACHINECPU}, can be confronted with results reported in previous studies -- table \ref{tab:FrankMACHINECPU}.
\begin{table}[!h]
\begin{center}
\begin{footnotesize}
\subtable[Results reported in \cite{Frank1999}.]{
\begin{tabular}{|c|c|c|c|}
\hline
& C4.5 ORD & C 4.5 & C4.5-1 PC\\
\hline
\hline
3 classes & 26.1 & 28.2 & 25.7\\
5 classes & 41.9 & 43.2 & 43.4\\
10 classes & 63.5 & 63.8 & 69.4\\
\hline
\end{tabular}}\hfill
\subtable[Results reported in \cite{Lee2003}.]{
\begin{tabular}{|c|c|c|}
\hline
& CRT & MDT\\
\hline
\hline
3 classes & 45.9 & 31.1 \\
5 classes & 45.0 & 40.7 \\
10 classes & 57.9 & 57.4 \\
\hline
\end{tabular}}
\end{footnotesize}
\end{center}
\caption{MER (\%) for the Machine CPU dataset, using decision trees.}
\label{tab:FrankMACHINECPU}
\end{table}
\begin{table}[!h]
\begin{center}
\begin{footnotesize}
\subtable[SVMs' results. $C=1000$, $h = 1$, $s=1$, $K(\textbf{x},\textbf{y})=\textbf{x}^t\textbf{y}$, training set size = 190.]{
\begin{tabular}{|c|c|c|c|c|}
\hline
& cSVM & pSVM & hSVM & oSVM\\
\hline
\hline
3 classes  & 23.9 & 22.9 & NA & 23.4\\
5 classes  & 39.7 & 43.0 & NA & 44.6\\
10 classes & 69.0 & 65.4 & NA & 67.8\\
\hline
\end{tabular}}\hfill
\subtable[NNs' results.  $h=1$, without hidden layers.]{
\begin{tabular}{|c|c|c|c|c|c|}
\hline
 & cNN & pNN & iNN & uNN & oNN\\
\hline
\hline
3 classes  & 23.8 & 22.3 & 25.5 & 24.3 & 23.4\\
5 classes  & 42.0 & 42.4 & 42.7 & 42.5 & 42.8\\
10 classes & 65.7 & 66.6 & 68.1 & 68.3 & 65.3\\
\hline
\end{tabular}}
\end{footnotesize}
\end{center}
\caption{MER (\%) for the Machine CPU dataset.}
\label{tab:MACHINECPU}
\end{table}

These results continue to suggest the merit of specific methods for ordinal data over conventional methods, attaining the best performance at the greatest simplicity.

\chapter{Conclusion}
\setcounter{footnote}{1}
\label{chap:discussion}

This study focuses on the application of machine learning methods, and in particular of neural networks and support vector machines, to the problem of classifying ordinal data.
Two novel approaches to train learning algorithms for ordinal data were presented.
The first idea is to reduce the problem to the standard two-class setting, using the so called \emph{data replication method}, a nonparametric procedure for the classification of ordinal categorical data. This method was mapped into neural networks and support vector machines. 
Two well-known approaches for the classification of ordinal categorical data were unified under this framework, the minimum margin principle \cite{Shashua2002A} and the generic approach by Frank and Hall \cite{Frank1999}.
Finally, it was also presented a probabilistic interpretation for the neural network model.

The second idea is to retain the ordinality of the classes by imposing a parametric model for the output probabilities.
The introduced unimodal model, mapped to neural networks, was then confronted with established regression methods.

The study compares the results of the proposed models with conventional learning algorithms for nominal classes and with models proposed in the literature specifically for ordinal data.
Simple misclassification, mean absolute error, root mean square error, Spearman and Kendall's tau-b coefficients are used as measures of performance for all models and used for model comparison.
The new methods are likely to produce simpler and more robust classifiers, and compare favourably with state-of-the-art methods.
However, the reported results must be taken with caution. In most of the experiments the effort to find the correct setting for the parameters of the algorithms was limited (although unbiased among methods). So, although reasonable conclusions are drawn from the experiments, we do not wish to overstate our claims. 

This thesis has covered the multiclass classification accuracy and classifier simplicity, but a brief word on speed is in order. Comparing different machine learning algorithms for speed is notoriously difficult; we are simultaneously judging mathematical algorithms and specific implementations. However, some useful general observations can be made. Empirically, SVM training time tends to be superlinear in the number of the training points \cite{Rifkin2004}. Armed only with this assumption, it is a simple exercise to conclude that the complexity of the data replication formulation is placed between the simple approach of Frank and Hall and the pairwise of Herbrich.  

An issue intentionally avoided until now was the very own definition of ordinal classes. Although we do not wish to delve deeply on that now, a few comments are in order.

Apparently, a model that restricts the search to noncrossing boundaries is too restrictive, imposing unnecessary and unnatural constraints on the solution, limiting this way the feasible solutions to a subset of what we would expect to be a valid solution to an ordinal data problem. 
On the other side, the unimodal model, more plausible and intuitive, seems to capture better the essence of the problem.
However, it is a simple exercise to verify that the unimodal model does not allow boundaries' intersections -- the intersection point would indicate an example where three or more classes are equally probable. 
For that reason, the unimodal model (parametric or not) seems to be a subset of the noncrossing boundaries model.
It is also reasonable to accept that each noncrossing boundary solution may be explained by, at least, an unimodal model (however, there is not a bijection between the two, as different unimodal models may lead to the same noncrossing boundary solution; in fact, non-unimodal models may also lead to noncrossing boundaries). 

It is visible here a resemblance with the parallelism between parametric classifiers that must estimate the probability density function for each class in order to apply the bayes likelihood ratio test and classifiers that specify the mathematical form of the classifier (linear, quadratic, etc), leaving a finite set of parameters to be determined.

We are not advocating any model in particular. Pragmatically, see them as two more tools: only testing will say which is best in a specific machine learning problem.

Finally, as all unfinished jobs, this also leaves some interesting anchors for future work. 
As mentioned in this thesis, several unimodal models were implemented making use of a generic optimization function available in Matlab. It would be most interesting to adapt the backpropagation method to all unimodal models and perform a fair comparison. 
The data replication method is parameterised by $h$ (and $C$); because it may be difficult and time consuming to choose the best value for $h$, it would be interesting to study possible ways to automatically set this parameter, probably as a function of the data and $C$.
It would also be interesting to study if these algorithms can be successfully applied to nominal data. 
Although the data replication method was designed for ordinal classes, nothing impedes its application to nominal classes.
It is expected that the classifier should be evaluated for each possible order of the classes, choosing the one conducting to the best performance (feasible only when the number of classes is small).
A systematic study of decision trees' algorithms for ordinal data is also indispensable. It would be a significant accomplishment to map the models introduced in this thesis to decision trees.

\renewcommand{\bibname}{References}
\addcontentsline{toc}{chapter}{References}
\bibliographystyle{IEEEtran}
\bibliography{classificationOrdinalData}

\end{document}